\documentclass[journal]{IEEEtran}

\usepackage{xspace}
\makeatletter
\DeclareRobustCommand\onedot{\futurelet\@let@token\@onedot}
\def\@onedot{\ifx\@let@token.\else.\null\fi\xspace}

\def\eg{\emph{e.g}\onedot} 
\def\ie{\emph{i.e}\onedot}

\def\etal{\emph{et al}\onedot}
\makeatother
\ifCLASSINFOpdf
\else
\fi
\usepackage{amssymb}
\usepackage{amsmath}

\usepackage[colorlinks,
            linkcolor=red,
            anchorcolor=blue,
            citecolor=green]{hyperref}

\usepackage{tabu}
\usepackage{comment}
\usepackage{graphicx}
\usepackage{amsmath,bm}
\usepackage{mathtools}
\usepackage{booktabs}
\usepackage{multirow}
\usepackage{pifont}
\usepackage{color,xcolor}
\usepackage{adjustbox}
\usepackage{url}            % simple URL typesetting
\usepackage{subcaption}
\usepackage{caption}
\usepackage{enumitem}
\usepackage{multicol}
\usepackage{adjustbox}
\usepackage{colortbl}
\usepackage{tikz, pgfplots}

\begin{document}
\title{OccNeRF: Advancing 3D Occupancy Prediction in LiDAR-Free Environments} 
\author{Chubin Zhang, Juncheng Yan, Yi Wei, Jiaxin Li, Li Liu,\\ Yansong Tang,~\IEEEmembership{Member,~IEEE}, Yueqi Duan,~\IEEEmembership{Member,~IEEE},	and~Jiwen~Lu, \IEEEmembership{Fellow,~IEEE} 
	\thanks{\IEEEcompsocthanksitem The first three authors contribute equally.}
        \thanks{\IEEEcompsocthanksitem Chubin Zhang and Yansong Tang are with the Shenzhen International Graduate School, Tsinghua University, Shenzhen, 518055, China. Email: zcb24@mails.tsinghua.edu.cn; tang.yansong@sz.tsinghua.edu.cn.}
        \thanks{\IEEEcompsocthanksitem Juncheng Yan, Yi Wei and Jiwen Lu are with the Department of Automation, Yueqi Duan is with the Department of Electronic Engineering, Tsinghua University, Beijing, 100084, China. Jiaxin Li is with Gaussian Robotics, Shanghai, 201100, China. Li Liu is with Xiaomi EV, Beijing, 100085, China.}}

\maketitle

%%%%%%%%% ABSTRACT

\begin{abstract}
   Occupancy prediction reconstructs 3D structures of surrounding environments. It provides detailed information for autonomous driving planning and navigation. However, most existing methods heavily rely on the LiDAR point clouds to generate occupancy ground truth, which is not available in the vision-based system. In this paper, we propose an OccNeRF method for training occupancy networks without 3D supervision. Different from previous works which consider a bounded scene, we parameterize the reconstructed occupancy fields and reorganize the sampling strategy to align with the cameras' infinite perceptive range. The neural rendering is adopted to convert occupancy fields to multi-camera depth maps, supervised by multi-frame photometric consistency.  Moreover, for semantic occupancy prediction, we design several strategies to polish the prompts and filter the outputs of a pretrained open-vocabulary 2D segmentation model. Extensive experiments for both self-supervised depth estimation and 3D occupancy prediction tasks on nuScenes and SemanticKITTI datasets demonstrate the effectiveness of our method. The code is available at \url{https://github.com/LinShan-Bin/OccNeRF}.
\end{abstract}



% \fi
\IEEEpeerreviewmaketitle

\section{Introduction}
\label{sec:intro}
\IEEEPARstart{R}{ecent} years have witnessed the great process of autonomous driving ~\cite{bevdepth,bevformer,surroundocc,bevfusion}. As a crucial component, 3D perception helps the model to understand the real 3D world. Although LiDAR provides a direct means to capture geometric data, its adoption is hindered by the expense of sensors and the sparsity of scanned points. In contrast, as a cheap while effective solution, the vision-centric methods~\cite{lu2019monocular,beverse,fiery,stretchbev,bevfusion,bevfusion2} have received more and more attention.  Among various 3D scene understanding tasks, multi-camera 3D object detection~\cite{bevformer,bevdepth,petr,bevdet} plays an important role in autonomous systems. However, it struggles to detect objects from infinite classes and suffers from long-tail problems as illustrated in~\cite{surroundocc,yu2023flashocc}. % TODO

Complementary to 3D object detection, 3D occupancy prediction~\cite{occ3d,occformer,openoccupancy,monoscene} reconstructs the geometric structure of surrounding scenes directly, which naturally alleviates the problems mentioned above. As mentioned in ~\cite{surroundocc}, 3D occupancy is a good 3D representation for multi-camera scene reconstruction since it has the potential to reconstruct occluded parts and guarantees multi-camera consistency. Recently, some methods have been proposed to lift image features to the 3D space and further predict 3D occupancy. However, most of these methods need 3D occupancy labels for supervision. While some previous research~\cite{surroundocc,occ3d} has employed multi-frame LiDAR point accumulation to automatically label occupancy ground truth, the volume of LiDAR data is significantly less than that of image data. Collecting LiDAR data requires specialized vehicles equipped with LiDAR sensors, which is costly. Moreover, this approach neglects a vast quantity of unlabeled multi-camera image data. Consequently, investigating LiDAR-free methods for training occupancy presents a promising research avenue.

\begin{figure}[tb]%{0.6\textwidth}
    %\vspace{-30pt}
      \begin{center}
       \includegraphics[width=0.48\textwidth]{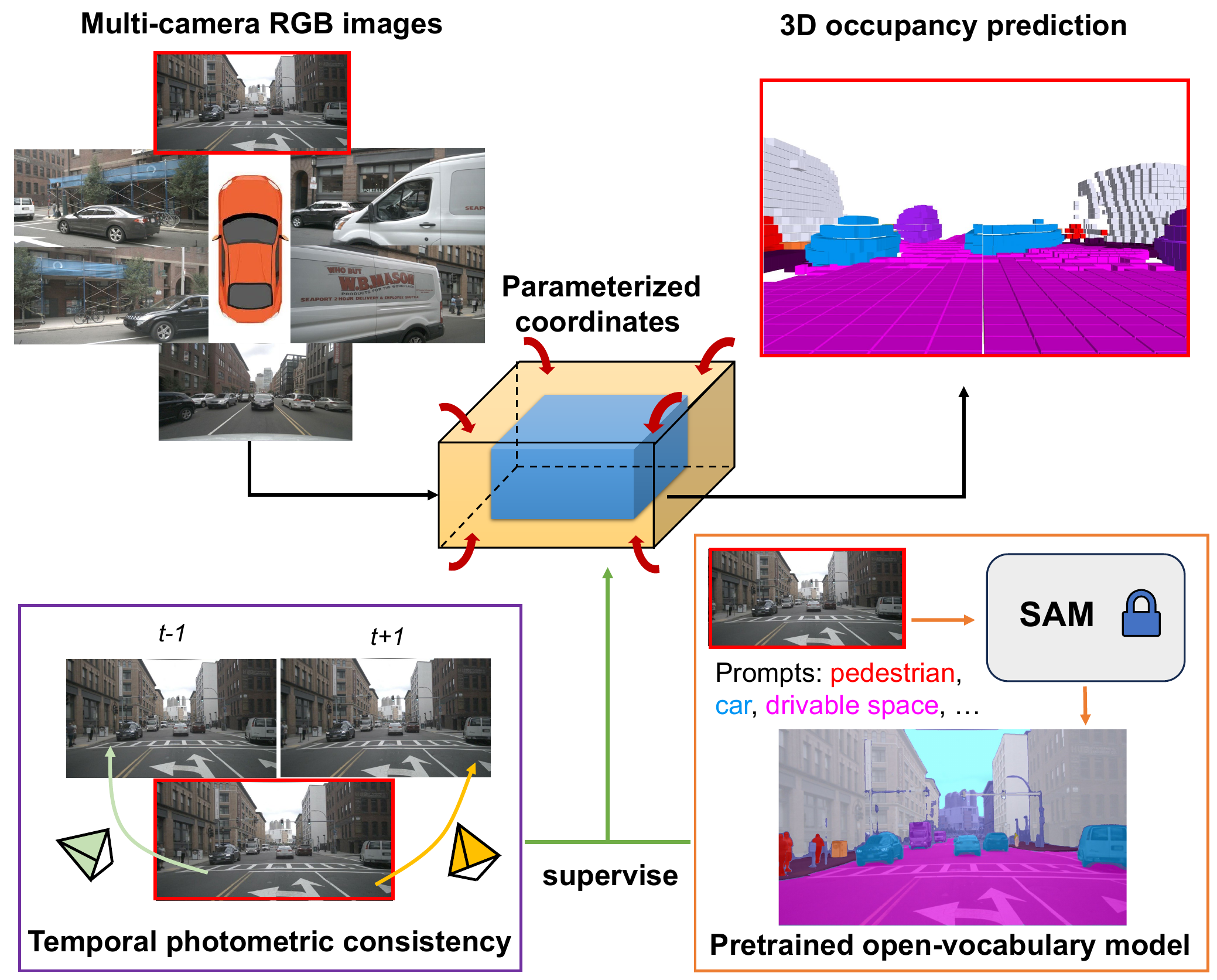}
      \end{center}
      %\vspace{-10pt}
      % \captionsetup{font={small}}
      \caption{\textbf{The overview of OccNeRF.} To represent unbounded scenes, we propose a parameterized coordinate to contract infinite space to bounded occupancy fields. Without using any LiDAR data or annotated labels, we leverage temporal photometric constraints and pretrained open-vocabulary segmentation models to provide geometric and semantic supervision.}
    %   The bottom part of the figure depicts the individual channels of SCER and the saturation of the dot color represents the relative intensity change corresponding to the central latent sharp frame.}
    % \vspace{-20pt}
  
    \label{fig:overview}
\end{figure}

To address this, we propose an OccNeRF method, which targets at training multi-camera occupancy networks without 3D supervision. The overview of our proposed method is shown in Fig.~\ref{fig:overview}. We first utilize a 2D backbone to extract multi-camera 2D features. To save memory, we directly interpolate 2D features to obtain 3D volume features instead of using heavy cross-view attention. In previous works, the volume features are supervised by the bounded occupancy labels (\eg 50m range) and they only need to predict the occupancy with finite resolution (\eg 200 $\times$ 200 $\times$ 16). Differently, for LiDAR-free training, we should consider unbounded scenes since the RGB images perceive an infinite range. To this end, we parameterize the occupancy fields to represent unbounded environments. Specifically, we split the whole 3D space into the inside and outside regions. The inside one maintains the original coordinate while the outside one adopts a contracted coordinate. A specific sampling strategy is designed to transfer parameterized occupancy fields to 2D depth maps with the neural rendering algorithm.

A straightforward way to supervise predicted occupancy is to calculate loss between rendered images and training images, which is the same as the loss function used in NeRF~\cite{nerf}. However, our experiments indicate this method's ineffectiveness due to the sparse nature of surrounding views, where minimal image overlap fails to supply sufficient geometric information. As an alternative, we take full advantage of temporal information by rendering multiple frames in a sequence and employing photometric consistency between adjacent frames as the primary supervision signal. For semantic occupancy, we propose three strategies to map the class names to the prompts, which are fed to a pretrained open-vocabulary segmentation model~\cite{sam,dino} to get 2D semantic labels. Then an additional semantic head is employed to render semantic images and supervised by these labels. To verify the effectiveness of our method, we conduct experiments on both self-supervised multi-camera depth estimation and 3D occupancy prediction tasks. Experimental results show that our OccNeRF outperforms other depth estimation methods by a large margin and achieves comparable performance with some methods using stronger supervision on the nuScenes~\cite{nuscenes} and SemanticKITTI~\cite{semantickitti} datasets.

% In summary, our key contributions are:
% \begin{itemize}
%     \item We design a system that trains an occupancy network without LiDAR data, addressing the challenge of sparse surrounding views through the incorporation of temporal information.
%     \item We introduce the parameterized occupancy field that allows vision-centric systems to efficiently represent unbounded scenes, matching the extensive perceptual range of cameras.
%     \item We devise three strategies to improve the quality of pseudo labels generated by pretrained open-vocabulary segmentation models.
% \end{itemize}

In summary, our principal contributions include:
\begin{itemize}
    \item We develop a system that trains an occupancy network without the need for LiDAR data, addressing the challenge of sparse surrounding views by integrating temporal information for more geometry information.
    \item We introduce a parameterized occupancy field that enables vision-centric systems to efficiently represent unbounded scenes, aligning with the extensive perceptual capabilities of the cameras.
    \item We devise a pipeline to generate high-quality pseudo labels with pretrained open-vocabulary segmentation models, with three prompt strategies to improve the accuracy.
\end{itemize}

\section{Related Work}
\label{sec:related}
This section examines three interrelated areas in computer vision: 3D occupancy prediction, neural radiance fields, and self-supervised depth estimation. We highlight key advancements and ongoing challenges, providing a critical overview that identifies gaps in current research and suggests avenues for further investigation.
%-------------------------------------------------------------------------
\subsection{3D Occupancy Prediction}
Due to the significance of the vision-centric autonomous driving systems, more and more researchers begin to focus on 3D occupancy prediction tasks~\cite{monoscene,tpvformer,surroundocc,occ3d,pointocc,occformer,openoccupancy,tong2023scene,scenerf,hayler2023s4c,voxformer,fbbev,fbocc}.
In the industry community, 3D occupancy is treated as an alternative to LiDAR perception. 
As one of the pioneering works, MonoScene~\cite{monoscene} extracts the voxel features generated by sight projection to reconstruct scenes from a single image. 
TPVFormer~\cite{tpvformer}  further extends it to multi-camera fashion with tri-perspective view representation. 
Beyond TPVFormer, SurroundOcc~\cite{surroundocc} designs a pipeline to generate dense occupancy labels instead of using sparse LiDAR points as the ground truth. 
In addition, a 2D-3D UNet with cross-view attention layers is proposed to predict dense occupancy. 
RenderOcc~\cite{renderocc} uses the 2D depth maps and semantic labels to train the model, reducing the dependence on expensive 3D occupancy annotations. 
Compared with these methods, our method does not need any annotated 3D or 2D labels. 
Occ3D~\cite{occ3d} establishes the occupancy benchmarks used in CVPR 2023 occupancy prediction challenge and proposes a coarse-to-fine occupancy network. 
SimpleOccupancy~\cite{simpleocc} presents a simple while effective framework for occupancy estimation. 
Although SimpleOccupancy~\cite{simpleocc} and SelfOcc~\cite{selfocc} investigate the vision-centric setting, they do not consider the infinite perception range of cameras.

%-------------------------------------------------------------------------
\subsection{Neural Radiance Fields}
As one of the most popular topics in 3D area, 
neural radiance fields (NeRF)~\cite{nerf}
have made great achievement in recent years. NeRF~\cite{nerf} learns the geometry of a scene by optimizing a continuous volumetric scene function with multi-view images. 
To obtain the novel views, volume rendering is performed to convert the radiance fields to RGB images. 
As a follow-up, mip-NeRF~\cite{mipnerf} represents the scene at a continuously valued and replaces rays as anti-aliased conical frustums. 
Beyond mip-NeRF, Zip-NeRF~\cite{zipnerf} integrates mip-NeRF with a grid-based model for faster training and better quality. 
There are several extensions of original NeRF, including dynamic scenes~\cite{park2021nerfies,gao2021dynamic,pumarola2021d,tretschk2021non,li2021neural,du2021neural,park2021hypernerf}, 
3D reconstruction~\cite{wang2021neus,yariv2021volume,yu2022monosdf,oechsle2021unisurf,wei2021nerfingmvs}, 
model accelerating~\cite{chen2022tensorf,garbin2021fastnerf,reiser2021kilonerf,fridovich2022plenoxels,muller2022instant,sun2022direct,zhang2023fast,chen2024learning}, etc. 
As one of these extensions, some works aim to describe unbounded scenes~\cite{nerf++,mipnerf360}. 
NeRF++~\cite{nerf++} split the 3D space as an inner unit sphere and an outer volume and proposes 
inverted sphere parameterization to represent outside regions. 
Further, mip-NeRF 360~\cite{mipnerf360} embeds this idea into mip-NeRF and 
applies the smooth parameterization to volumes. Inspired by these methods, 
we also design a parameterization scheme to model unbounded scenes for the occupancy prediction task.

\begin{figure*}[t]
    \centering
    \includegraphics[width=1.0\linewidth]{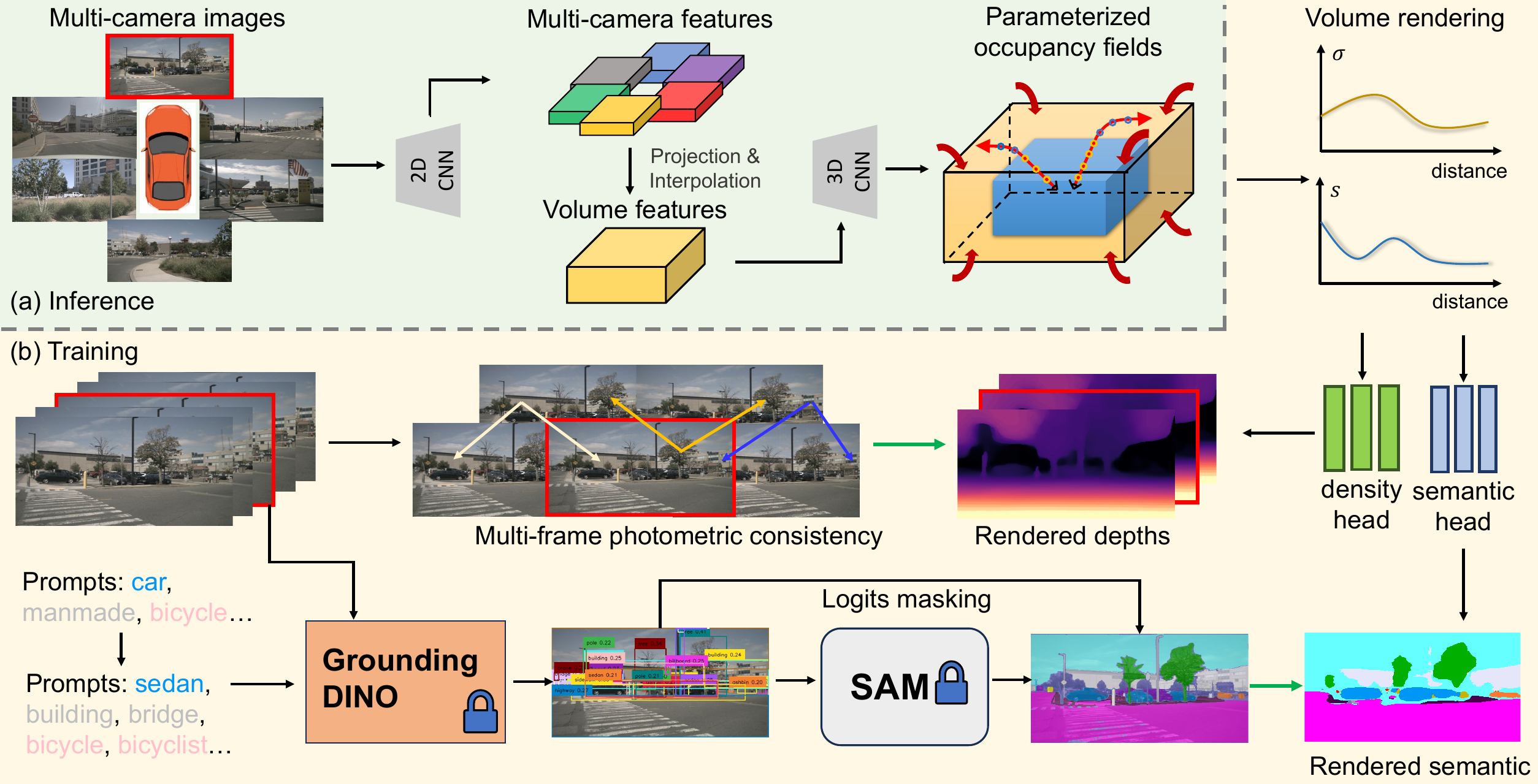}
    \caption{(a) During inference, a 2D backbone first extracts features from multiple cameras, which are then projected and interpolated into 3D space to create volume features. These are used to reconstruct parameterized occupancy fields that capture the extent of unbounded scenes.
    (b) During training, to generate rendered depth and semantic maps, we employ volume rendering using a redesigned sampling strategy. The depths from multiple frames are refined through the photometric loss. For the semantic prediction, we utilize pretrained Grounded-SAM model enhanced with prompt cleaning. The green arrow denotes the supervision signal.}
    \label{fig:pipeline}
    \vspace{-5mm}
\end{figure*}

%-------------------------------------------------------------------------
\subsection{Self-supervised Depth Estimation}
While early works~\cite{lee2019big,zhang2018progressive,fu2018deep,liu2015learning,roy2016monocular} require dense depth annotations, 
recent depth estimation methods~\cite{guizilini20203d,watson2019self,mahjourian2018unsupervised,yin2018geonet,wang2018learning,ranjan2019competitive,tosi2019learning,bian2019unsupervised,zhang2022self,xu2021multi,ye2021unsupervised,bello2024self,li2023sense} 
are designed in a self-supervised manner. 
Most of these methods predict depth maps and ego-motions simultaneously, 
adopting the photometric constraints~\cite{zhou2017unsupervised,godard2017unsupervised} between successive
frames as the supervision signal. As a classical work in this field, Monodepth2~\cite{monodepth2} proposes 
some techniques to improve the quality of depth predictions, including the minimum re-projection loss, 
full-resolution multi-scale sampling, and auto-masking loss. 
Since modern self-driving vehicles are usually equipped with multiple cameras to capture the surrounding views, 
researchers begin to concentrate on the multi-camera self-supervised depth estimation task~\cite{fsm,surrounddepth,r3d3,kim2022self,xu2022multi,shi2023ega}. 
FSM~\cite{fsm} is the first work to extend monocular depth estimation to full surrounding views by leveraging spatiotemporal contexts and pose consistency constraints. 
To predict the real-world scale, SurroundDepth~\cite{surrounddepth} uses structure-from-motion to generate scale-aware pseudo depths to pretrain the models. 
Further, it proposes the cross-view transformer and joint pose estimation to incorporate the multi-camera information. 
Recently, R3D3~\cite{r3d3} combines the feature correlation with bundle adjustment operators for robust depth and pose estimation. 
Different from these methods, our approach directly extracts features in 3D space, achieving multi-camera consistency and better reconstruction quality.  

\section{Method}
\label{sec:method}

Fig.~\ref{fig:pipeline} shows the pipeline of our approach. With the multi-camera images $\{{I}^{i}\}_{i=1}^{N}$ as inputs, we first utilize a 2D backbone to extract $N$ cameras' features  $\{{X}^{i}\}_{i=1}^{N}$. Then the 2D features are interpolated to the 3D space to obtain the volume features with known intrinsic $\{{K}^{i}\}_{i=1}^{N}$ and extrinsic $\{{T}^{i}\}_{i=1}^{N}$. As discussed in Section \ref{sec:parameterize}, to represent the unbounded scenes, we propose a coordinate parameterization to contract the infinite range to a limited occupancy field. The volume rendering is performed to convert occupancy fields to multi-frame depth maps, which are supervised by photometric loss. Section \ref{sec:depth} introduces this part in detail. Finally, Section \ref{sec:semantic} shows how we use a pretrained open-vocabulary segmentation model to get 2D semantic labels. 

\subsection{Parameterized Occupancy Fields}
\label{sec:parameterize}
Different from previous works~\cite{surroundocc,occformer}, we need to consider unbounded scenes in the LiDAR-free setting. On the one hand, we should preserve high resolution for the inside region (\eg [-40m, -40m, -1m, 40m, 40m, 5.4m]), since this part covers most regions of interest. On the other hand, the outside region is necessary but less informative and should be represented within a contracted space to reduce memory consumption. Inspired by~\cite{mipnerf360}, we propose a transformation function with adjustable regions of interest and contraction threshold to parameterize the coordinates $r=(x,y,z)$ of each voxel grid:

\begin{equation}
    f(r) = 
    \begin{cases}
        \alpha \cdot r' & \left| r' \right| \leq 1 \\
        \frac{r'}{\left|r'\right|} \cdot \left(1 - \frac{a}{\left|r'\right|+b}\right) & \left| r' \right| > 1
    \end{cases},
    \label{eq:cc_form}
\end{equation}
where $\alpha \in [0,1]$ represents the proportion of the region of interest in the parameterized space. Higher $\alpha$ indicates we use more space to describe the inside region. $r'=r/r_b$ denotes the normalized coordinate based on the input $r$ and pre-defined inside region bound $r_b$. The parameters $a$ and $b$ are introduced to maintain the continuity of the first derivative. The determination of these parameters is achieved through the resolution of the ensuing equations:
\begin{equation}
\begin{cases}
    \lim\limits_{r\rightarrow r_b^+} f(r) = \lim\limits_{r\rightarrow r_b^-} f(r)\\
    \lim\limits_{r\rightarrow r_b^+} f'(r) = \lim\limits_{r\rightarrow r_b^-} f'(r)
    \end{cases},
\label{eq:cc_deri}
\end{equation}
The derived solutions are presented as:
\begin{equation}
    \begin{cases}
        a = \frac{(1-\alpha)^2}{\alpha}\\
        b = \frac{1-2\alpha}{\alpha}\\
    \end{cases}.
    \label{eq:cc_deri2}
\end{equation}

\begin{figure}[t]
    \centering
    \includegraphics[width=1.0\linewidth]{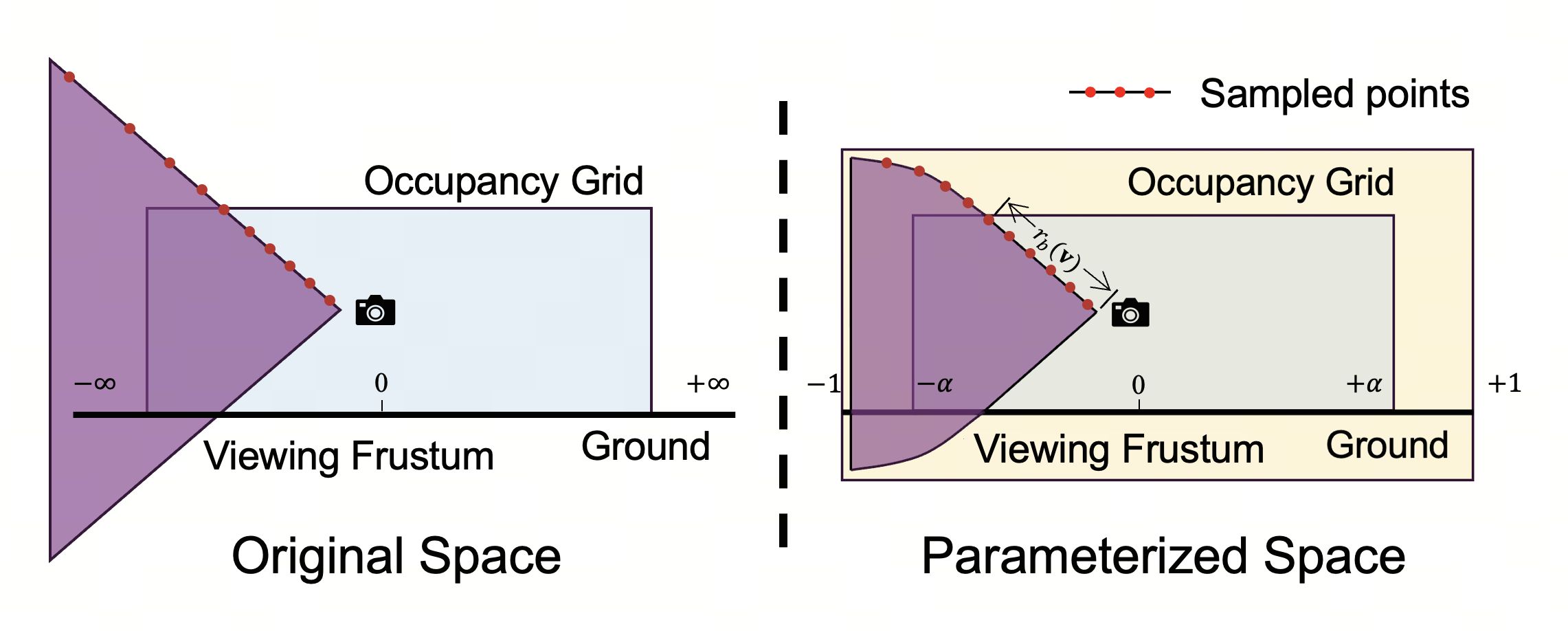}
    \caption{\textbf{Comparison between original space and parameterized space}. The original space utilizes the conventional Euclidean space, emphasizing linear mapping. The parameterized space is divided into two parts: an inner space with linear mapping to preserve high-resolution details and an outer space where point distribution is scaled inversely with distance. This design facilitates the representation of an infinite range within a finite spatial domain.}
    \label{fig:cc_explained}
\end{figure}

To obtain 3D voxel features from 2D views, we first generate the corresponding points $\mathcal{P}_{pc} = [\boldsymbol{x}_{pc}, \boldsymbol{y}_{pc}, \boldsymbol{z}_{pc}]^T$ for each voxel in the parameterized coordinate system and map them back to the ego coordinate system:
\begin{equation}
    \mathcal{P} = [f^{-1}_{x}(\boldsymbol{x}_{pc}), f^{-1}_{y}(\boldsymbol{y}_{pc}), f^{-1}_{z}(\boldsymbol{x}_{pc})]^T.
\end{equation}
Then we project these points to the 2D image feature planes and use bilinear interpolation to get the 2D features:
\begin{equation}
    \mathcal{F}^i = X^i \left< \text{proj} (\mathcal{P}, T^i, K^i) \right>.
\end{equation}
where $\text{proj}$ is the function projecting 3D points $\mathcal{P}$ to the 2D image plane defined by the camera extrinsic $T^i$ and intrinsic $K^i$, $\left<\right>$ is the bilinear interpolation operator, $\mathcal{F}^i$ is the interpolation result. To simplify the aggregation process and reduce computation costs, we directly average the multi-camera 2D features to get volume features, which is the same as the method used in ~\cite{simpleocc,simplebev}. Finally, a 3D convolution network~\cite{hybridnet} is employed to extract features and predict the final occupancy outputs.

\subsection{Multi-frame Depth Estimation}
\label{sec:depth}
To project the occupancy fields to multi-camera depth maps, we adopt volume rendering~\cite{max1995optical}, which is widely used in NeRF-based methods~\cite{nerf,nerf++,mipnerf}. To render the depth value of a given pixel, we cast a ray from the camera center $\boldsymbol{\text{o}}$ along the direction $\boldsymbol{\text{d}}$ pointing to the pixel. The ray is represented by $\boldsymbol{\text{v}}(t) = \boldsymbol{\text{o}} + t\boldsymbol{\text{d}}, t \in [t_n, t_f]$. Then, we sample $L$ points $\{t_k\}_{k=1}^{L}$ along the ray in 3D space to get the density $\sigma(t_k)$. For the selected $L$ quadrature points, the depth of the corresponding pixel is computed by:
\begin{equation}
    D(\boldsymbol{\text{v}}) = \sum_{k=1}^L T(t_k) (1 - \exp(-\sigma(t_k) \delta_k)) t_k,
    \label{eq:depth}
\end{equation}
where $T(t_k) = \exp\left({-\sum_{k'=1}^{k-1} \sigma(t_k)\delta_k}\right)$, and $\delta_k = t_{k+1} - t_{k}$ are intervals between sampled points.

A vital problem here is how to sample $\{t_k\}_{k=1}^{L}$ in our proposed coordinate system. Uniform sampling in the depth space or disparity space will result in an unbalanced series of points in either the outside or inside region of our parameterized grid, which is to the detriment of the optimization process. With the assumption that $o$ is around the coordinate system's origin, we directly sample $L(\boldsymbol{\text{r}})$ points from $U[0, 1]$ in parameterized coordinate and use the inverse function of Equation \ref{eq:cc_form} to calculate the $\{t_k\}_{k=1}^{L(\boldsymbol{\text{v}})}$ in the ego coordinates. The specific $L(\boldsymbol{\text{v}})$ and $r_b(\boldsymbol{\text{v}})$ for a ray are calculated by:
\begin{equation}
\begin{aligned}
    r_b(\boldsymbol{\text{v}}) =& \frac{\sqrt{(\boldsymbol{\text{d}} \cdot \boldsymbol{\text{i}} l_x)^2 + (\boldsymbol{\text{d}} \cdot \boldsymbol{\text{j}} l_y)^2 + (\boldsymbol{\text{d}} \cdot \boldsymbol{\text{k}} l_z)^2}}{2\Vert \boldsymbol{\text{d}} \Vert},\\
    & \quad  \quad  \quad L(\boldsymbol{\text{v}}) = \frac{2r_b(\boldsymbol{\text{v}})}{\alpha d_v}\\
\end{aligned}
\end{equation}
where $\boldsymbol{\text{i}}, \boldsymbol{\text{j}}, \boldsymbol{\text{k}}$ are the unit vectors in the x, y, z directions, $l_x, l_y, l_z$ are the lengths of the inside region, and $d_v$ is the voxel size. To better adapt to the occupancy representation, we directly predict the rendering weight instead of the density.

A conventional supervision method is to calculate the difference between rendered images and raw images, which is employed in NeRF~\cite{nerf}. However, our experimental results show that it does not work well. The possible reason is that the large-scale scene and few view supervision make it difficult for NeRF to converge. To better make use of temporal information, we employ the photometric loss proposed in~\cite{monodepth2,zhou2017unsupervised}. Specifically, we project adjacent frames to the current frames according to the rendered depths and given relative poses. Then we calculate the reconstruction error between projected images and raw images:
\begin{equation}
    \mathcal{L}_{pe}^i = \frac{\beta}{2}(1-\text{SSIM}(I^i, \hat{I}^i)) + (1 - \beta) \Vert  I^i, \hat{I}^i \Vert_1,
    \label{pm loss}
\end{equation}
where $\hat{I}^i$ is the projected image and $\beta=0.85$. Moreover, we adopt the techniques introduced in ~\cite{monodepth2}, \ie per-pixel minimum reprojection loss and auto-masking stationary pixels. For each camera view, we render a short sequence instead of a single frame and perform multi-frame photometric loss.

\subsection{Semantic Supervision}
\label{sec:semantic}

\begin{figure}[t] %{r}{0.68\textwidth}

    \centering
    \begin{subfigure}{0.48\linewidth}
        \includegraphics[width=\textwidth]{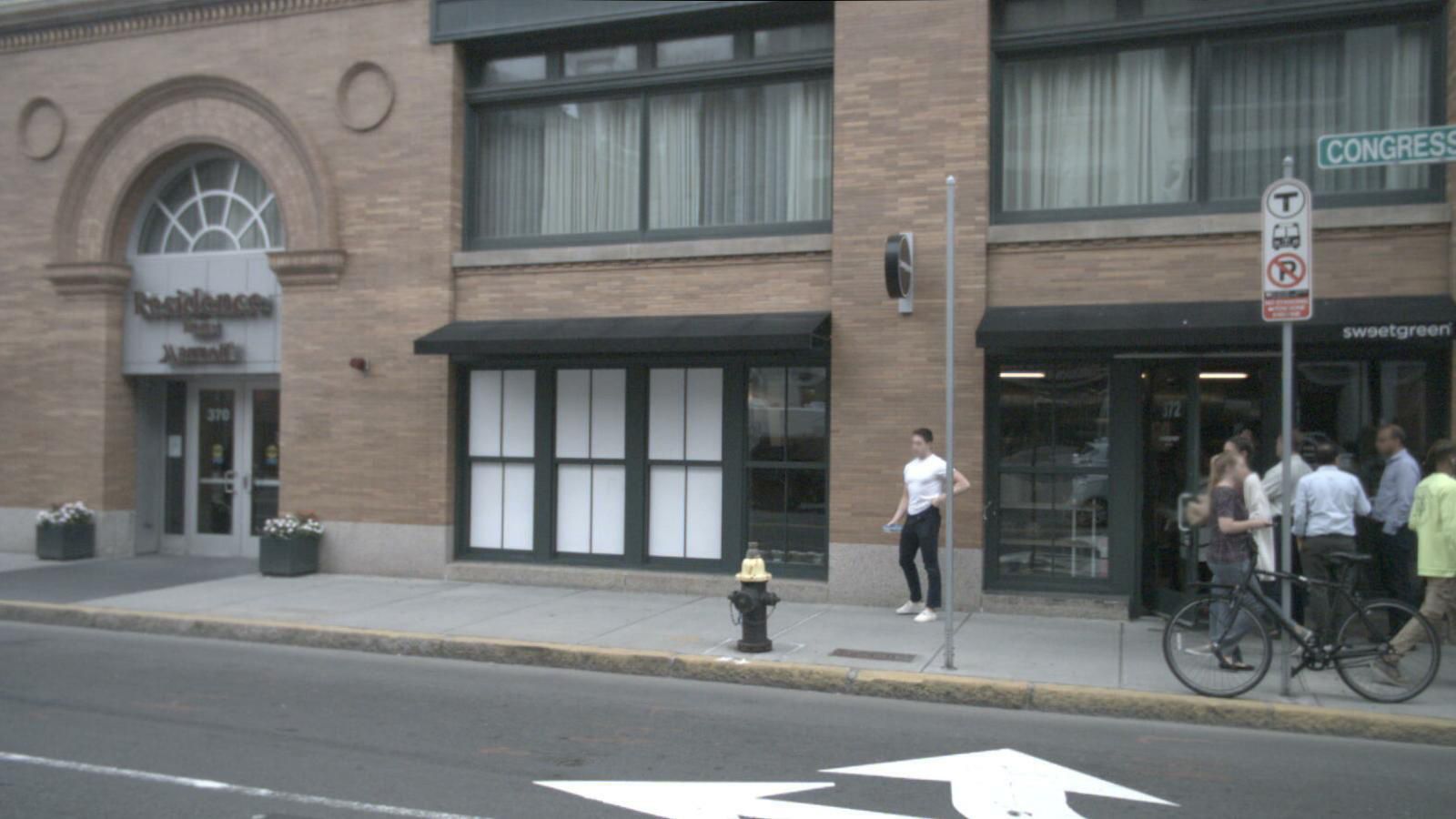}
        \caption{RGB image}
    \end{subfigure}
    \begin{subfigure}{0.48\linewidth}
        \includegraphics[width=\textwidth]{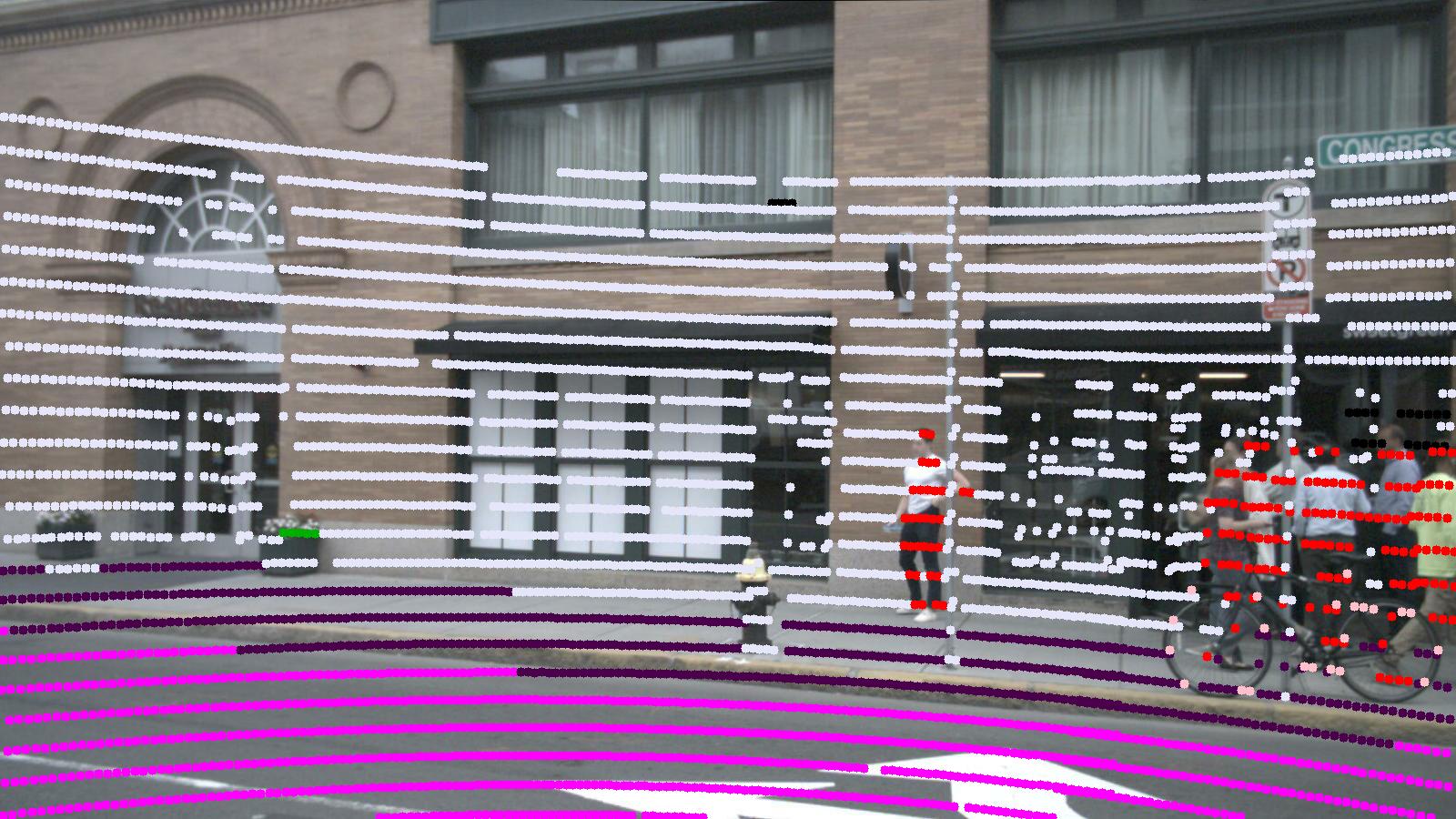}
        \caption{LiDAR projection labels}
    \end{subfigure} \\ %\relax
    \begin{subfigure}{0.48\linewidth}
        \includegraphics[width=\textwidth]{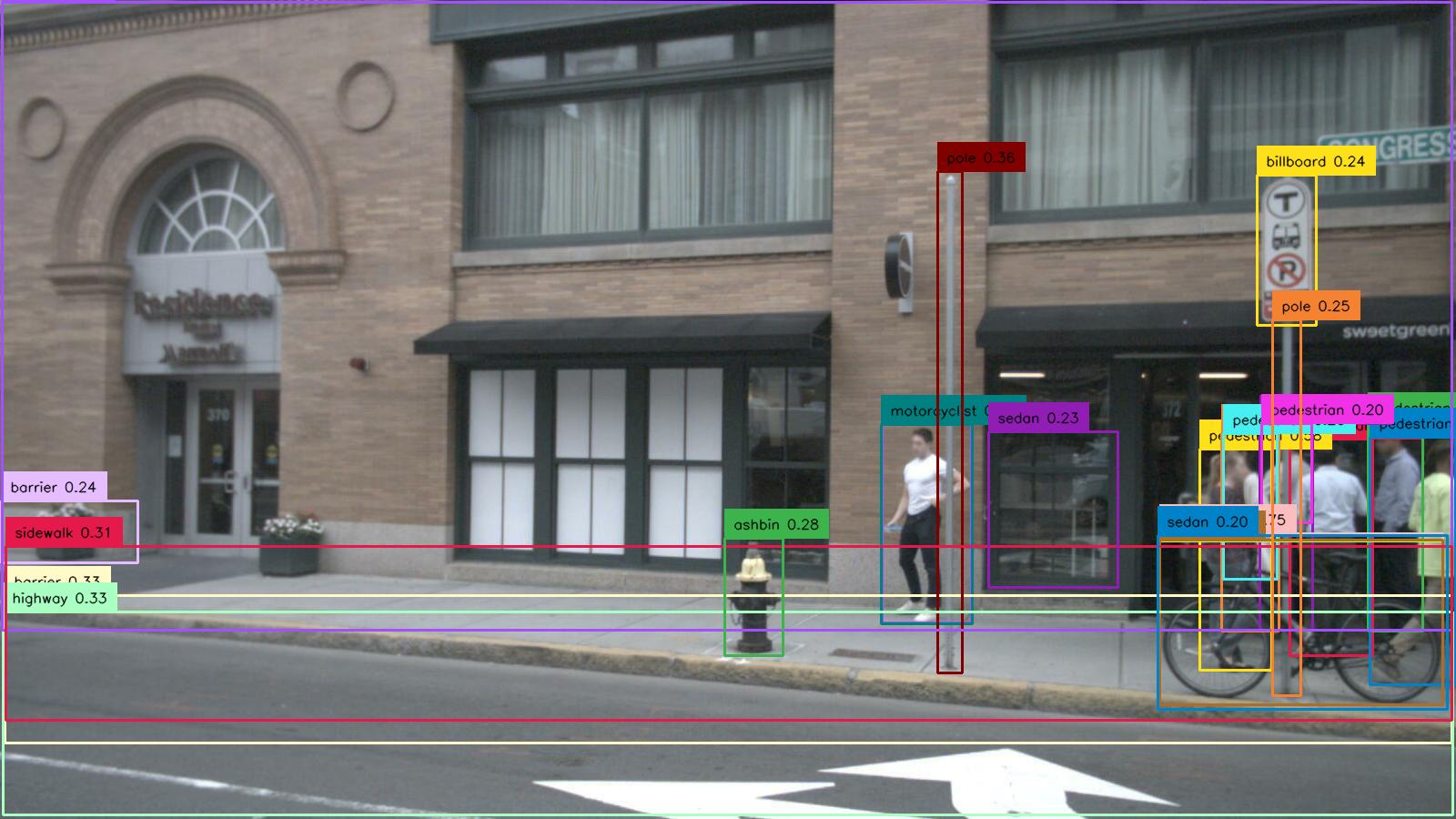}
        \caption{Our Grounding DINO bboxs}
    \end{subfigure}
    \begin{subfigure}{0.48\linewidth}
        \includegraphics[width=\textwidth]{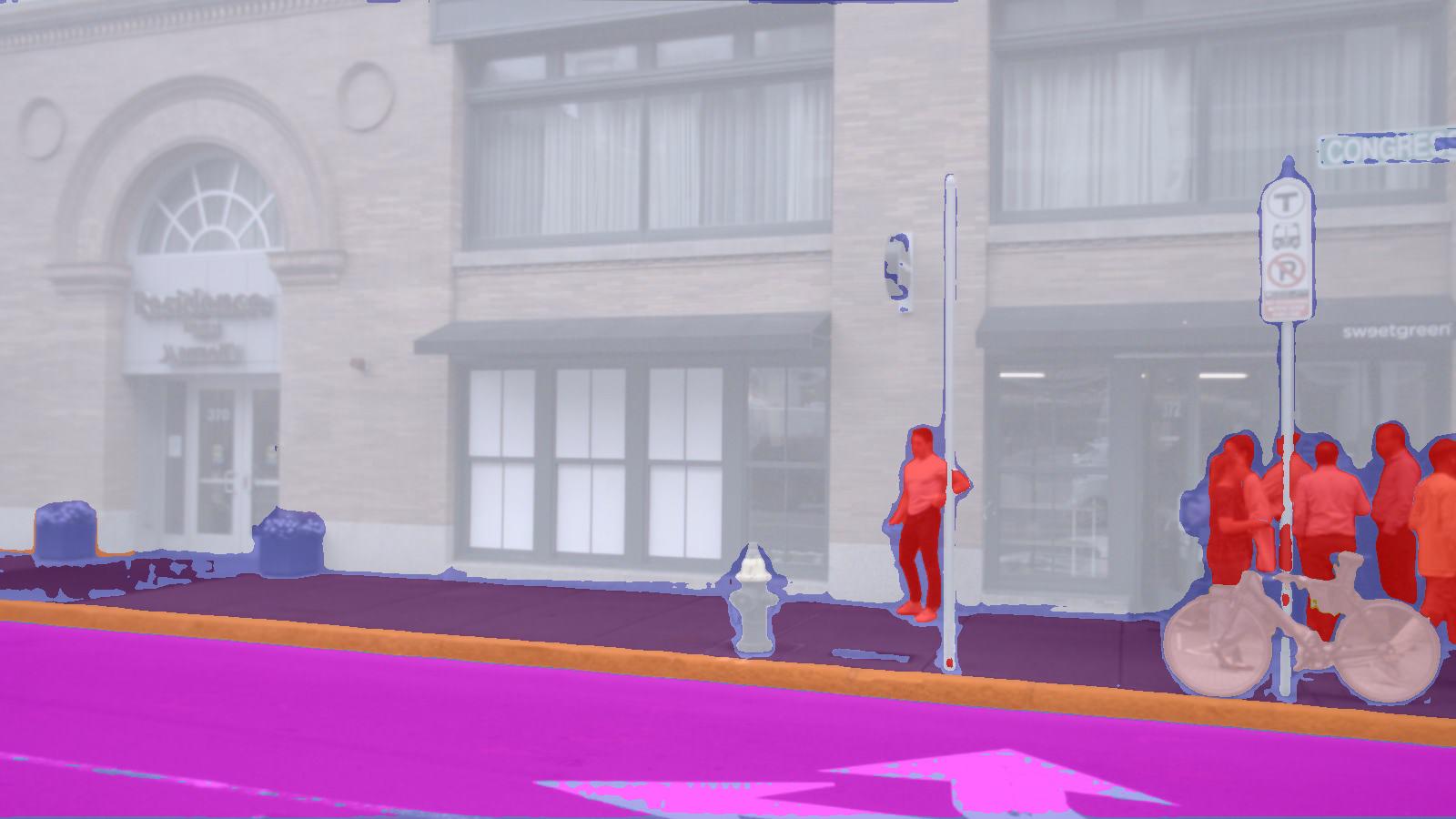}
        \caption{Our Grounded-SAM labels}
    \end{subfigure}
    
    \caption{\textbf{Label generation}. Detection bounding boxes generated by our Grounding DINO and semantic labels predicted by SAM in our method exhibit precision, which is comparable with that of LiDAR points projection labels.} 
    % \vspace{-15pt}
\label{fig:label_show_and_comparison}
\end{figure}

To enhance the richness of occupancy voxel information and facilitate comparison with existing methods, we introduce 2D labels to provide semantic supervision. Previous works~\cite{occ3d,sscbench} project 3D LiDAR points with segmentation labels to the image space to avoid the expensive cost of annotating dense 3D occupancy. However, we aim to predict semantic occupancy in a fully vision-centric system and use 2D data only. To this end, we leverage a pretrained open-vocabulary model Grounded-SAM~\cite{sam, dino, GroundedSAM} to generate 2D semantic segmentation labels. Without any 2D or 3D ground truth data, the pretrained open-vocabulary model enables us to obtain 2D labels which closely match the semantics of the given category names. This method can easily extend to any dataset, making our approach efficient and generalizable.

Specifically, when dealing with $\boldsymbol{c}$ categories, we employ three strategies to determine the prompts provided to the Grounding DINO. These strategies consist of synonymous substitution, where we replace words with their synonyms (e.g., changing `car' to `sedan' to enable the model to distinguish it from `truck' and `bus'); splitting single words into multiple entities (e.g., `manmade' is divided into `building', `billboard', and `bridge' etc. to enhance differentiation); and incorporating additional information (e.g., introducing `bicyclist' to facilitate the detection of a person on a bike). See TABLE~\ref{tab:semantic_gen} for more details. Subsequently, we obtain detection bounding boxes along with their corresponding logits and phrases, which are fed to SAM~\cite{sam} to generate $M$ precise segmentation binary masks. After multiplying the Grounding DINO logits with binary masks, every pixel has $\{l_i\}_{i=1}^{M}$ logits. We get the per-pixel label $\mathcal{S}^{pix}$ using:
\begin{equation}
    \mathcal{S}^{pix} = \psi (\mathop{\arg\max}_{i} l_i),
\end{equation} 
where  $\psi(\cdot)$ is a function that maps the index of $l_i$ to the category label according to the phrases. If a pixel does not belong to any categories and gets $M$ zero logits, we will give it an `uncertain' label. The generated detection bounding boxes and semantic labels are shown in Fig.~\ref{fig:label_show_and_comparison}.

To leverage the 2D semantic supervision, we initially utilize a semantic head with $\boldsymbol{c}$ output channels to map volume features extracted to semantic outputs, denoted as $S(x)$. Similar to the method outlined in Section \ref{sec:depth}, we engage in volume rendering once more using the subsequent equation:
\begin{equation}
\hat{\mathcal{S}}^{pix}(\boldsymbol{\text{r}}) = \sum_{k=1}^{L_s} T(t_k) (1 - \exp(-\sigma(t_k) \delta_k)) S(t_k),
\end{equation}
where $\hat{\mathcal{S}}^{pix}$ represents the per-pixel semantic rendering output. To save the memory and improve efficiency, we do not render the pixels that are assigned with ‘uncertain’ labels. Moreover, we only render the central frame instead of multiple frames and reduce the sample ratio to $L_s=L/4$. Our overall loss function is expressed as:
\begin{equation}
\mathcal{L}_{total} = \sum_{i} \mathcal{L}_{pe}^i + \lambda\mathcal{L}_{sem}^i (\hat{\mathcal{S}}^{pix}, \mathcal{S}^{pix})
\end{equation}
where $\mathcal{L}_{sem}$ is the cross-entropy loss function and $\lambda$ is the semantic loss weight. %It is worth mentioning that $\mathcal{L}_{sem}$ does not compute the pixel where $\mathcal{S}^{pix}$ is an `uncertain' label.

\begin{table}[t]
	\caption{\textbf{Details of prompt strategy.}}
	\begin{center}
	\resizebox{0.4\textwidth}{!}{
		\begin{tabular}{|c|c|}
			\hline
			Original labels & Ours \\
			\hline

			car & sedan \\
            bicycle & bicycle bicyclist\\
            vegetation &  tree \\
            motorcycle  & motorcycle motorcyclist \\
            drivable surface & highway \\
            traffic cone & cone \\
            construction vehicle & crane \\
            \multirow{2}{*}{manmade} & building compound bridge \\ 
            & pole billboard light ashbin \\
			\hline
	    \end{tabular}}
	\end{center}
	\label{tab:semantic_gen}
\end{table}

\definecolor{col1}{RGB}{232, 161, 148}
\definecolor{col2}{RGB}{148, 187, 232}

\begin{table*}[t]
    \caption{\textbf{Comparisons for self-supervised multi-camera depth estimation on the nuScenes dataset~\cite{nuscenes}}. The results are averaged over all views without median scaling at test time. `FSM*' is the reproduced result in~\cite{kim2022self} .}
	\centering
	\resizebox{1.0\textwidth}{!}{
	\begin{tabular}{|l|c|c|c|c|c|c|c|}
			\hline
			Method &\cellcolor{col1}Abs Rel & \cellcolor{col1}Sq Rel & \cellcolor{col1}RMSE  & \cellcolor{col1}RMSE log & \cellcolor{col2}$\delta < 1.25 $ & \cellcolor{col2}$\delta < 1.25^{2}$ & \cellcolor{col2}$\delta < 1.25^{3}$\\
			\hline
			FSM~\cite{fsm} &  0.297  &   -  &   -  &   -  &   -  &   -  &   -
\\

			FSM*~\cite{fsm}&   0.319 &7.534 &7.860 &0.362 &0.716 &0.874 &0.931

 \\
			SurroundDepth~\cite{surrounddepth} &   0.280 &4.401 &7.467 &0.364 &0.661 &0.844& 0.917

\\
            Kim \etal~\cite{kim2022self} &    0.289 &5.718 &7.551 &0.348 &0.709 &0.876 &0.932

\\
            R3D3~\cite{r3d3} &    0.253 &4.759 &7.150 &- &0.729 &- &-
\\
            SimpleOcc~\cite{simpleocc} &    0.224 &3.383 & 7.165 & 0.333 &  0.753 &0.877 &0.930
\\
			  OccNeRF & \bf{0.202}  &   \bf{2.883}  &  \bf{6.697}  &   \bf{0.319}  &   \bf{0.768}  &   \bf{0.882}  &   \bf{0.931}

  \\
			\hline
	\end{tabular}}
	
	\label{tab:depth}
\end{table*}

\begin{figure*}[t]
    \centering
    \vspace{-5pt}
    %\scriptsize
    \setlength\tabcolsep{1.0pt} % default value: 6pt
    \renewcommand{\arraystretch}{1.0}
    \begin{tabular}{cccccc}
    {\includegraphics[width=0.165\linewidth]{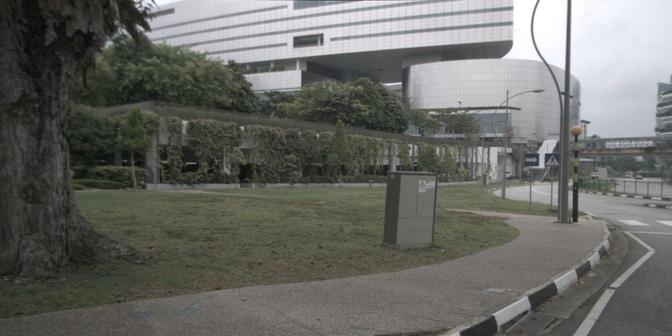}} &
    {\includegraphics[width=0.165\linewidth]{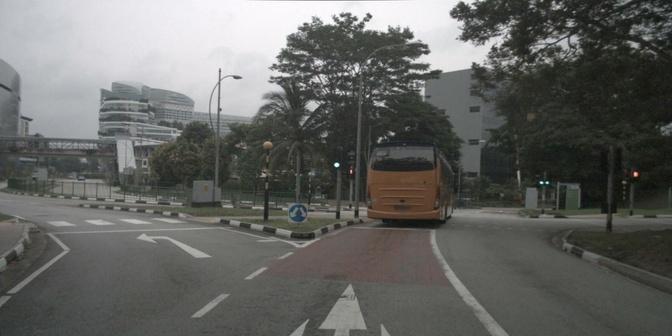}} &
    {\includegraphics[width=0.165\linewidth]{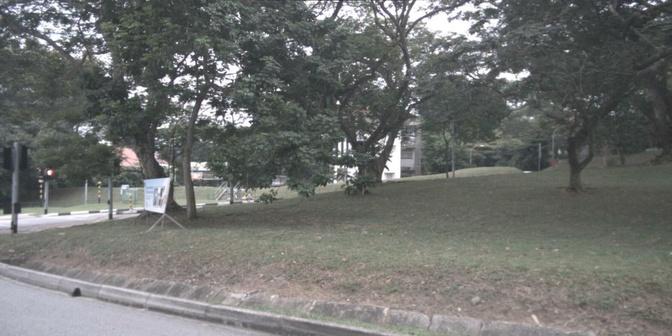}} &
    {\includegraphics[width=0.165\linewidth]{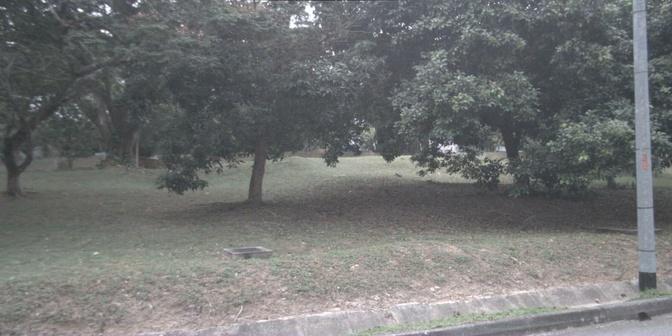}} &
    {\includegraphics[width=0.165\linewidth]{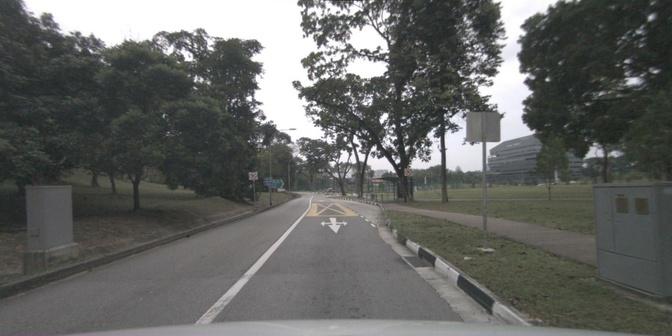}} &
    {\includegraphics[width=0.165\linewidth]{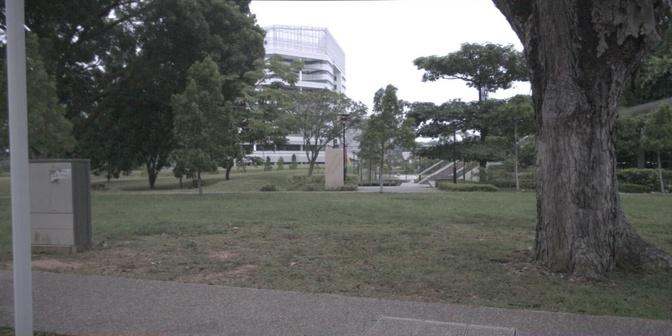}} \\
    
    {\includegraphics[width=0.165\linewidth]{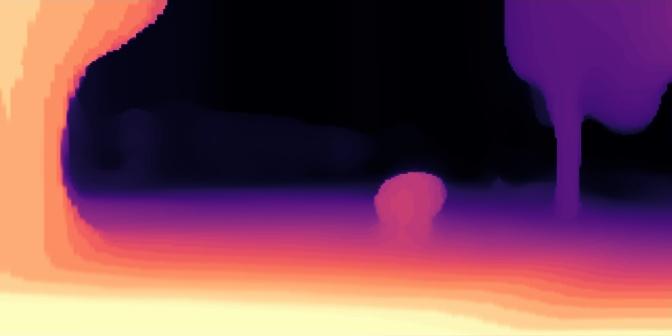}} &
    {\includegraphics[width=0.165\linewidth]{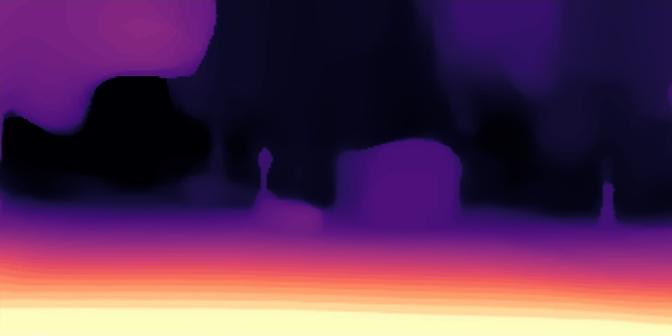}} &
    {\includegraphics[width=0.165\linewidth]{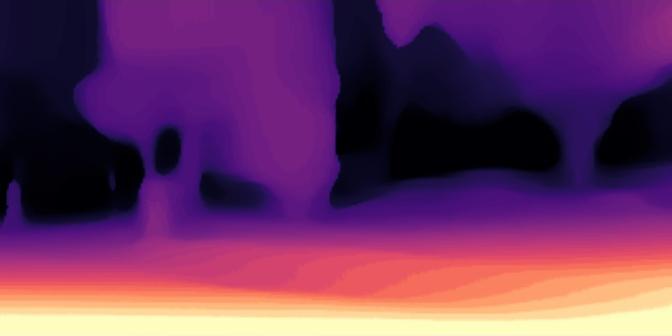}} &
    {\includegraphics[width=0.165\linewidth]{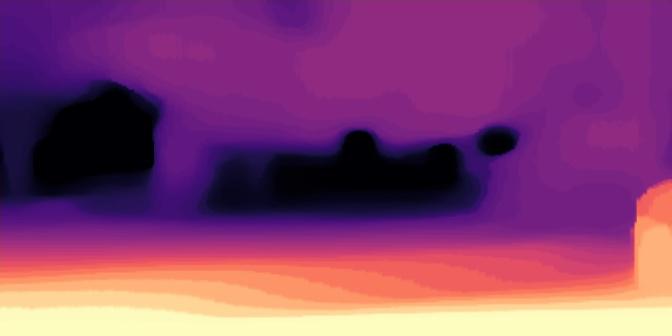}} &
    {\includegraphics[width=0.165\linewidth]{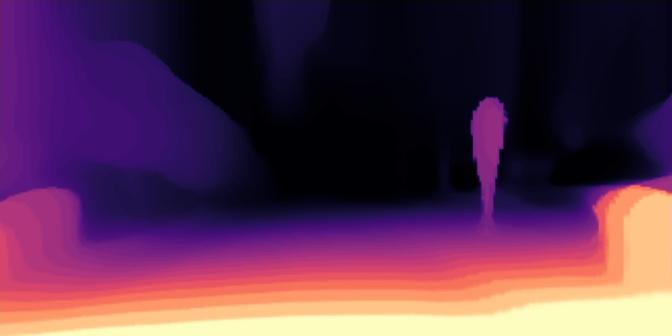}} &
    {\includegraphics[width=0.165\linewidth]{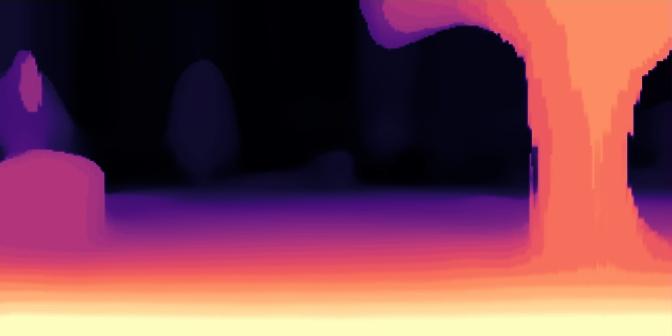}} \\
    
    {\includegraphics[width=0.165\linewidth]{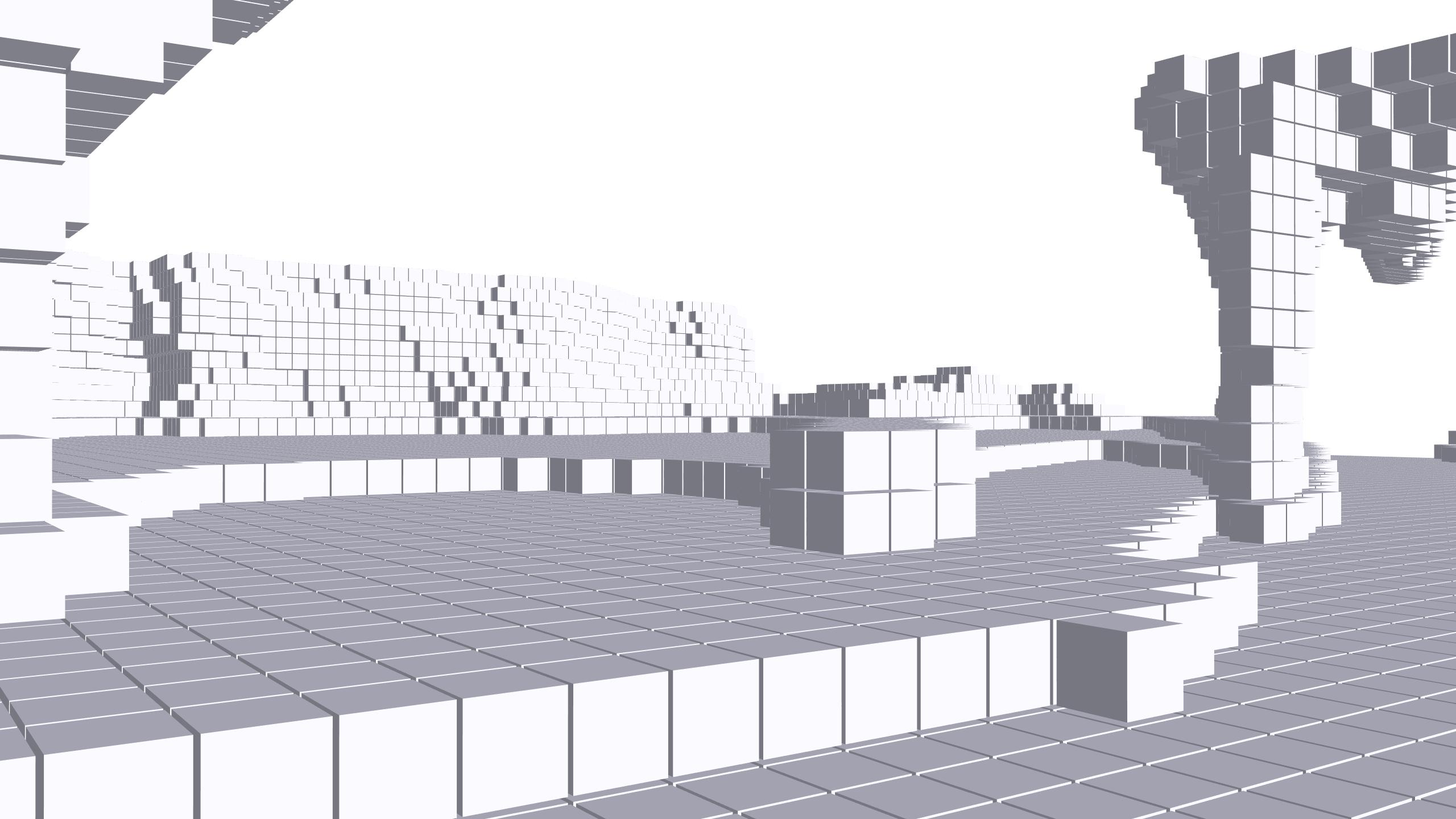}} &
    {\includegraphics[width=0.165\linewidth]{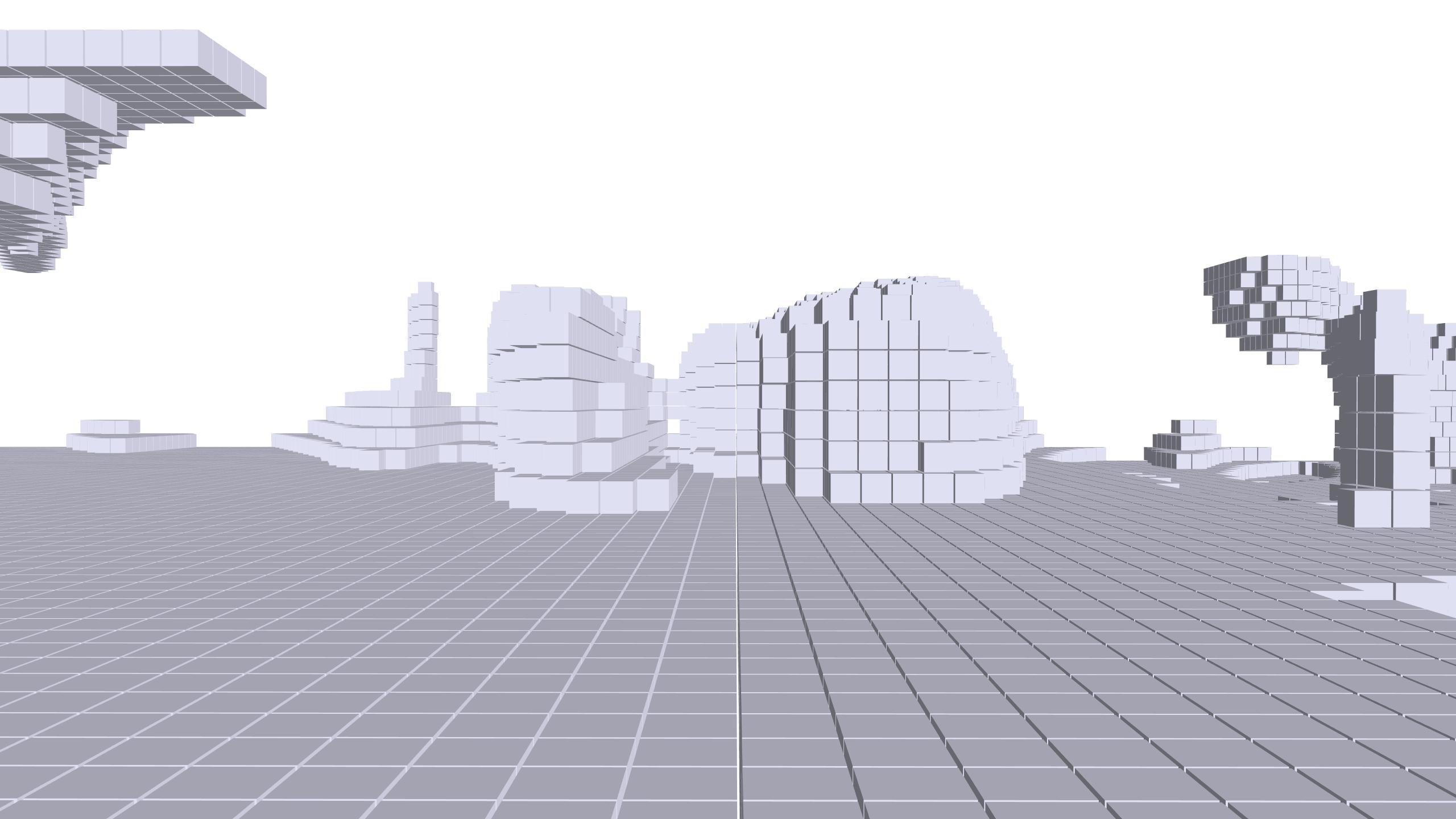}} &
    {\includegraphics[width=0.165\linewidth]{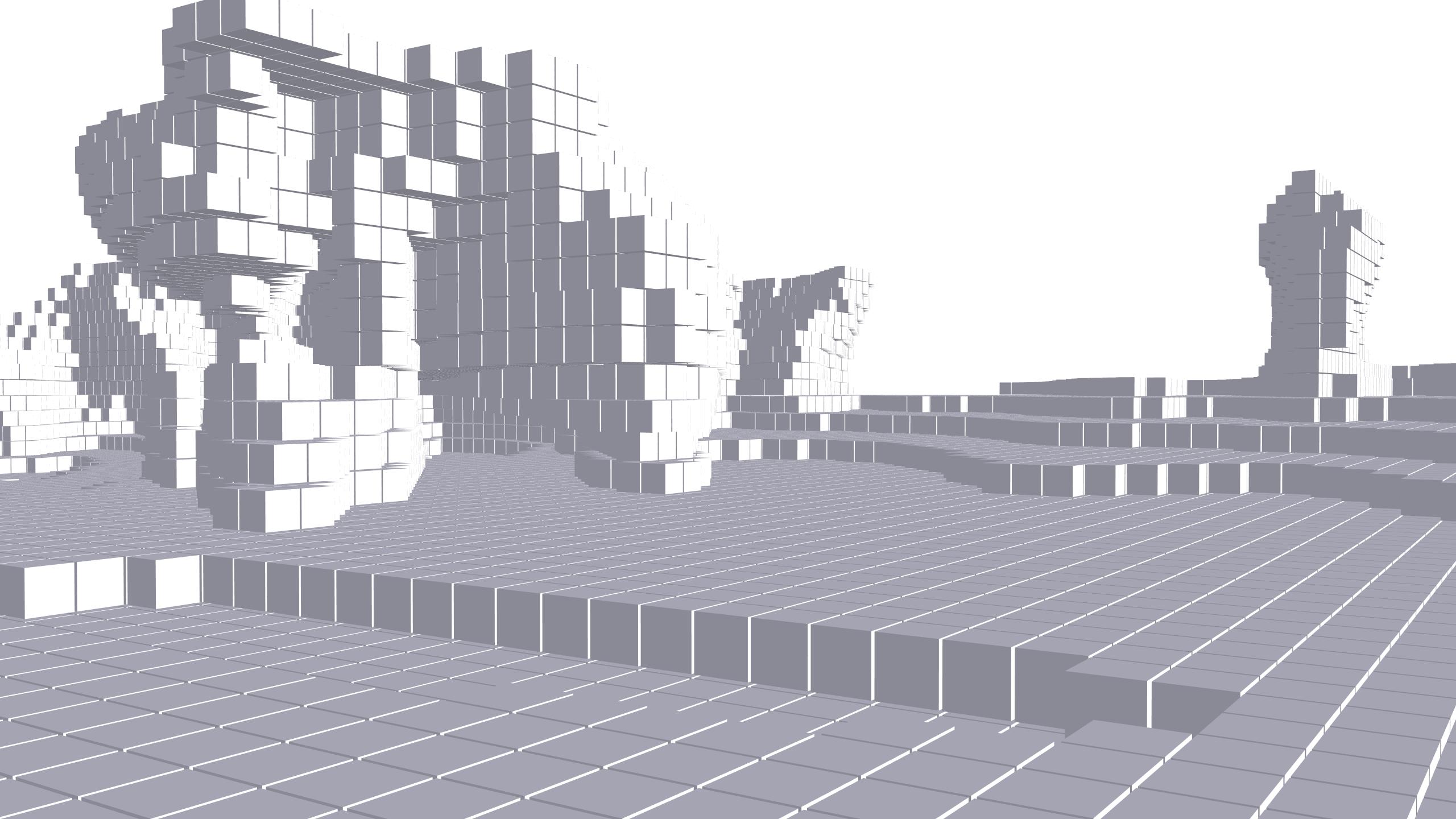}} &
    {\includegraphics[width=0.165\linewidth]{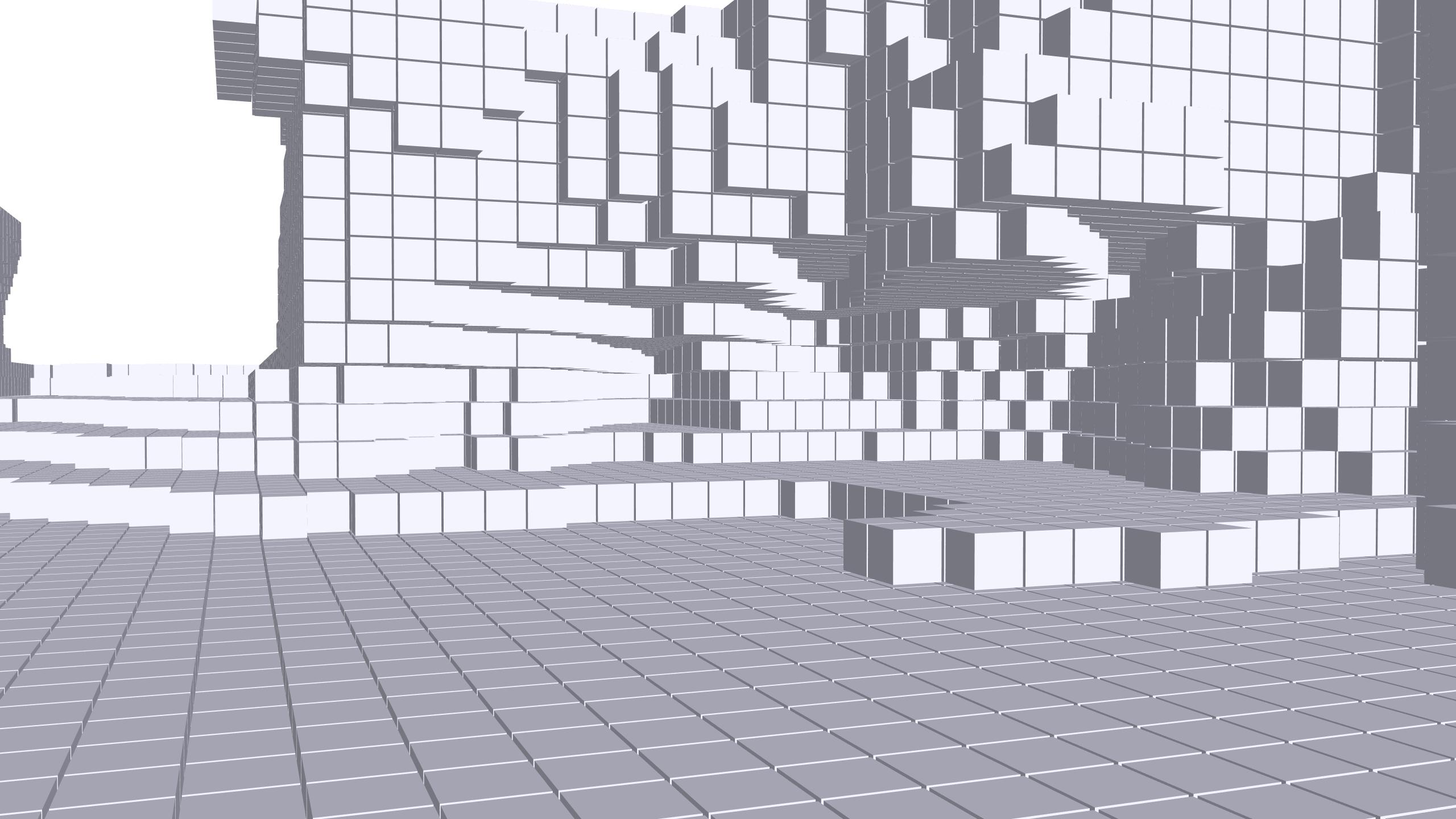}} &
    {\includegraphics[width=0.165\linewidth]{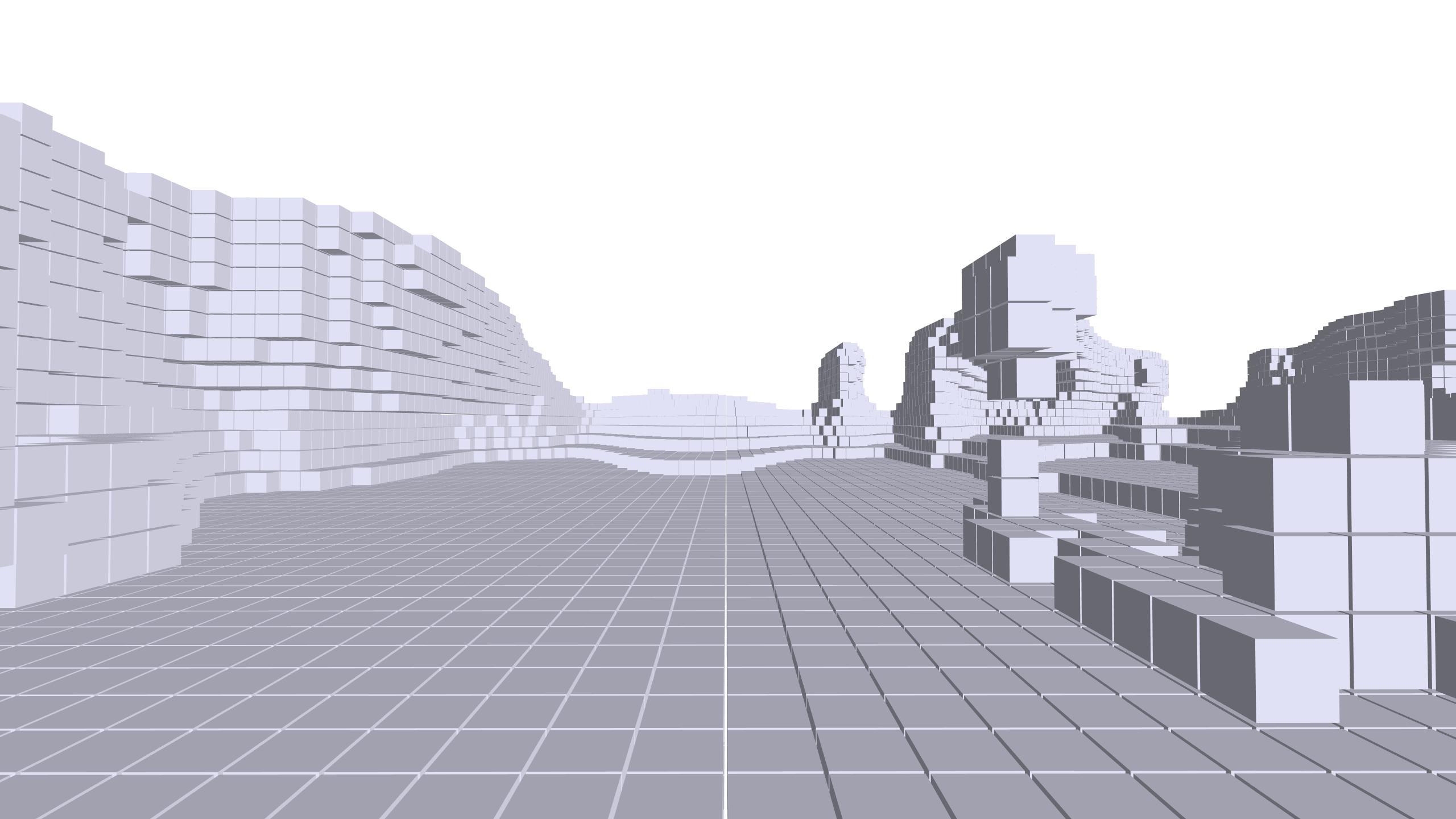}} &
    {\includegraphics[width=0.165\linewidth]{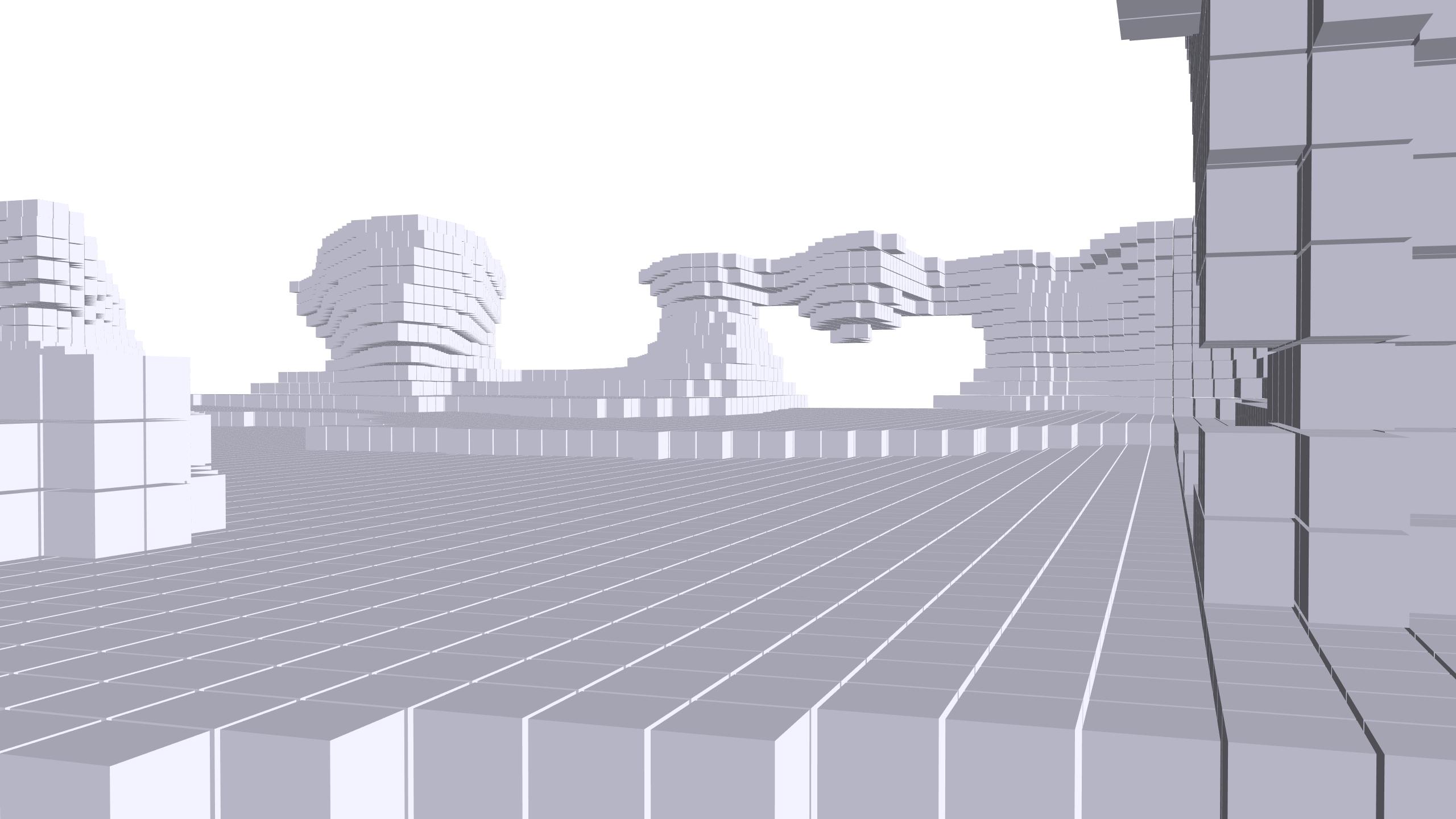}} \\

    {\includegraphics[width=0.165\linewidth]{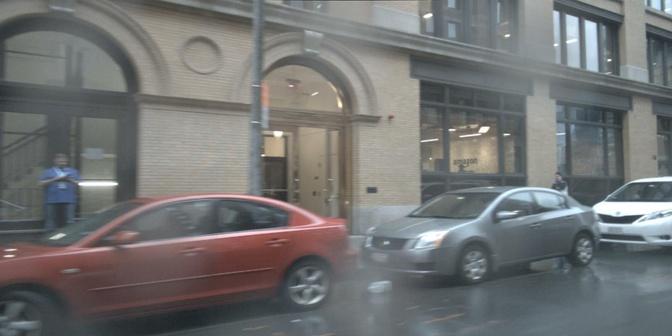}} &
    {\includegraphics[width=0.165\linewidth]{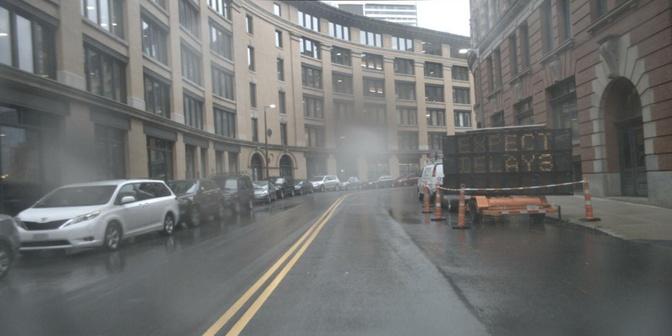}} &
    {\includegraphics[width=0.165\linewidth]{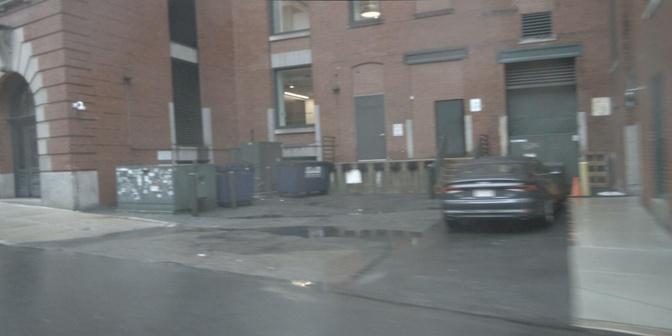}} &
    {\includegraphics[width=0.165\linewidth]{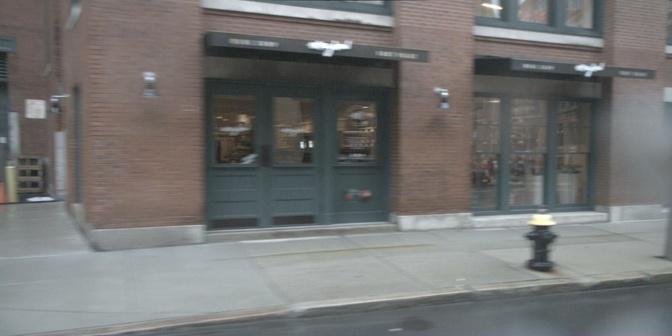}} &
    {\includegraphics[width=0.165\linewidth]{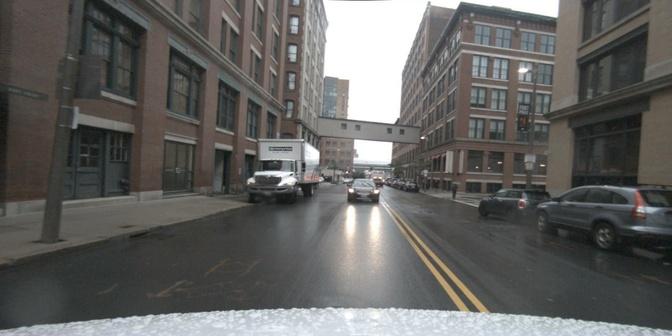}} &
    {\includegraphics[width=0.165\linewidth]{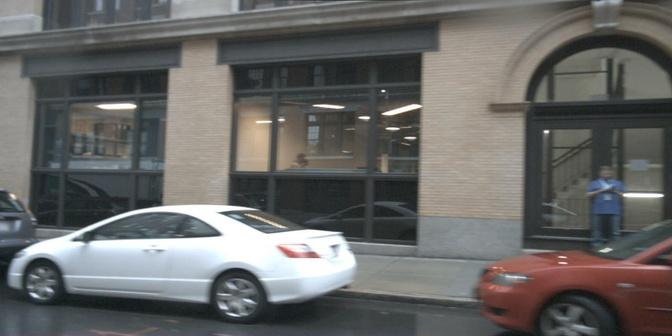}} \\
    
    {\includegraphics[width=0.165\linewidth]{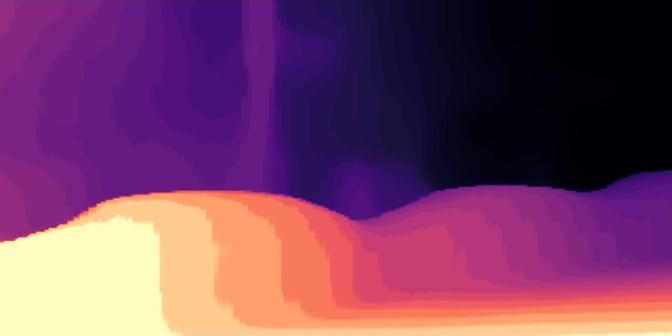}} &
    {\includegraphics[width=0.165\linewidth]{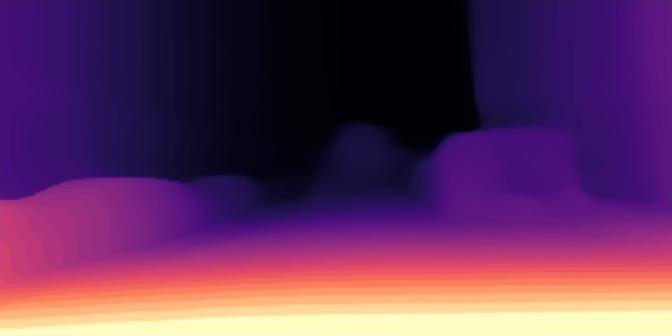}} &
    {\includegraphics[width=0.165\linewidth]{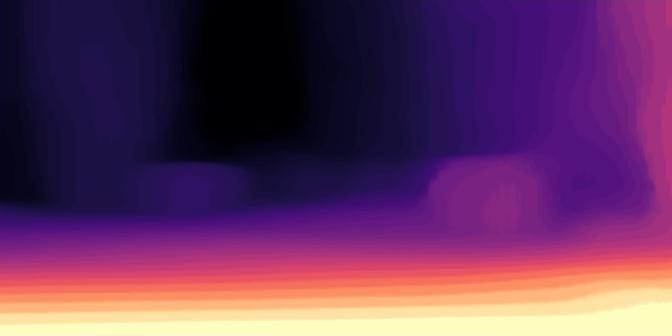}} &
    {\includegraphics[width=0.165\linewidth]{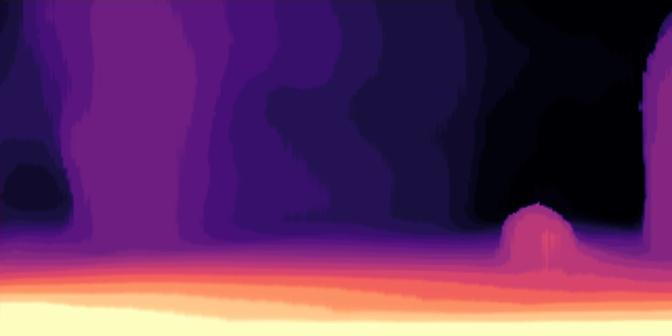}} &
    {\includegraphics[width=0.165\linewidth]{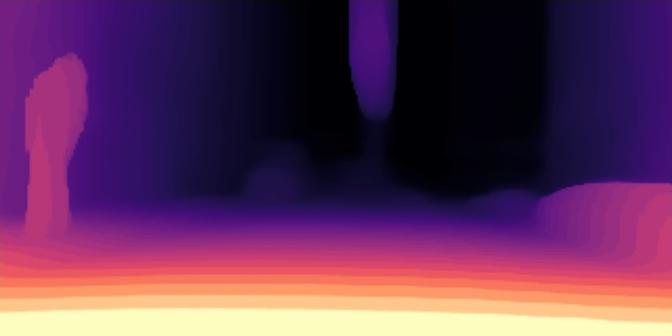}} &
    {\includegraphics[width=0.165\linewidth]{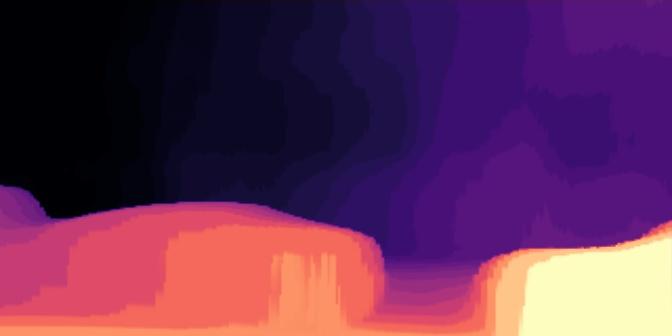}} \\
    
    {\includegraphics[width=0.165\linewidth]{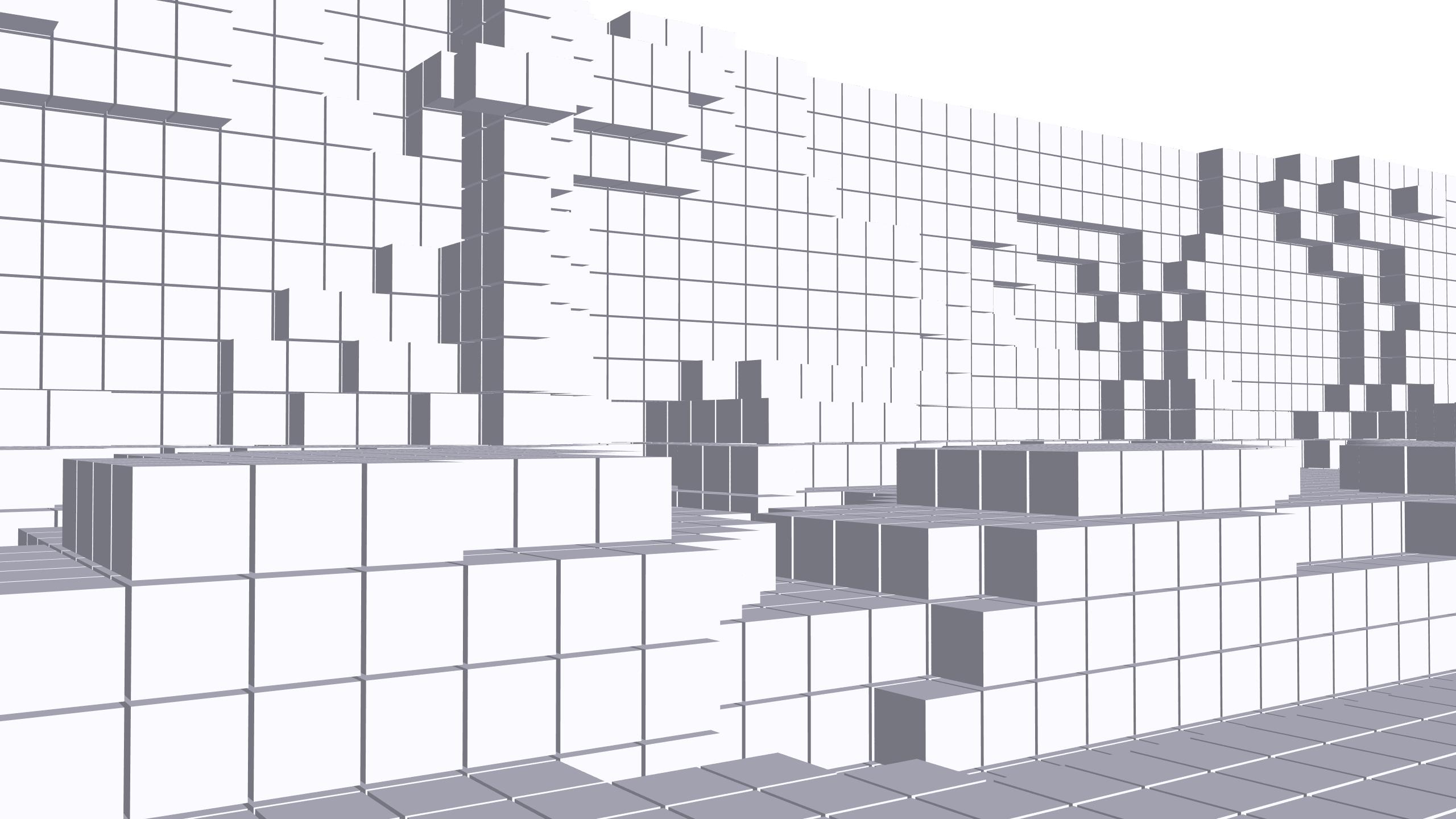}} &
    {\includegraphics[width=0.165\linewidth]{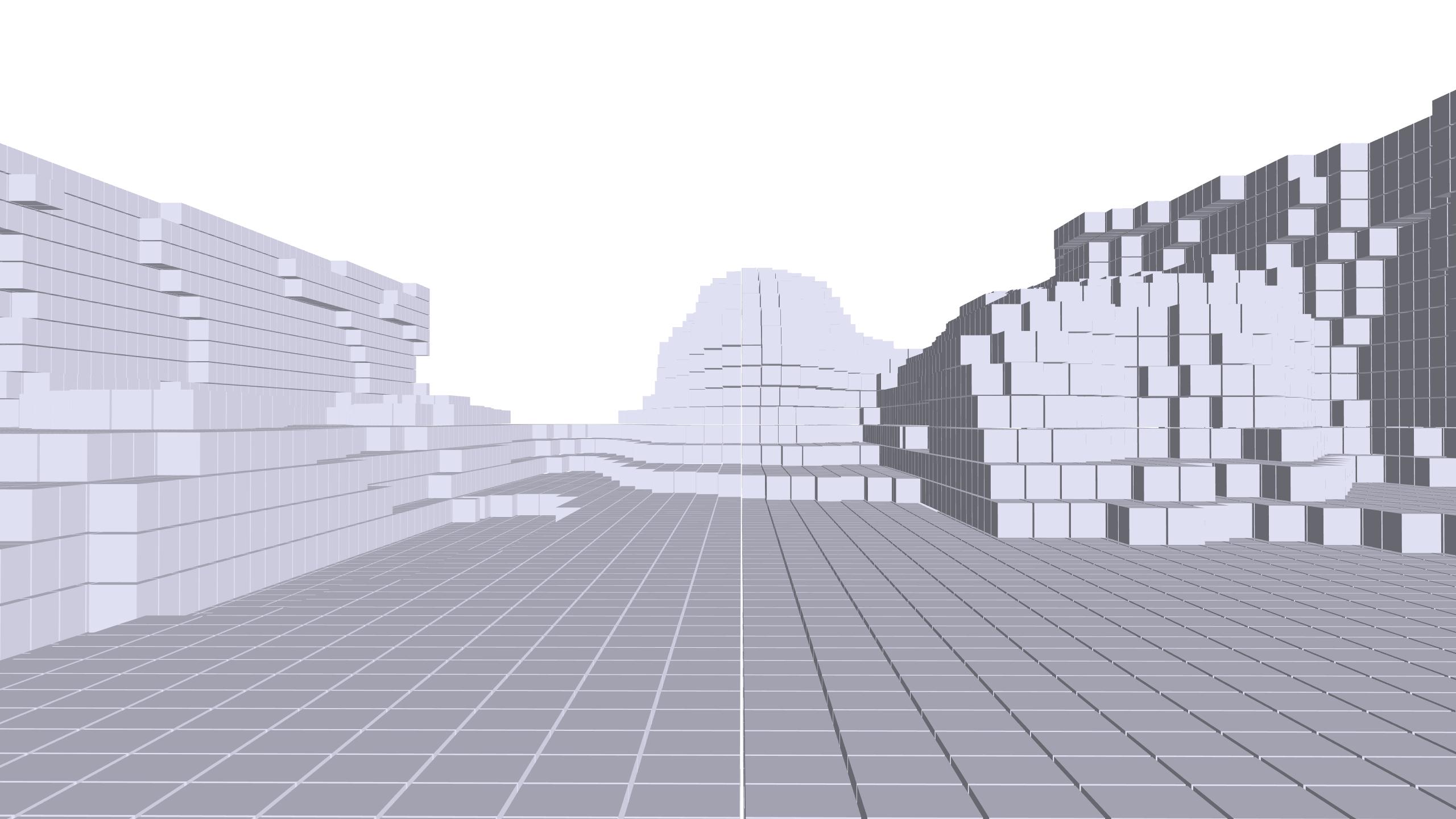}} &
    {\includegraphics[width=0.165\linewidth]{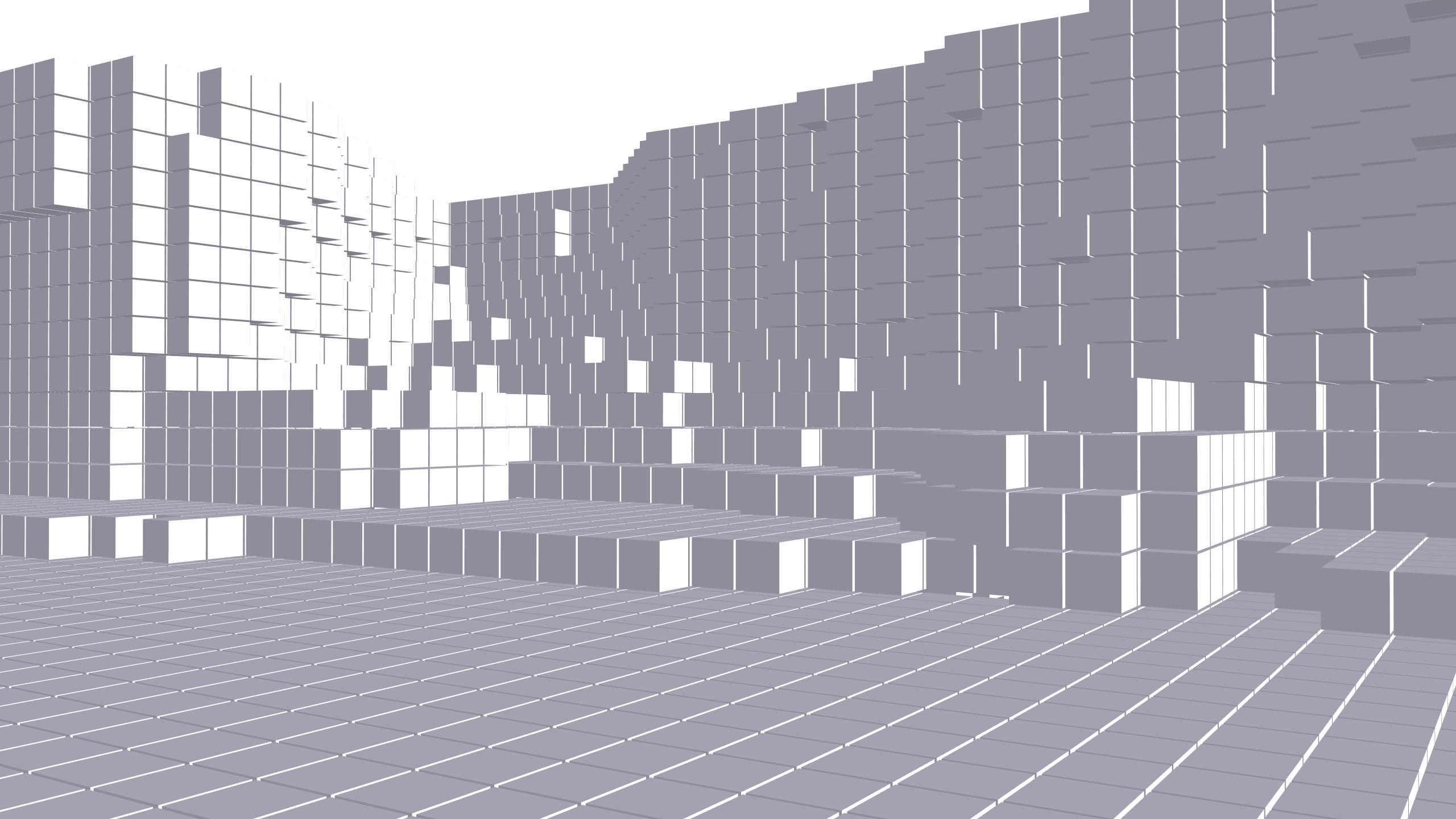}} &
    {\includegraphics[width=0.165\linewidth]{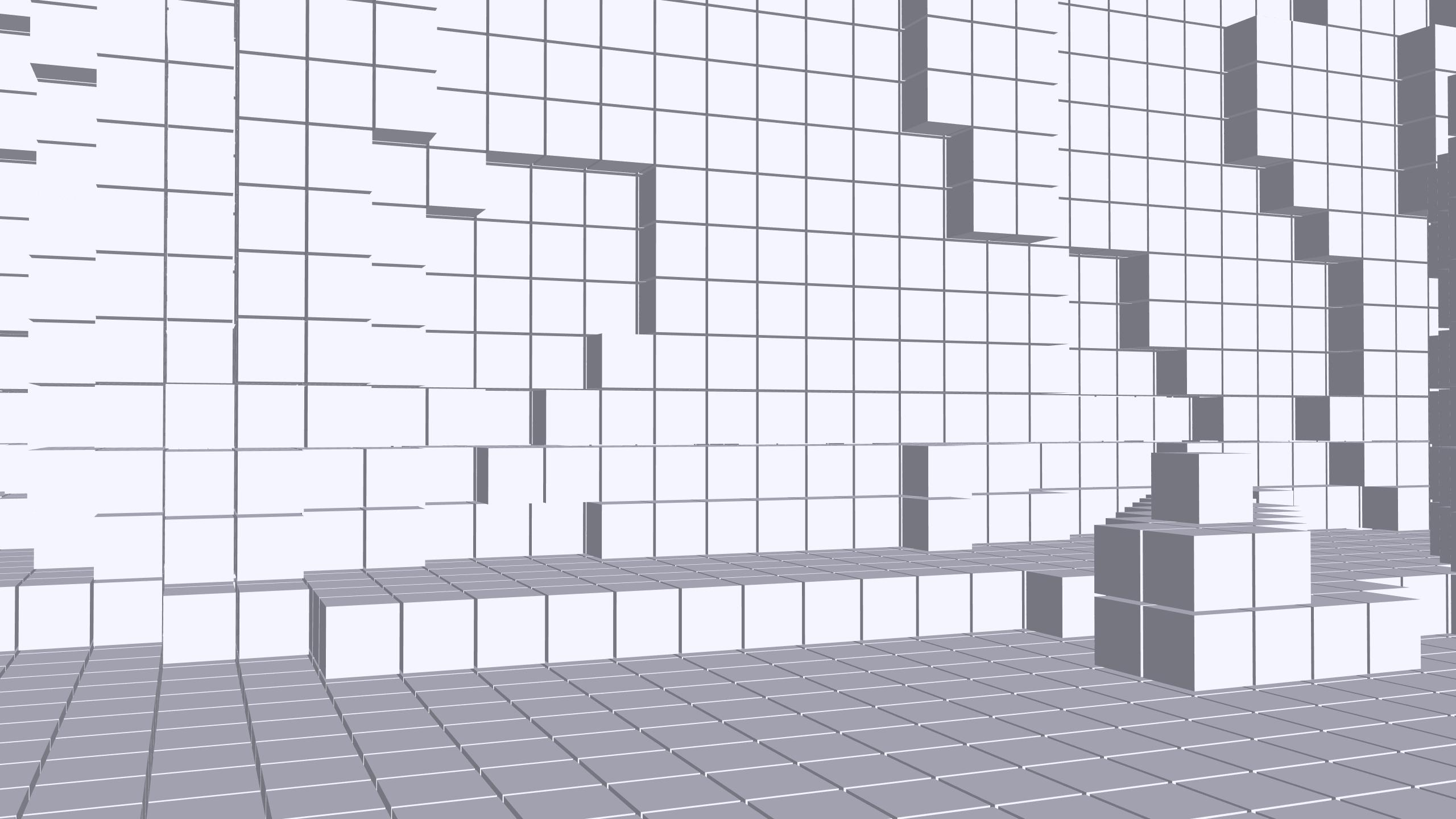}} &
    {\includegraphics[width=0.165\linewidth]{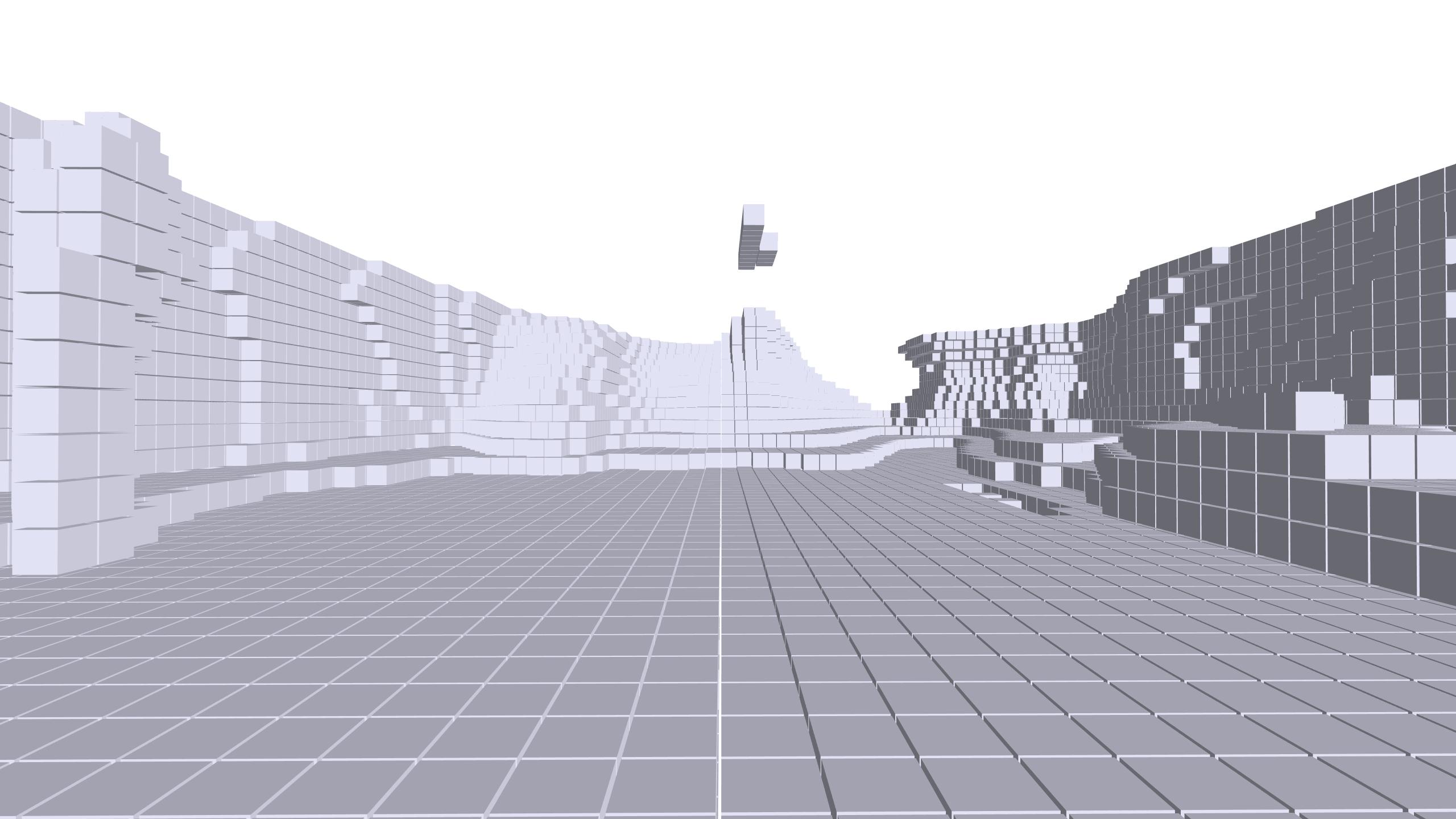}} &
    {\includegraphics[width=0.165\linewidth]{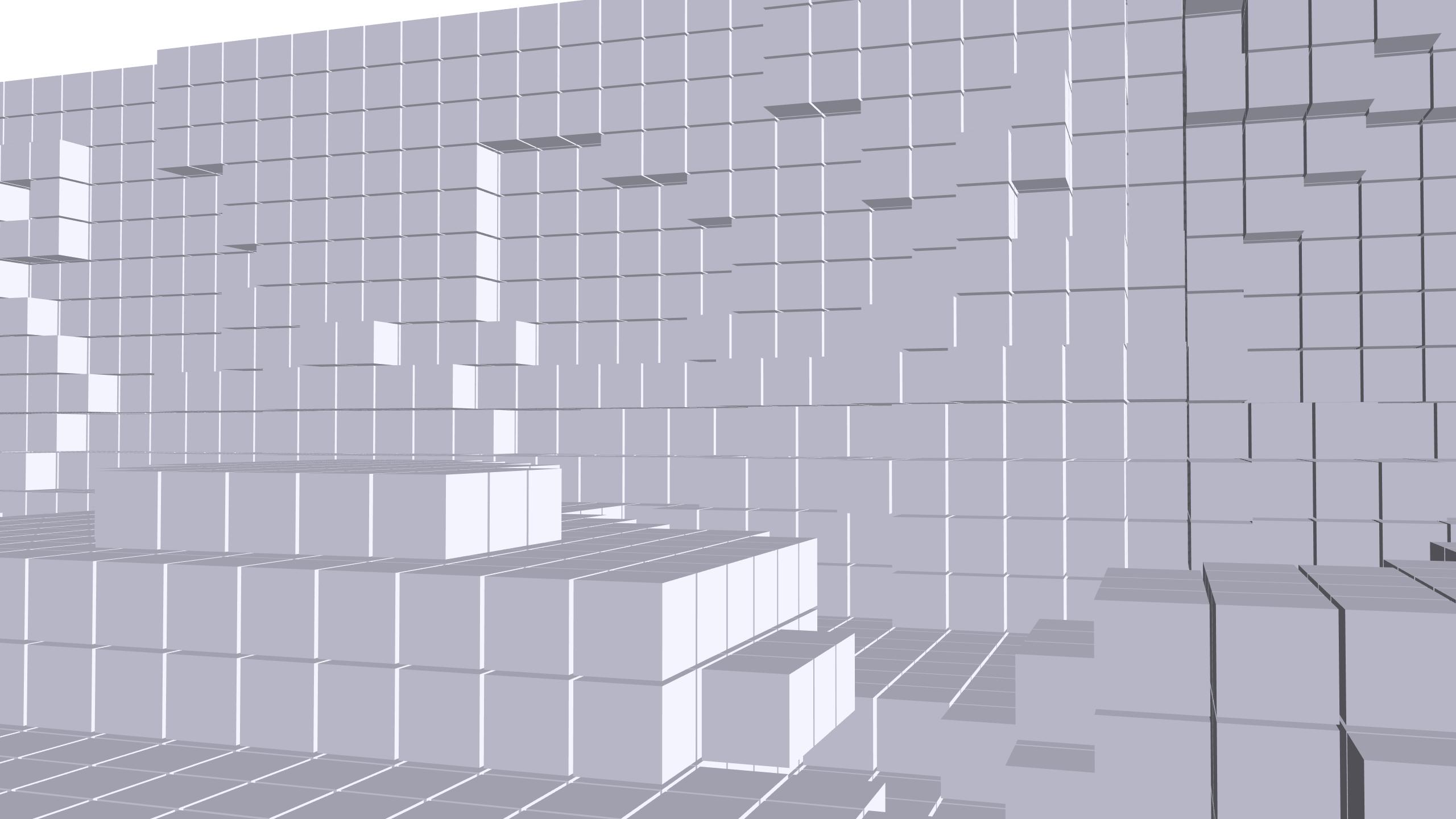}} \\
    
    front left & front & front right & back right & back & back left \\
    \end{tabular}
    \centering
    \caption{\textbf{Qualitative results on nuScenes dataset~\cite{nuscenes}}. Our method can predict visually appealing depth maps with texture details and fine-grained occupancy. Better viewed when zoomed in.}
\label{fig:qualitative}
\end{figure*}

\definecolor{nbarrier}{RGB}{255, 120, 50}
\definecolor{nbicycle}{RGB}{255, 192, 203}
\definecolor{nbus}{RGB}{255, 255, 0}
\definecolor{ncar}{RGB}{0, 150, 245}
\definecolor{nconstruct}{RGB}{0, 255, 255}
\definecolor{nmotor}{RGB}{200, 180, 0}
\definecolor{npedestrian}{RGB}{255, 0, 0}
\definecolor{ntraffic}{RGB}{255, 240, 150}
\definecolor{ntrailer}{RGB}{135, 60, 0}
\definecolor{ntruck}{RGB}{160, 32, 240}
\definecolor{ndriveable}{RGB}{255, 0, 255}
\definecolor{nother}{RGB}{139, 137, 137}
\definecolor{nsidewalk}{RGB}{75, 0, 75}
\definecolor{nterrain}{RGB}{150, 240, 80}
\definecolor{nmanmade}{RGB}{230, 230, 250}
\definecolor{nvegetation}{RGB}{0, 175, 0}
\definecolor{nothers}{RGB}{0, 0, 0}

\begin{table*}[t]
    \caption{\textbf{3D Occupancy prediction performance on the Occ3D-nuScenes dataset~\cite{occ3d}.} `GT' indicates occupancy ground truth. Since `other' and `other flat' classes are the invalid prompts for open-vocabulary models, we do not consider these two classes during evaluation. `mIoU*' is the original result, and `mIoU' is the result ignoring the classes.}

	\footnotesize
 	\setlength{\tabcolsep}{0.0025\linewidth}
	
	\newcommand{\classfreq}[1]{{~\tiny(\nuscenesfreq{#1}\%)}}  %
    \begin{center}
	\resizebox{1.0\textwidth}{!}{
	\begin{tabular}{l|c|c c| c c c c c c c c c c c c c c c}
		\toprule
		Method
		& GT &  mIoU & mIoU*

		& \rotatebox{90}{\textcolor{nbarrier}{$\blacksquare$} barrier}
		
		& \rotatebox{90}{\textcolor{nbicycle}{$\blacksquare$} bicycle}
		
		& \rotatebox{90}{\textcolor{nbus}{$\blacksquare$} bus}

		& \rotatebox{90}{\textcolor{ncar}{$\blacksquare$} car}

		& \rotatebox{90}{\textcolor{nconstruct}{$\blacksquare$} const. veh.}

		& \rotatebox{90}{\textcolor{nmotor}{$\blacksquare$} motorcycle}

		& \rotatebox{90}{\textcolor{npedestrian}{$\blacksquare$} pedestrian}

		& \rotatebox{90}{\textcolor{ntraffic}{$\blacksquare$} traffic cone}

		& \rotatebox{90}{\textcolor{ntrailer}{$\blacksquare$} trailer}

		& \rotatebox{90}{\textcolor{ntruck}{$\blacksquare$} truck}

		& \rotatebox{90}{\textcolor{ndriveable}{$\blacksquare$} drive. suf.}

		& \rotatebox{90}{\textcolor{nsidewalk}{$\blacksquare$} sidewalk}

		& \rotatebox{90}{\textcolor{nterrain}{$\blacksquare$} terrain}

		& \rotatebox{90}{\textcolor{nmanmade}{$\blacksquare$} manmade}

		& \rotatebox{90}{\textcolor{nvegetation}{$\blacksquare$} vegetation}
        
		\\
		\midrule
    MonoScene~\cite{monoscene} & \checkmark  & 6.33 & 6.06  & 7.23 & 4.26 & 4.93 & 9.38 & 5.67 & 3.98 & 3.01 & 5.90 & 4.45 & 7.17 & 14.91  & 7.92 & 7.43 & 1.01 & 7.65  \\
    TPVFormer~\cite{tpvformer} & \checkmark  & 28.69 & 27.83 & 38.90 & 13.67 & \textbf{40.78} & \textbf{45.90} & \textbf{17.23} & 19.99 & 18.85 & 14.30 & \textbf{26.69} & \textbf{34.17}  & 55.65 & 37.55 & 30.70 & 19.40 & 16.78  \\
    BEVDet ~\cite{bevdet} & \checkmark  & 20.03 & 19.38 & 30.31 & 0.23 & 32.26 & 34.47 & 12.97 & 10.34 & 10.36 & 6.26 & 8.93 & 23.65 & 52.27 & 26.06 & 22.31 & 15.04 & 15.10  \\
    OccFormer~\cite{occformer}& \checkmark  & 22.39 & 21.93 & 30.29 & 12.32 & 34.40 & 39.17 & 14.44 & 16.45 & 17.22 & 9.27 & 13.90 & 26.36 & 50.99 &  34.66 & 22.73 & 6.76 & 6.97  \\
    BEVFormer~\cite{bevformer} & \checkmark  & 28.13 & 26.88 & 37.83 & 17.87 & 40.44 & 42.43 & 7.36 & 23.88 & 21.81 & 20.98 & 22.38 & 30.70 & 55.35 & 36.0 & 28.06 & 20.04 & 17.69 \\
    CTF-Occ~\cite{occ3d} & \checkmark  & \textbf{29.54} & \textbf{28.53} &\textbf{39.33} &\textbf{20.56} &38.29& 42.24 &16.93 &\textbf{24.52} &\textbf{22.72} &\textbf{21.05} &22.98 &31.11& 53.33 & 37.98 &33.23 &\textbf{20.79} &18.00 \\
    RenderOcc~\cite{renderocc} & \checkmark  & 24.53 &23.93&27.56 &14.36 &19.91 &20.56 &11.96 &12.42 &12.14 &14.34 &20.81 &18.94 &\textbf{68.85} & \textbf{42.01} &\textbf{43.94} &17.36& \textbf{22.61} \\ 
    
        \midrule
        SimpleOcc~\cite{simpleocc} & $\times$  & 7.99 & 7.05 & 0.67 &1.18 &3.21 &7.63 &1.02& 0.26 &1.80 &0.26 &1.07 &2.81 &40.44 &18.30 &17.01 &13.42 &10.84 \\
         OccNeRF &  $\times$ & 10.81 & 9.53 & 0.83 & 0.82 & 5.13 & 12.49 & 3.50 & 0.23 & 3.10 & 1.84 & 0.52 & 3.90 & 52.62 & 20.81 & 24.75 & 18.45 & 13.19 \\

		\bottomrule
	\end{tabular}}
    \end{center}
    \label{tab:occ}
\end{table*}

\begin{figure*}
    \centering
    \vspace{2mm}
    %\scriptsize
    \setlength\tabcolsep{1.0pt} % default value: 6pt
    \renewcommand{\arraystretch}{1.0}
    \begin{tabular}{cccccc}
    {\includegraphics[width=0.165\linewidth]{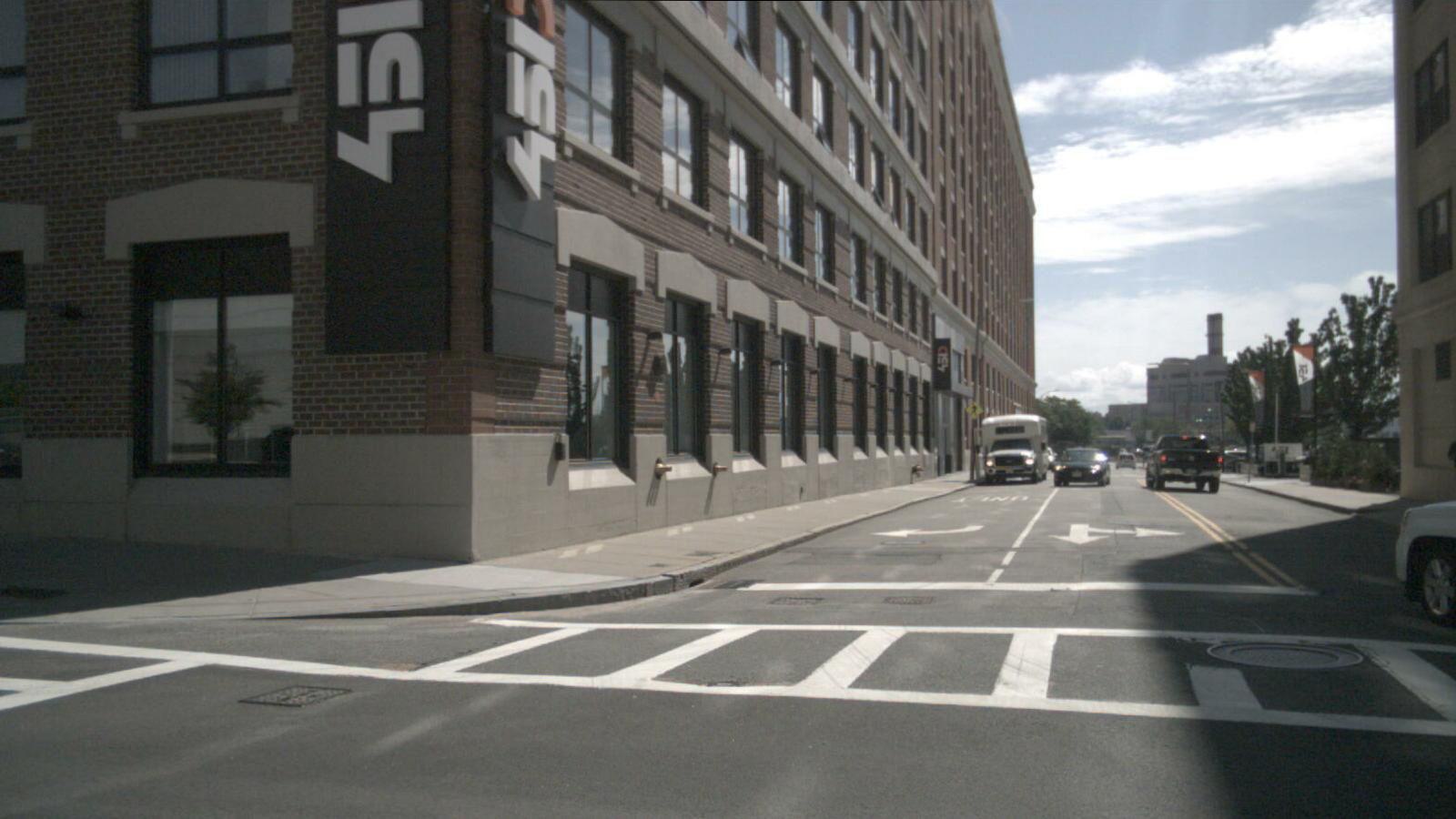}} &
    {\includegraphics[width=0.165\linewidth]{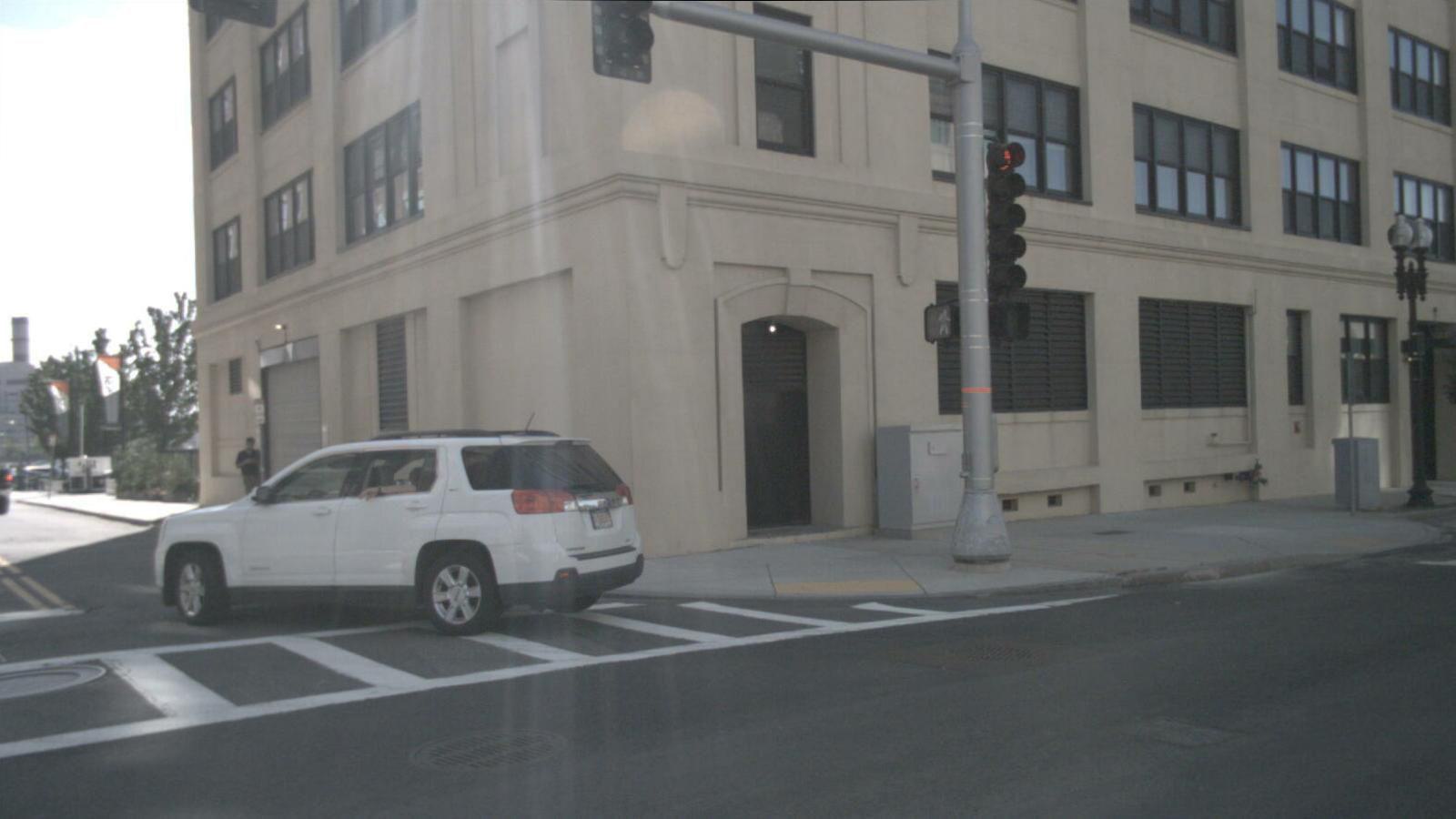}} &
    {\includegraphics[width=0.165\linewidth]{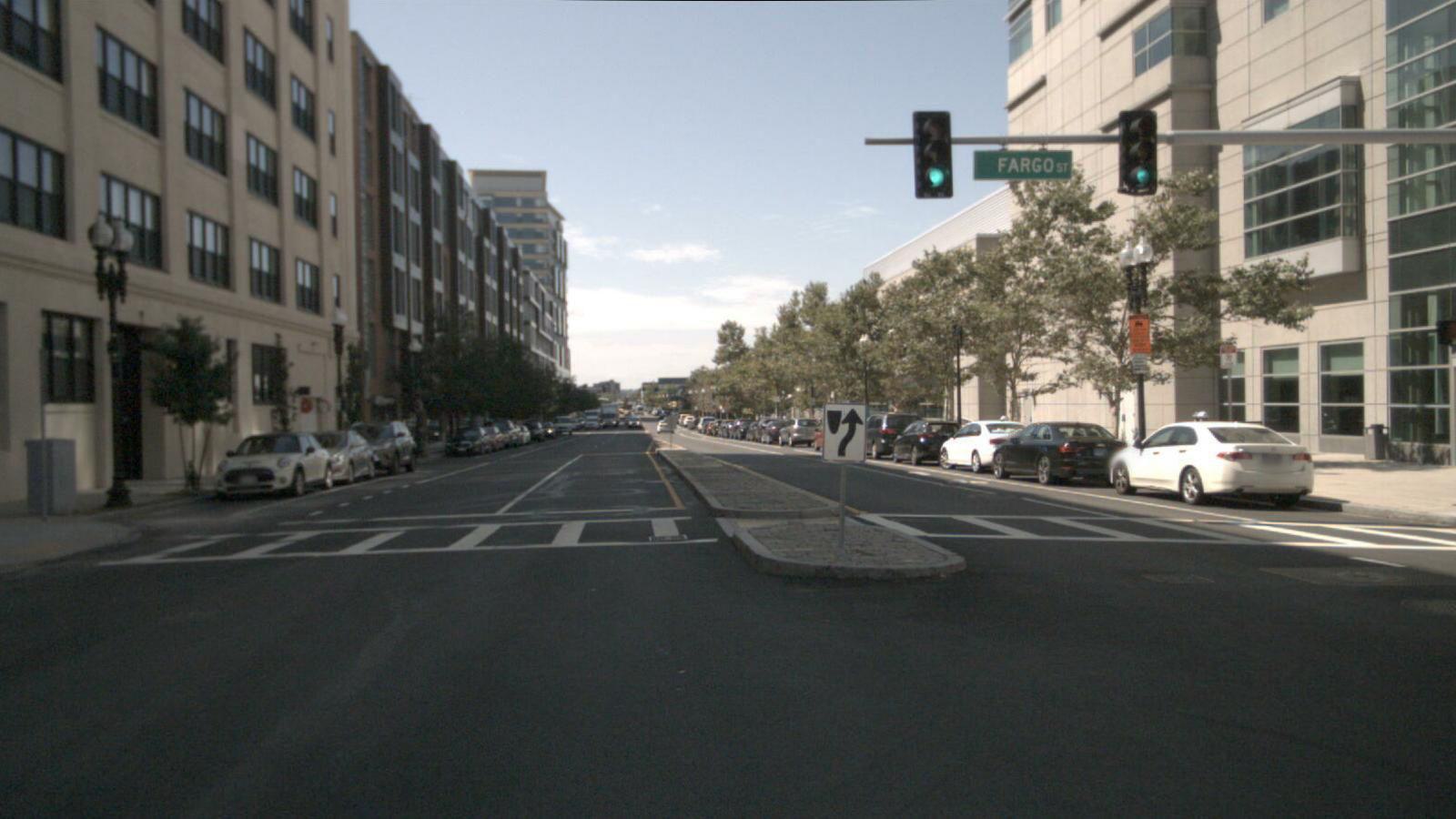}} &
    {\includegraphics[width=0.165\linewidth]{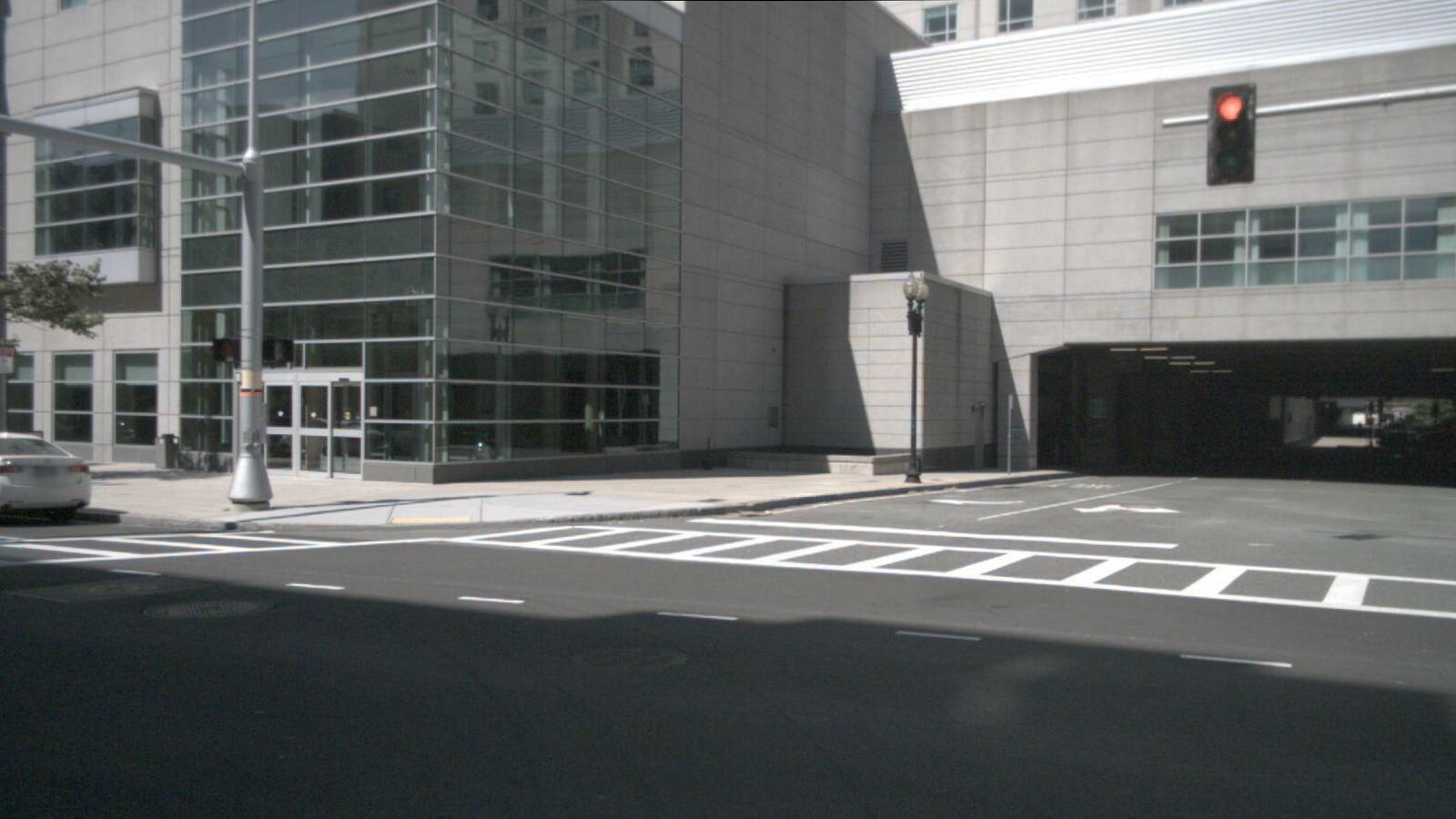}} &
    {\includegraphics[width=0.165\linewidth]{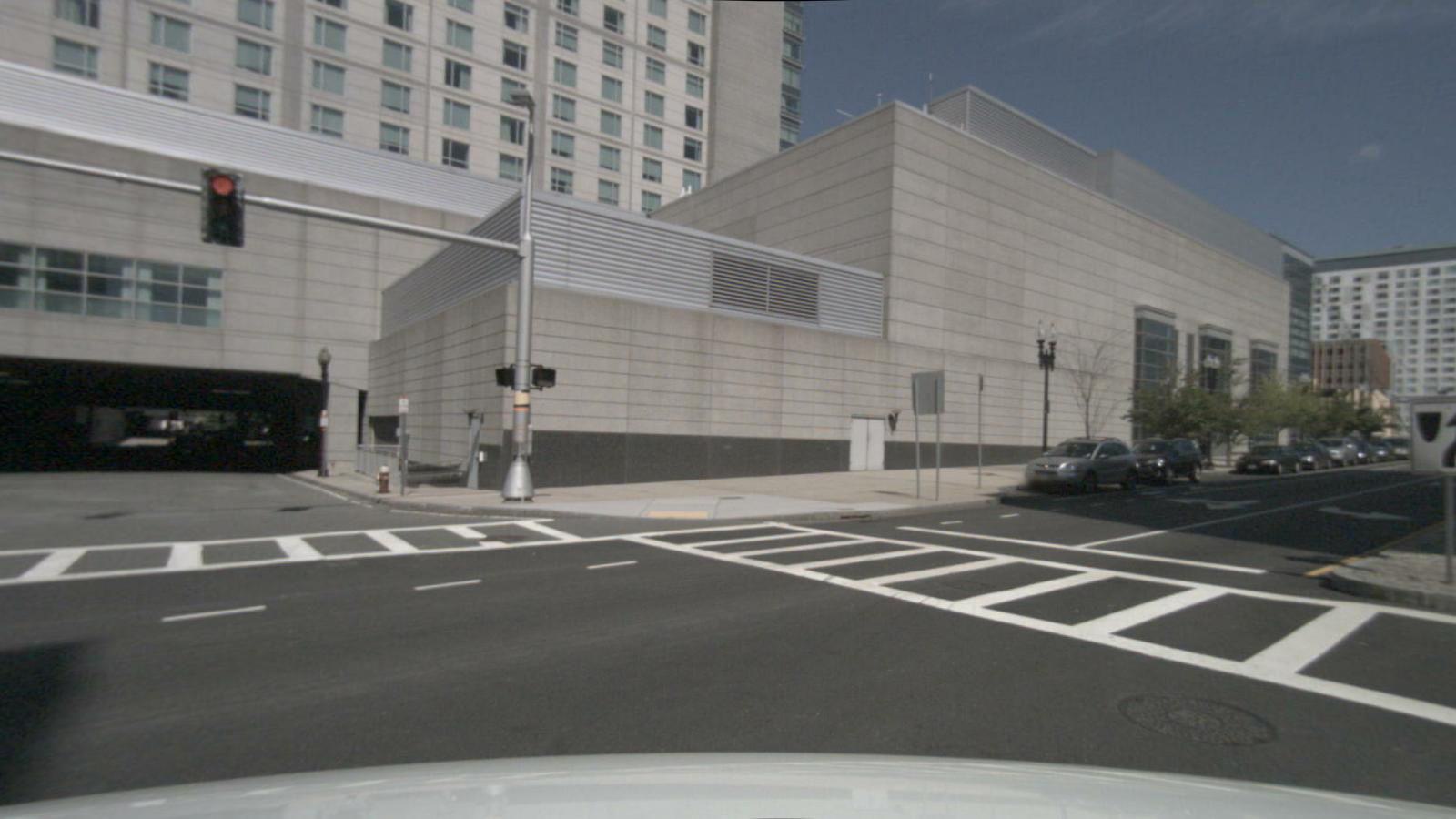}} &
    {\includegraphics[width=0.165\linewidth]{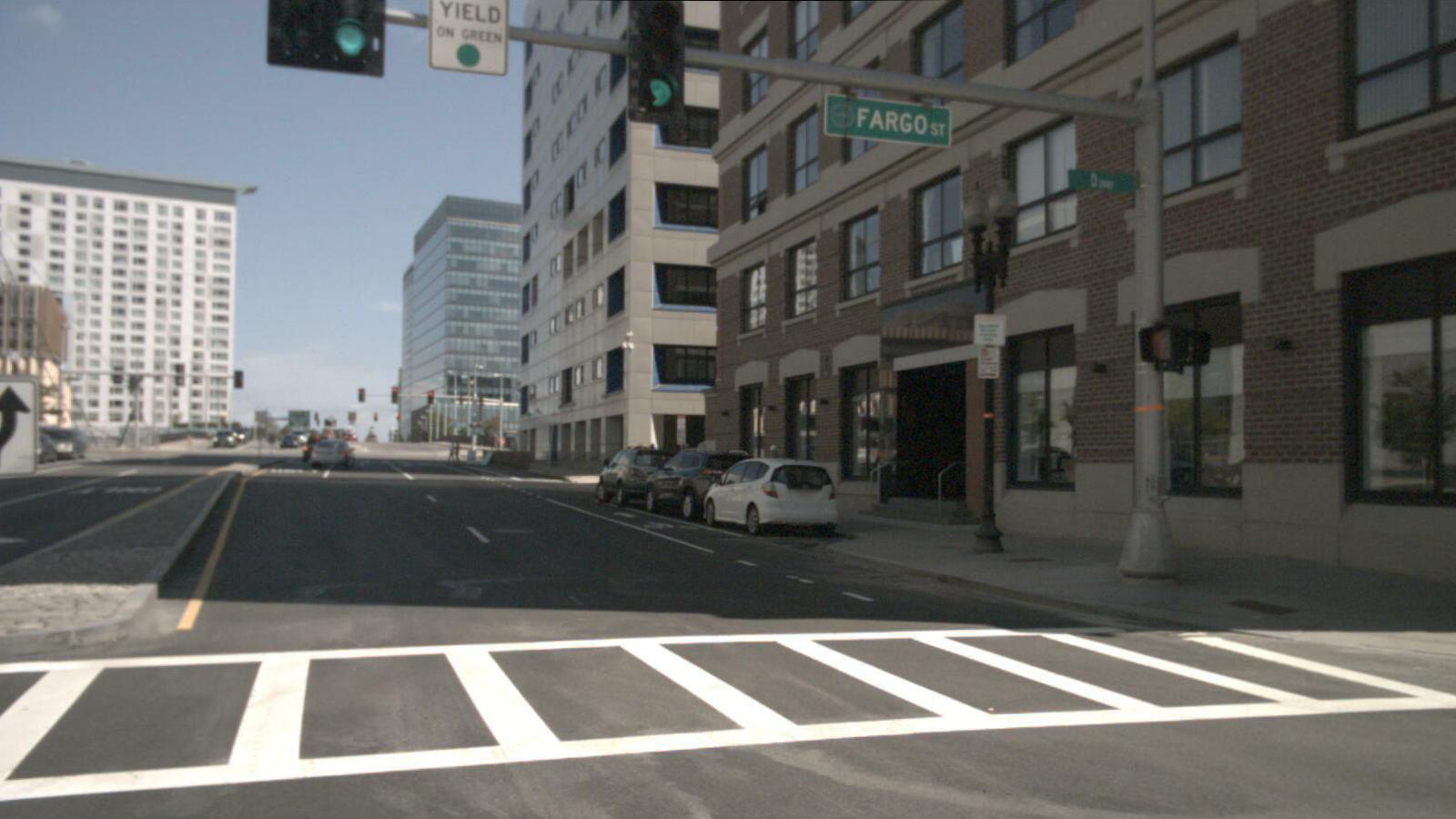}} \\
    
    {\includegraphics[width=0.165\linewidth]{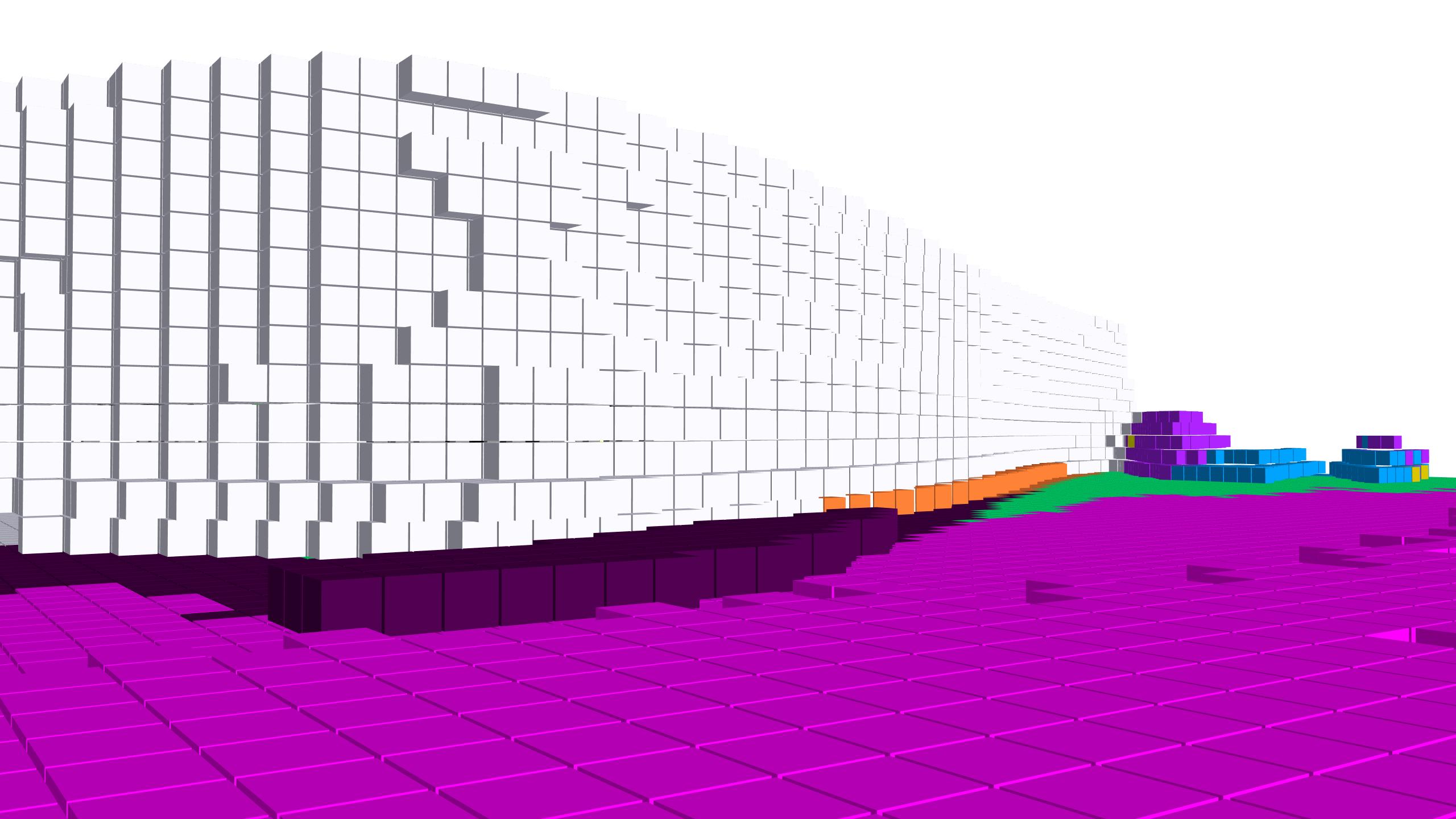}} &
    {\includegraphics[width=0.165\linewidth]{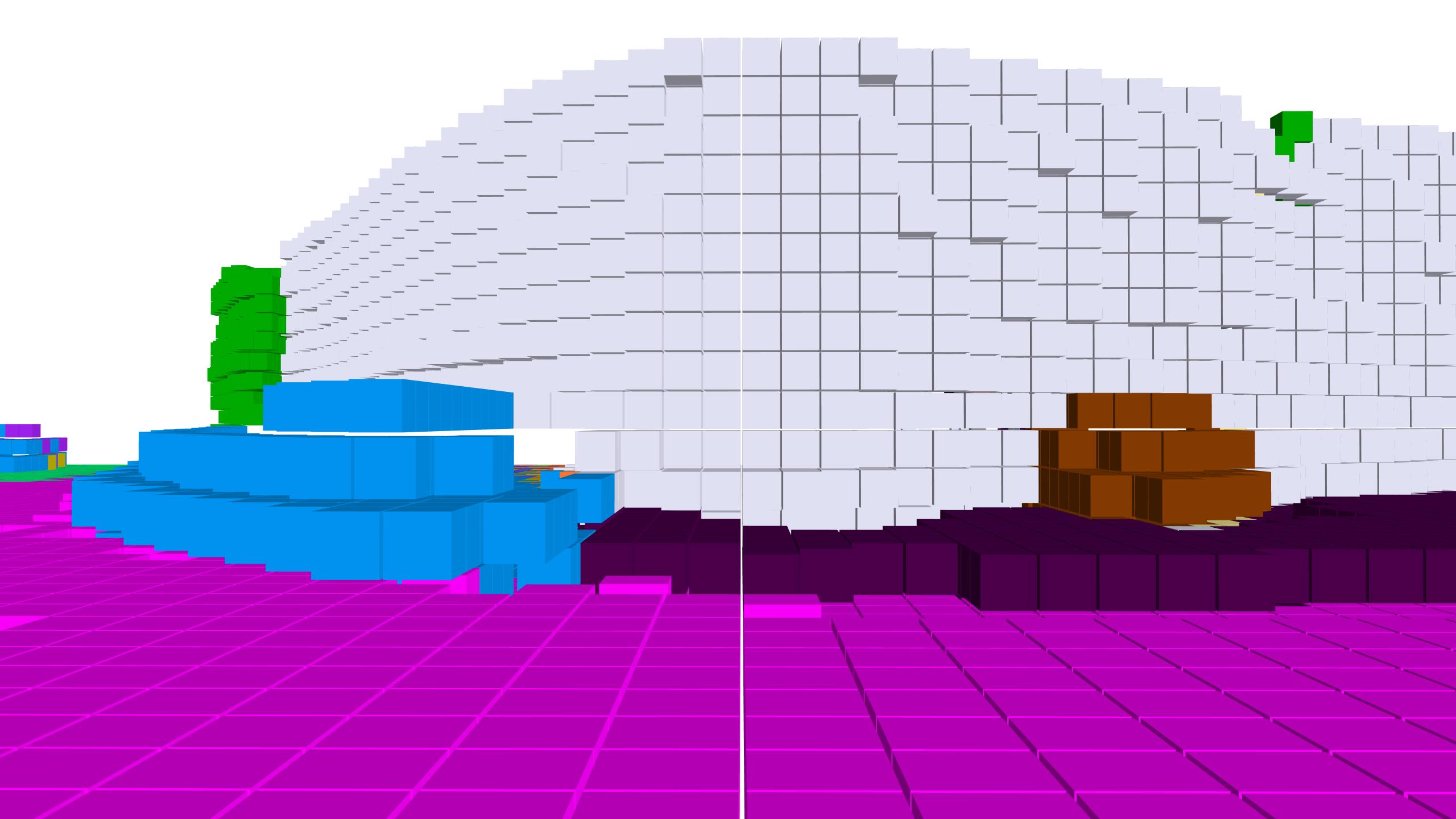}} &
    {\includegraphics[width=0.165\linewidth]{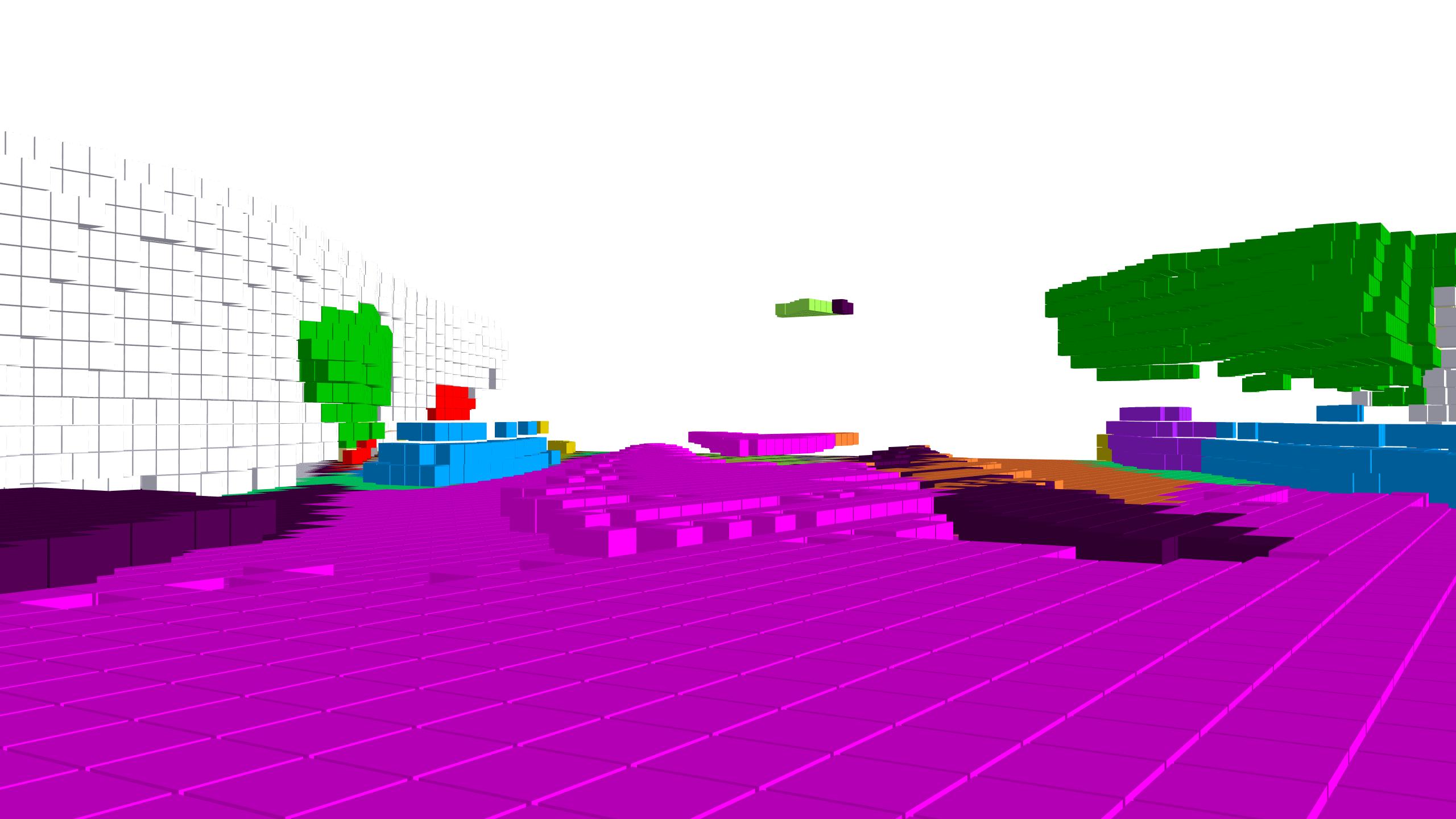}} &
    {\includegraphics[width=0.165\linewidth]{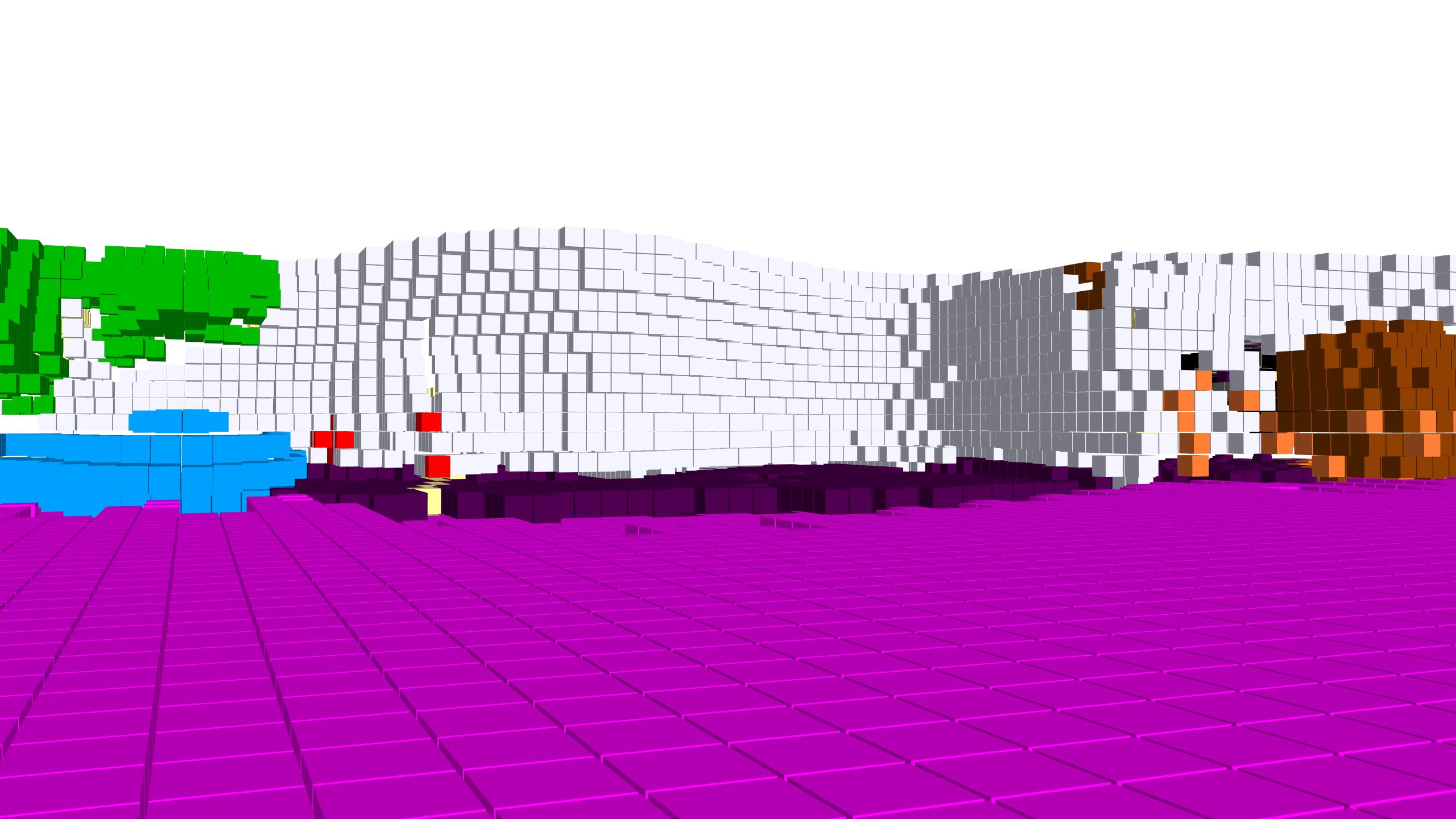}} &
    {\includegraphics[width=0.165\linewidth]{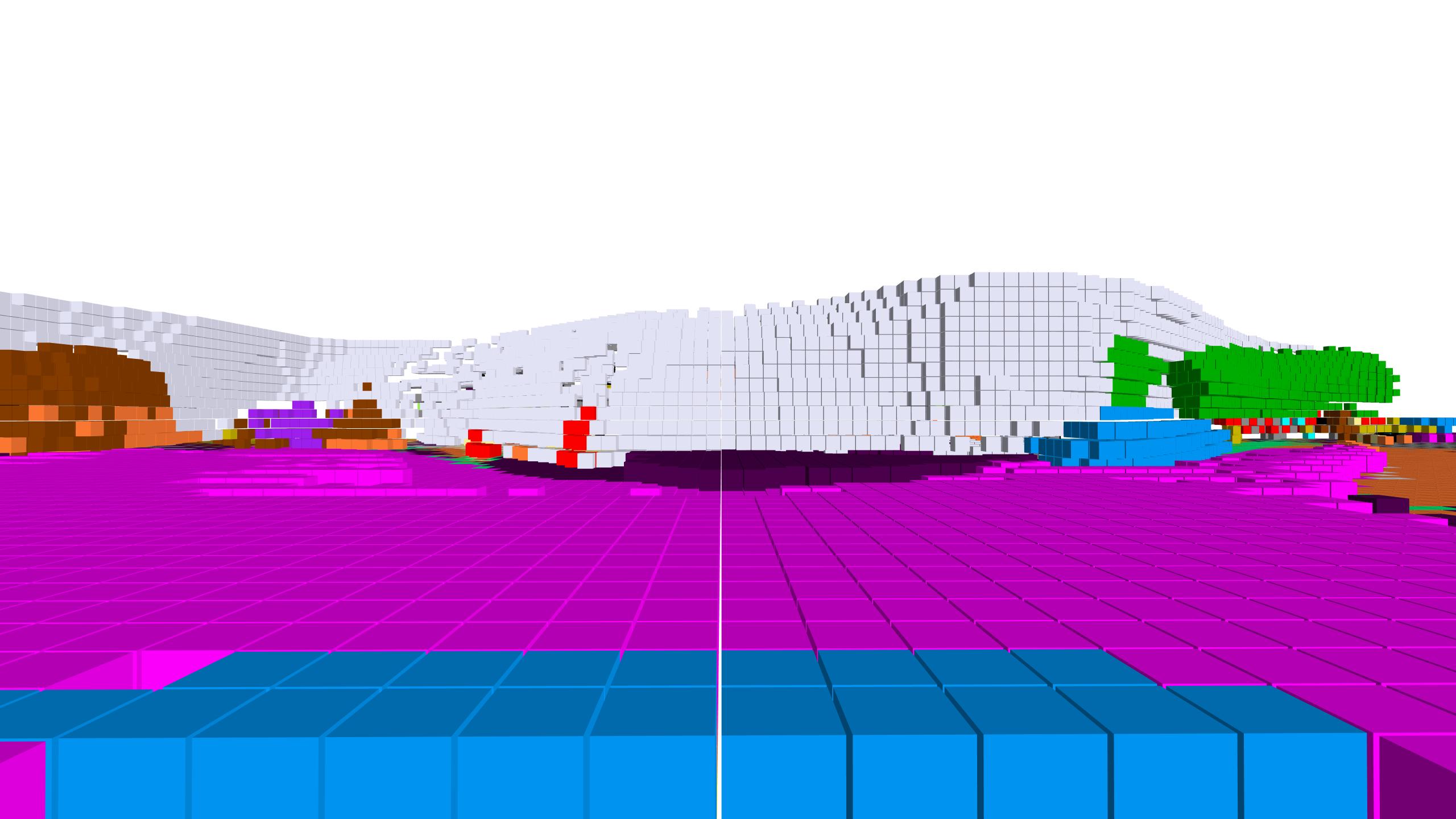}} &
    {\includegraphics[width=0.165\linewidth]{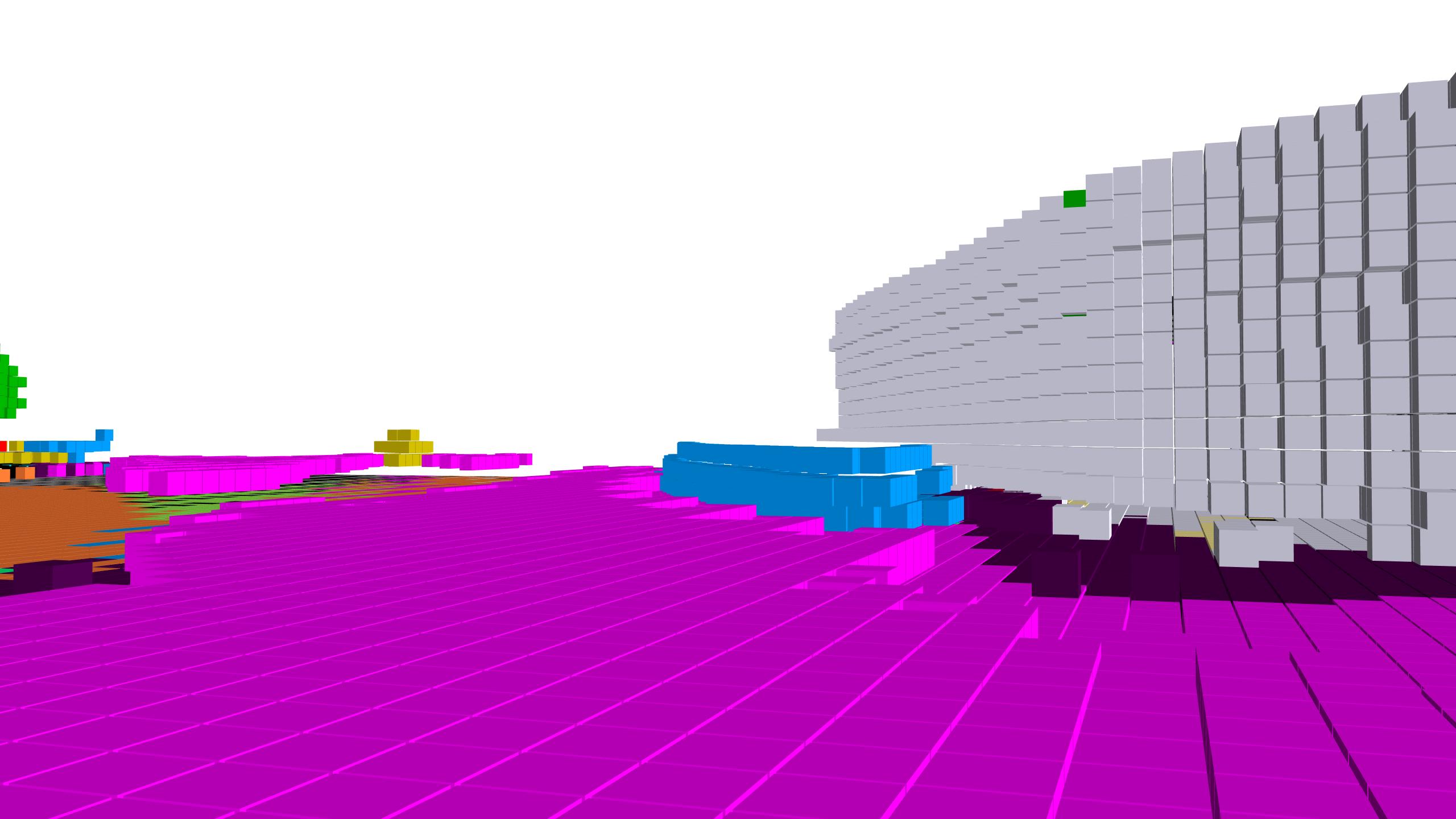}} \\

    {\includegraphics[width=0.165\linewidth]{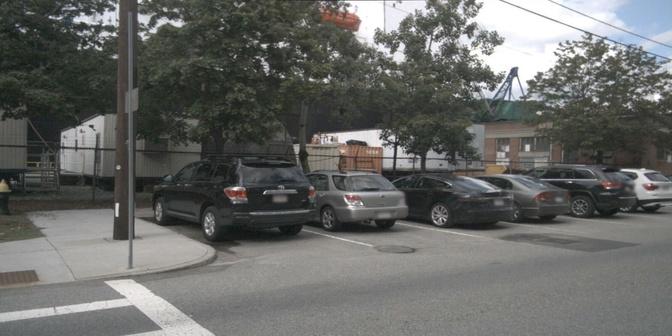}} &
    {\includegraphics[width=0.165\linewidth]{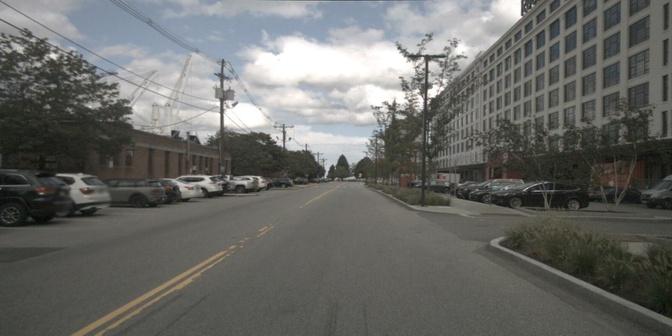}} &
    {\includegraphics[width=0.165\linewidth]{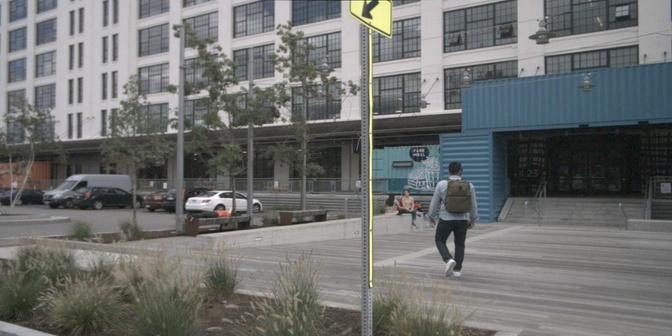}} &
    {\includegraphics[width=0.165\linewidth]{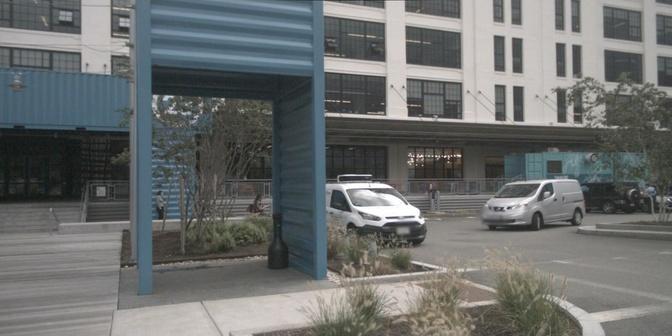}} &
    {\includegraphics[width=0.165\linewidth]{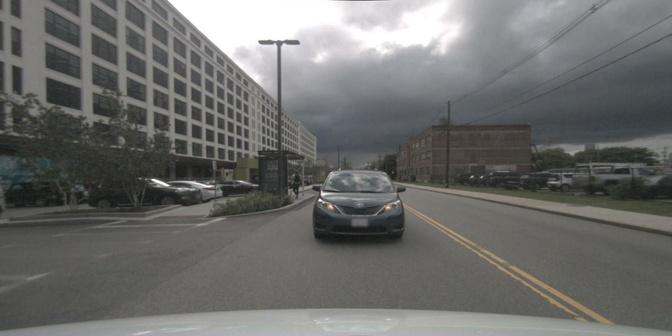}} &
    {\includegraphics[width=0.165\linewidth]{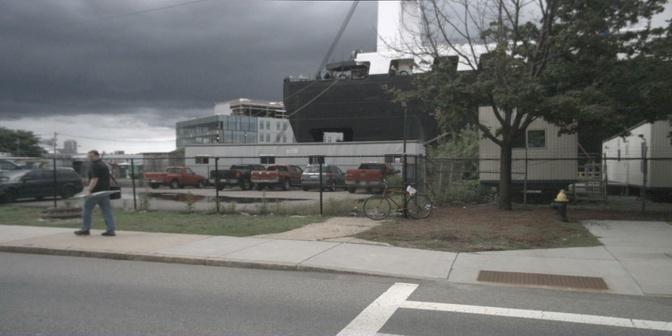}} \\
    
    {\includegraphics[width=0.165\linewidth]{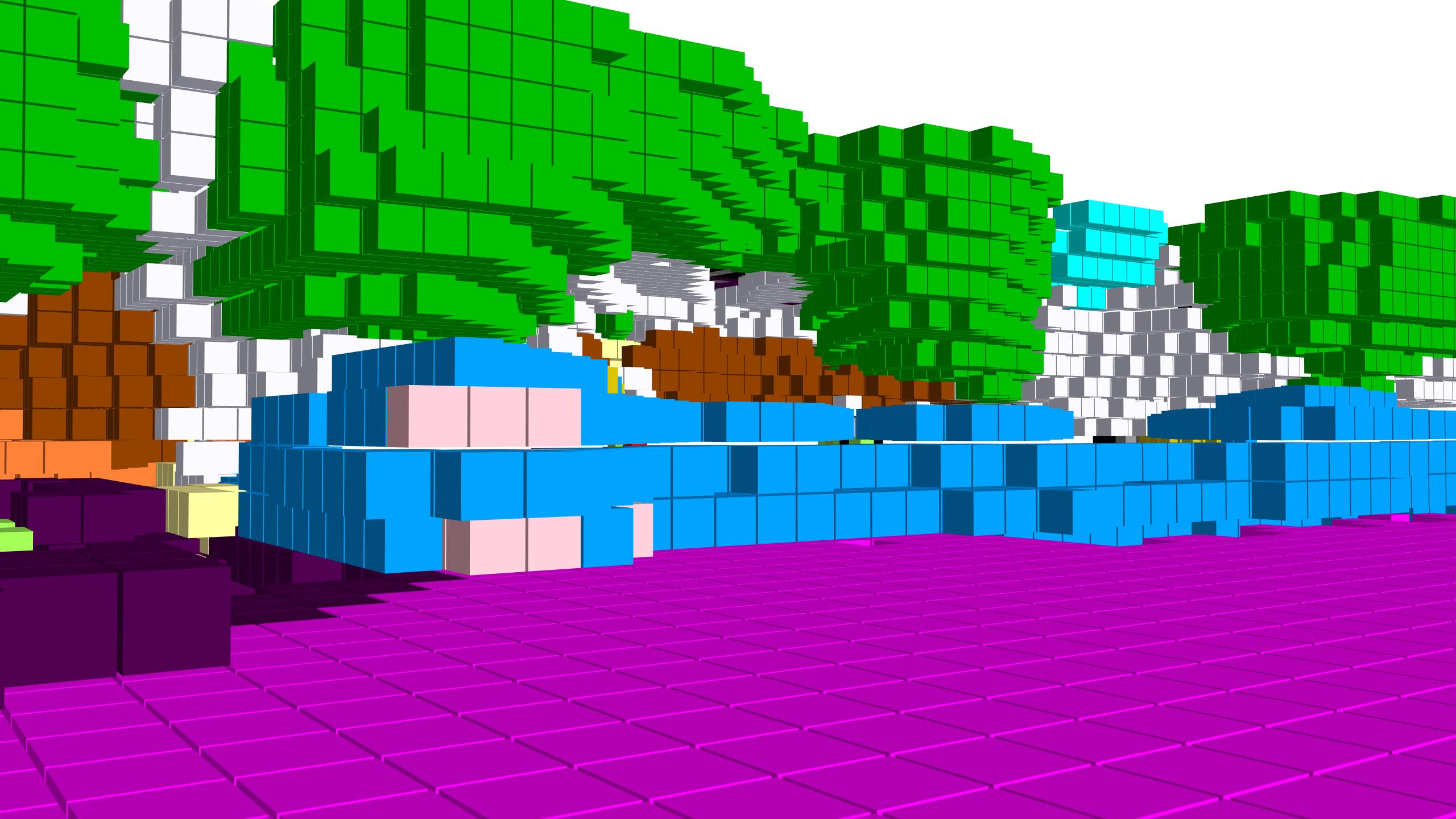}} &
    {\includegraphics[width=0.165\linewidth]{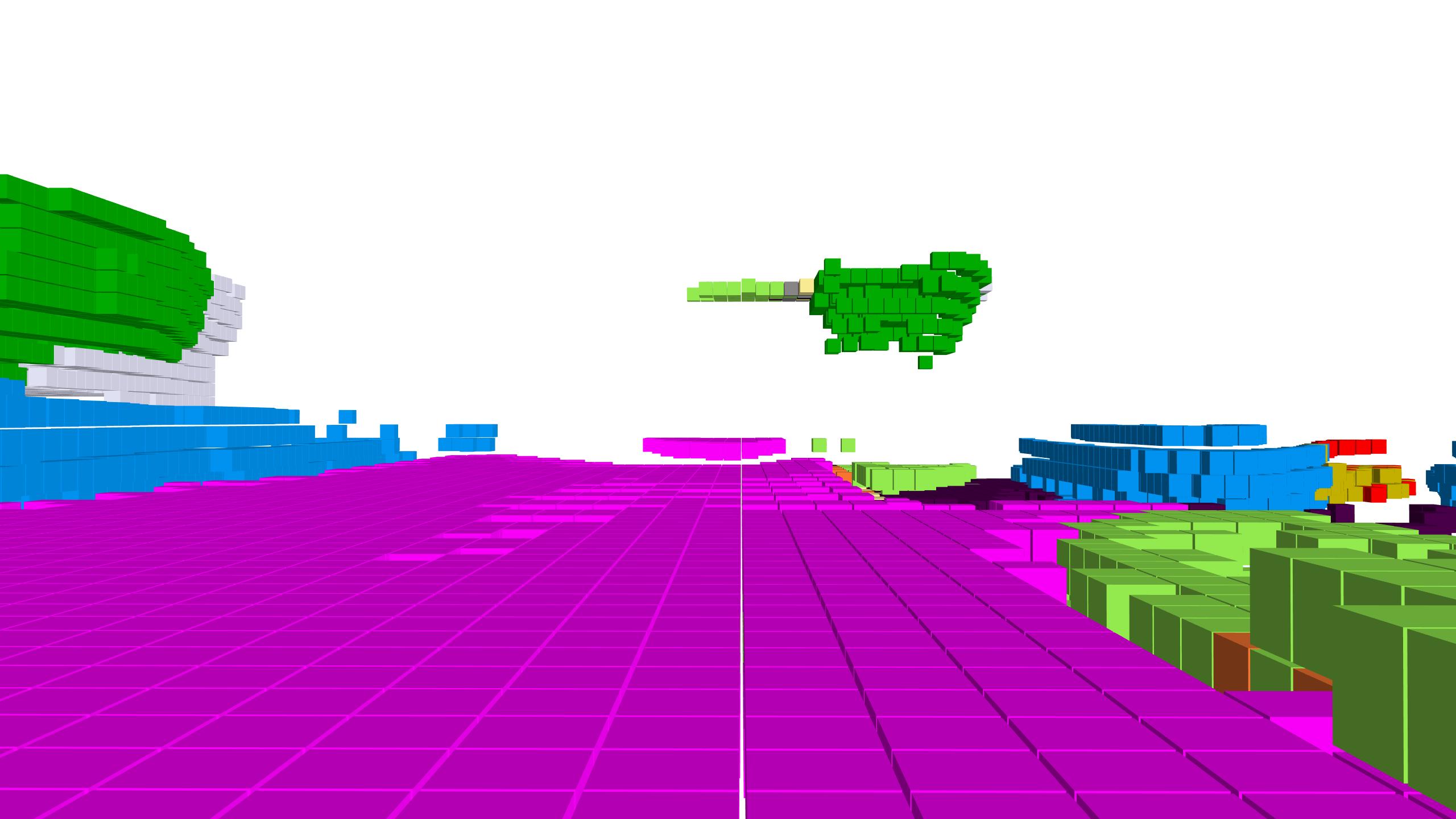}} &
    {\includegraphics[width=0.165\linewidth]{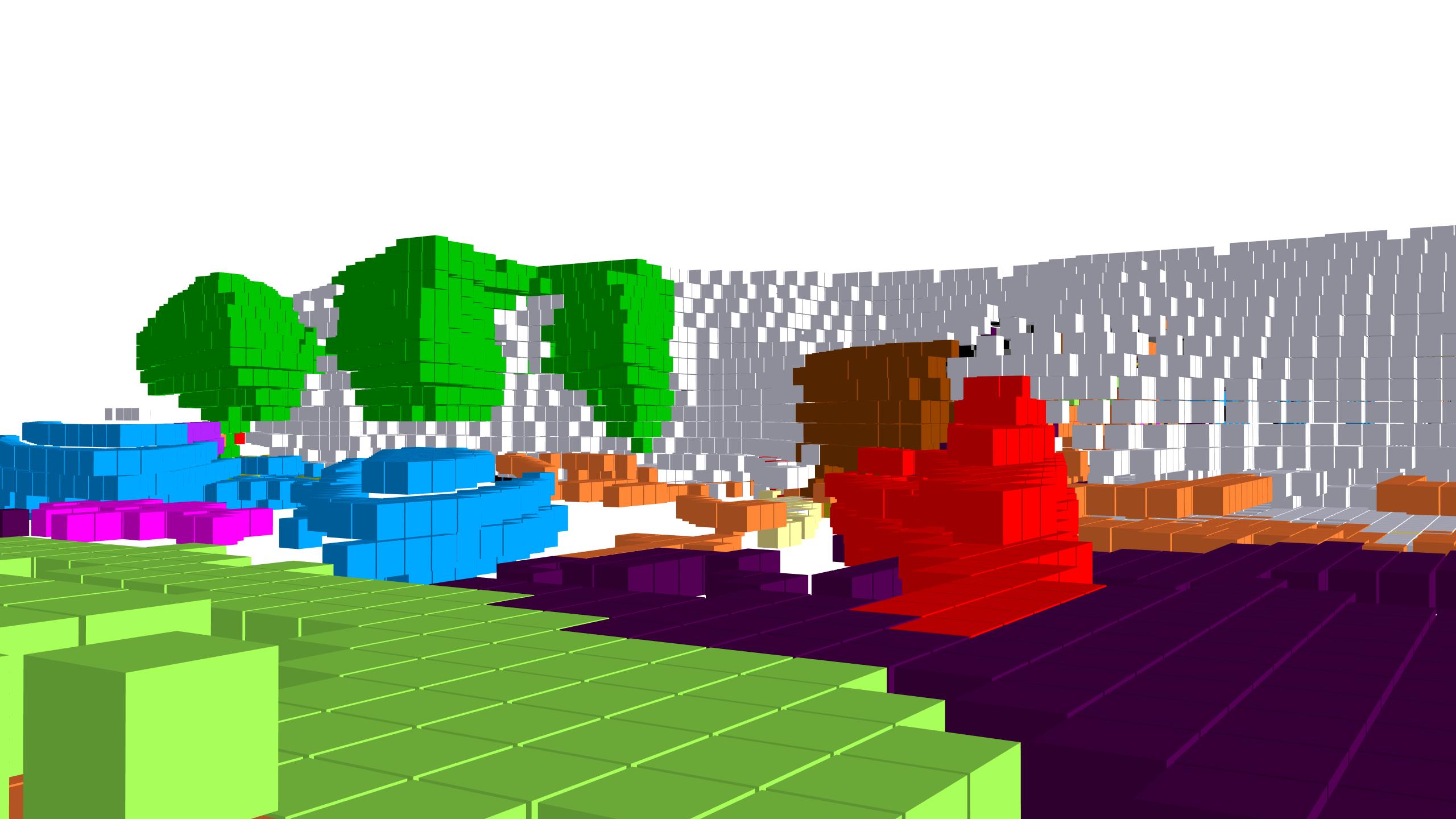}} &
    {\includegraphics[width=0.165\linewidth]{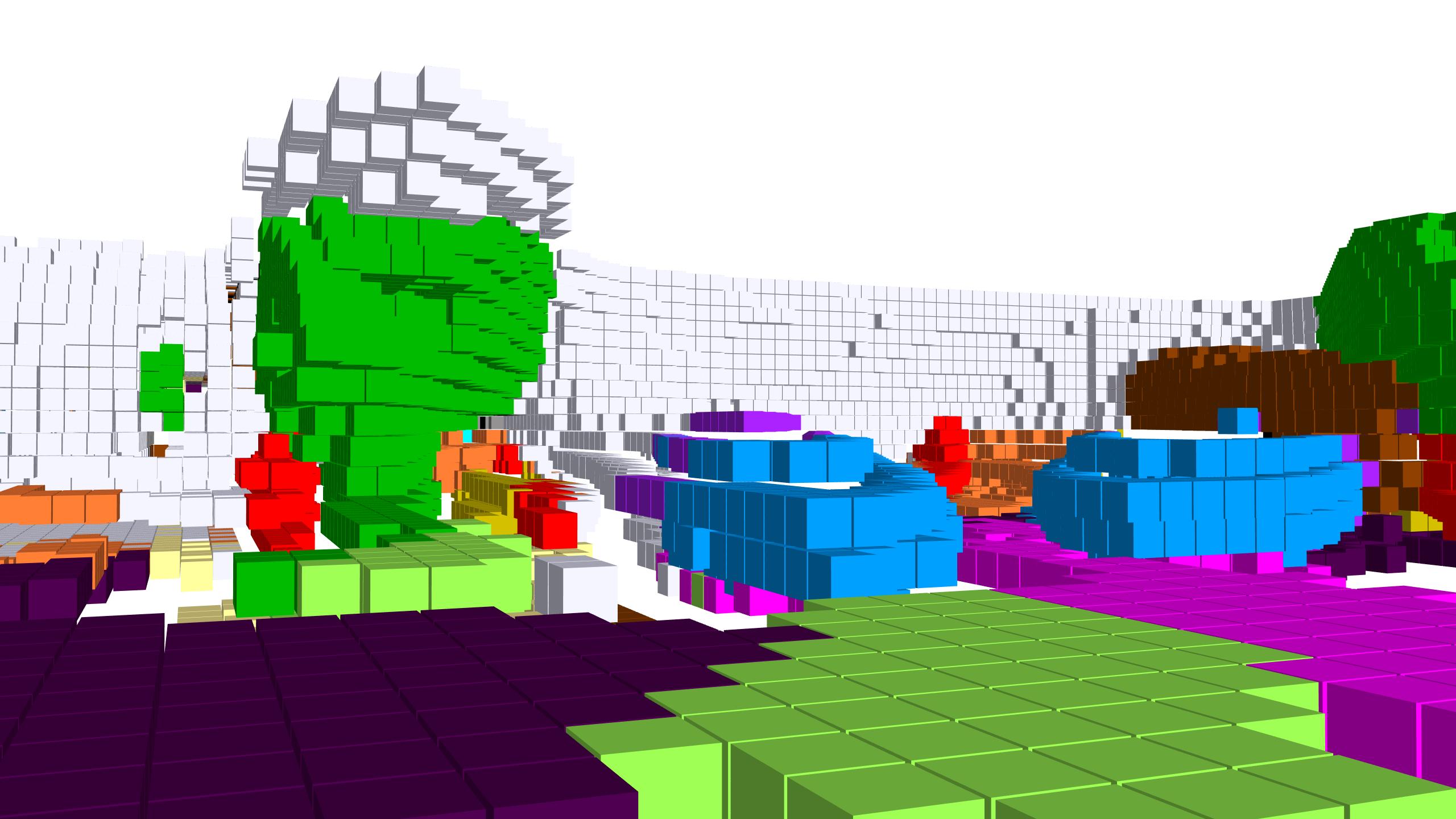}} &
    {\includegraphics[width=0.165\linewidth]{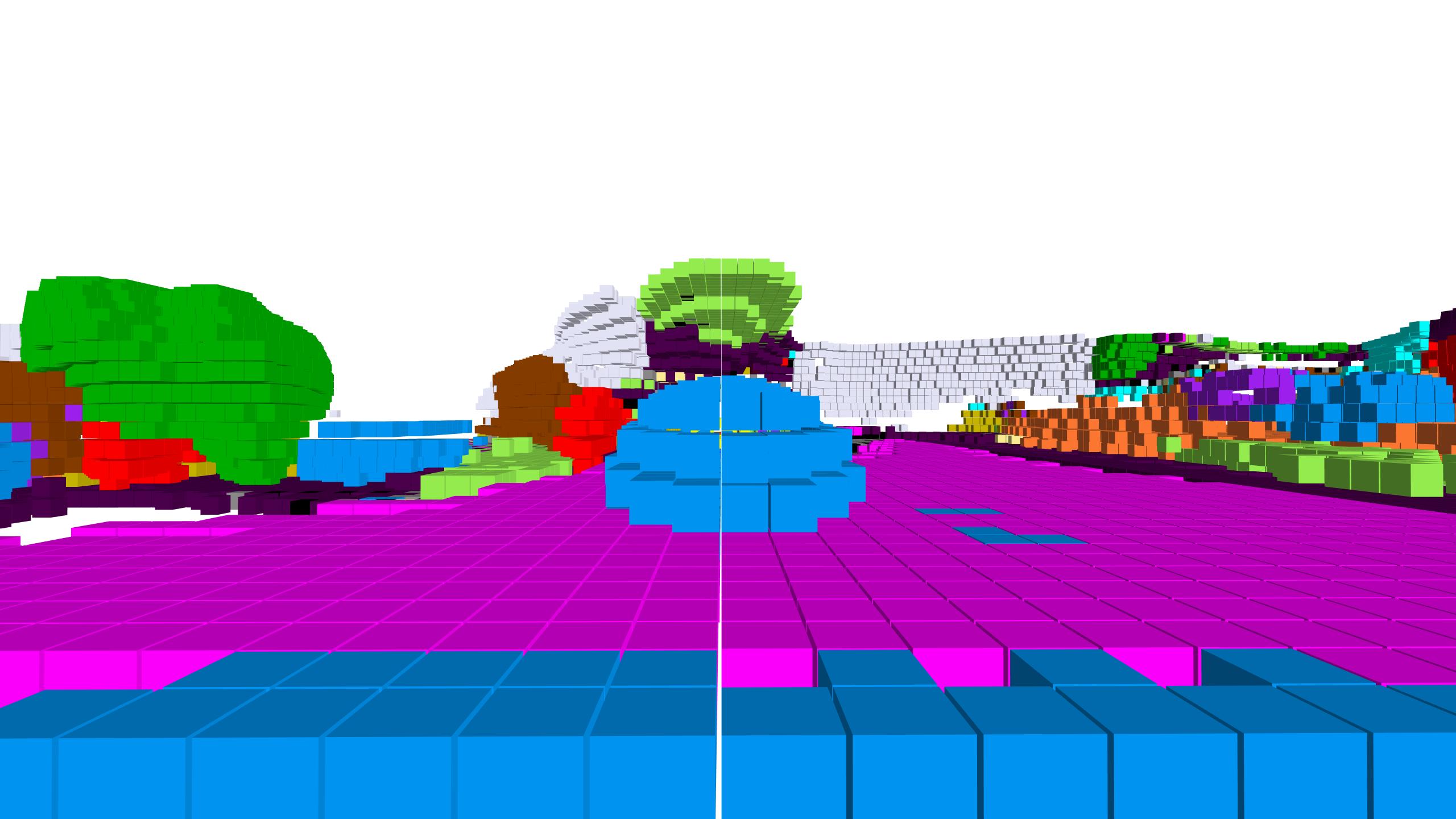}} &
    {\includegraphics[width=0.165\linewidth]{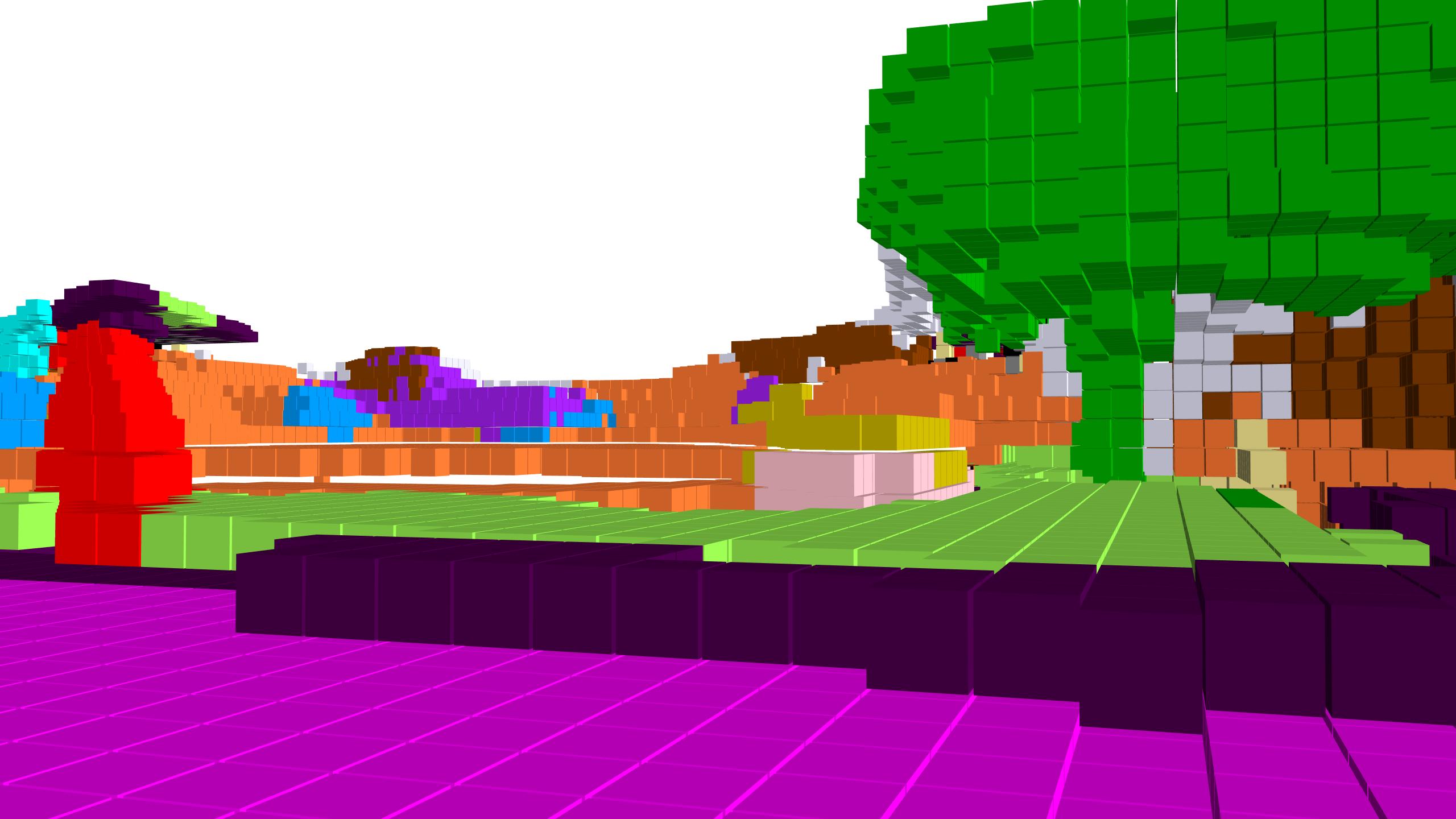}} \\
    
    front left & front & front right & back right & back & back left \\
    \end{tabular}
    \centering
    \caption{\textbf{Qualitative results of semantic occupancy on nuScenes dataset~\cite{nuscenes}}. Our method can predict visually appealing semantic occupancy with well geometry correspondence. Better viewed when zoomed in.}
\label{fig:semantic}
\end{figure*}

\section{Experiment}
\label{sec:experiment}

\definecolor{col1}{RGB}{232, 161, 148}
\definecolor{col2}{RGB}{148, 187, 232}

\subsection{Experimental Setup}
\noindent \textbf{Dataset:}
Our experiments are conducted on nuScenes~\cite{nuscenes} and SemanticKITTI~\cite{semantickitti} datasets. NuScenes~\cite{nuscenes} is a large-scale autonomous driving dataset which contains 600 scenes for training, 150 scenes for validation, and 150 for testing. The dataset has about 40000 frames and 17 classes in total. For self-supervised depth estimation, we project LiDAR point clouds to each view to get depth ground truth for evaluation. Following SurroundDepth~\cite{surrounddepth}, we clip the depth prediction and ground truth from 0.1m to 80m. To evaluate the semantic occupancy prediction, we use Occ3D-nuScenes~\cite{occ3d} benchmark. The range of each sample is [-40 m, -40 m, -1 m, 40 m, 40 m, 5.4 m] and the voxel size is 0.4 m. Among 17 classes, we do not consider `other' and `other flat' classes for evaluation since open-vocabulary models cannot recognize the semantic-ambiguous text. Following ~\cite{surrounddepth,occ3d}, we evaluate models on validation sets. 
Additionally, we explore 3D occupancy prediction using the SemanticKITTI~\cite{semantickitti} dataset. SemanticKITTI~\cite{semantickitti} is composed of 22 sequences (10 for training, 1 for validation and 11 for test) of scans and each scan contains voxelized LiDAR data and corresponding stereo images. The range of each sample is [-25.6 m, 0, -2.0 m, 25.6 m, 51.2 m, 4.4 m] and the voxel size is 0.2 m.

\noindent \textbf{Implementation Details:} % TODO
We adopt ResNet-101~\cite{he2016deep} with ImageNet~\cite{deng2009imagenet} pretrained weights as the 2D backbone to extract multi-camera features. For nuScenes~\cite{nuscenes}, the predicted occupancy field has the shape 300x300x24. The central 200x200x16 voxels represent inside regions: -40m to 40m for the X and Y axis, and -1m to 5.4m for the Z axis, which is the same as the scope defined in Occ3D-nuScenes. For SemanticKITTI~\cite{semantickitti}, the shape of the predicted occupancy field is 320x320x40. The central part is 256x256x32, also representing the scope defined by the dataset. We render 3 frame depth maps, which are supervised by the photometric loss with a sequence of 5 frame raw images (1 keyframe with 4 neighbored non-key frames). The $\alpha$ is set as 0.667. To predict semantic occupancy, the Grounded-SAM ~\cite{dino,sam} is employed as our pretrained open-vocabulary model. The text and box thresholds are set as 0.2 and we use the loss weight $\lambda=0.05$. All experiments are conducted on 8 A100.

\subsection{Self-supervised Depth Estimation}

\noindent \textbf{Evaluation Metric:}
For depth estimation, we use the commonly used depth evaluation metrics~\cite{monodepth2,zhou2017unsupervised,surrounddepth} as outlined in the following:
\begin{itemize}
\item Abs Rel: $\frac1{|T|}\sum_{d\in T}|d - d^*| / d^*$,
\item Sq Rel: $\frac1{|T|}\sum_{d\in T}|d - d^*|^2 / d^*$,
\item RMSE: $\sqrt{\frac1{|T|}\sum_{d\in T}|d - d^*|^2}$,
\item RMSE log: $\sqrt{\frac1{|T|}\sum_{d\in T}|\log d - \log d^*|^2}$,
\item $\delta < t$: \% of $d$ s.t. $\max(\frac{d}{d^*},\frac{d^*}{d}) = \delta < t$,
\end{itemize}
where $d$ and $d^*$ indicate predicted and ground truth depths respectively, and T indicates all pixels on the depth image $D$. In our experiments, all the predicted depth maps are scale-aware and we do not perform any scale alignment. 
The Abs Rel is the main metric for depth estimation tasks and it reveals the relative errors of estimated depths. Note that during evaluation we do not perform median scaling since our method can predict real-world scale given ground truth poses. For the 3D occupancy predictiontask, we use the mean intersection over union (mIoU) of all classes as the semantic-aware metric and the intersection over union (IoU), precision and recall as the semantic-agnostic metrics.

TABLE \ref{tab:depth} shows the self-supervised multi-camera depth estimation results on nuScenes dataset. We do not use pretrained segmentation models in this experiment. The results are averaged over 6 cameras and `FSM*' is the reproduced FSM~\cite{fsm} result reported in ~\cite{kim2022self}. We can see that our method outperforms other SOTA methods by a large margin, demonstrating the effectiveness of OccNeRF. Compared to previous depth estimation methods, our approach directly predicts an occupancy field in 3D space instead of predicting per-pixel depths. This naturally guarantees multi-camera consistency. Furthermore, our method eliminates the need for post-processing steps that lift 2D depths to 3D point clouds.

\begin{table}[tb]
	\caption{\textbf{3D Occupancy prediction performance on the SemanticKITTI dataset~\cite{semantickitti}.} The results of other methods are from the table in SceneRF~\cite{scenerf}. MonoScene* is supervised by depth predictions from~\cite{monodepth2}}
	\scriptsize
	\centering
	\resizebox{0.48\textwidth}{!}{
		\begin{tabular}{l|ccc|ccc}
			\toprule
			\multirow{2}{*}{Method} & \multicolumn{3}{ c |}{Supervision} & \multirow{2}{*}{IoU} & \multirow{2}{*}{Prec.} & \multirow{2}{*}{Rec.} \\
             &  3D & Depth & Image & & &  \\
			\midrule
   
			MonoScene~\cite{monoscene} & \checkmark &  &  &37.14 & 49.90 & 59.24  \\ 
            \midrule
            
			LMSCNet$^{\text{rgb}}$~\cite{lmscnet} &  & \checkmark &  & 12.08 & 13.00 & 63.16  \\ 
			3DSketch$^{\text{rgb}}$~\cite{3dsketch} &  & \checkmark &  &   12.01 & 12.95 & 62.31  \\
			AICNet$^{\text{rgb}}$~\cite{aicnet}  &  & \checkmark &  & 11.28 & 11.84 & 70.89  \\ 
			MonoScene~\cite{monoscene} &  & \checkmark &  &  13.53 & 16.98 & 40.06  \\ 
			\midrule
			MonoScene*~\cite{monoscene} &  & &  \checkmark &  11.18 &  13.15 & 40.22  \\
			SceneRF~\cite{scenerf} &  & &  \checkmark & 13.84 & 17.28 & \textbf{40.96}  \\
                OccNeRF &  & &  \checkmark & \textbf{22.81} & \textbf{35.25} & 39.27  \\
   %              \midrule 
			
			% RenderOcc~\cite{renderocc}  & & Depth \& Semantic & \textbf{53.09} & \textbf{59.97} & \textbf{82.23} \\
			% SimpleOcc~\cite{simpleocc} & nuScense~\cite{nuscenes} & Image & 33.92 & 41.91 & \underline{64.02}  \\
   %              OccNeRF & & Image & \underline{39.20} & \underline{57.20} & 55.47 \\
			\bottomrule
		\end{tabular}
	}
\label{tab:kitti}
\end{table}

\begin{table}[t]
    \centering
    \caption{\textbf{The scene reconstruction performance on the Occ3D-nuScenes dataset~\cite{occ3d}}. The results of other methods are reproduced with their released codes.}
    % \vspace{9pt}
    
	\resizebox{0.4\textwidth}{!}{
	\begin{tabular}{l|c|ccc }
		\toprule
		Method & GT & IoU & Prec. & Rec.\\
        \midrule
        RenderOcc~\cite{renderocc} &\checkmark& 53.09 & 59.97 & 82.23 \\
        \midrule
        SimpleOcc~\cite{simpleocc} &$\times$& 33.92 & 41.91 & 64.02 \\
        OccNeRF &$\times$& 39.20 & 57.20 & 55.47 \\
    \bottomrule
	\end{tabular}}
\label{tab:nus_occ}
\end{table}

\subsection{Occupancy Prediction}
We conduct experiments on semantic occupancy prediction using the Occ3D-nuScenes dataset. The pretrained open-vocabulary model~\cite{sam,dino} struggles with ambiguous prompts like `other' and `other flat', so we exclude them during the evaluation. To compare with the SimpleOcc method~\cite{simpleocc}, we add a semantic head to the original model and leverage the generated 2D semantic labels to train it. Our approach significantly outperforms SimpleOcc, as detailed in TABLE \ref{tab:occ}, and achieves competitive results against some fully-supervised methods. Notably, it excels in predicting `drivable space' and `manmade' classes, outdoing all supervised methods. However, it falls short in detecting small objects, such as bicycles and pedestrians, where it lags behind state-of-the-art supervised methods, likely due to the open-vocabulary model's limitations in capturing small details. 

We further explored the geometry occupancy prediction task with the nuScenes~\cite{nuscenes} and SemanticKITTI~\cite{semantickitti}. Since most works reported in occ3D-nuScenes~\cite{occ3d} do not provide codes, we can only evaluate RenderOcc~\cite{renderocc} and SeimpleOcc~\cite{simpleocc}. As detailed in TABLE~\ref{tab:kitti} and TABLE~\ref{tab:nus_occ}, our approach outperforms other methods supervised by images and achieves competitive results against methods with stronger supervision.

\begin{table}[t]
	\caption{\textbf{The ablation study of supervision method}. `Depth' means whether we use the temporal photometric constraints to train the model. `Multi' indicates whether we employ multi-frame rendering and supervision.}
	\scriptsize
	\centering
	\resizebox{0.48\textwidth}{!}{
		\begin{tabular}{|c|c|c|c|c|}
            \hline
            Depth & Multi &\cellcolor{col1}Abs Rel & \cellcolor{col1}RMSE &  \cellcolor{col2}$\delta < 1.25 $\\
            \hline
        
            & & 0.627& 15.901& 0.051\\
             & \checkmark&   0.489 & 9.352  & 0.362 \\
             \checkmark& & 0.216 & 6.752  & 0.764 \\
             \checkmark& \checkmark & \textbf{0.202} & \textbf{6.697}  & \textbf{0.768} \\
            
            \hline
        \end{tabular}}
\label{tab:ab_supervision}
\end{table}

\subsection{Ablation Study}
\noindent \textbf{Supervision Method:} A straightforward supervision signal is a difference between the rendered and true pixel colours, which is the same as the loss function used in NeRF~\cite{nerf}. However, as shown in TABLE \ref{tab:ab_supervision}, this supervision method yields terrible performance. We attribute this to the challenge NeRF faces in learning the scene structure with only six views. On the contrary, temporal photometric loss (`Depth' in the table) can better leverage geometric cues in adjacent frames, which is the golden metric in self-supervised depth estimation methods. Moreover, multi-frame training provides stronger supervision, further boosting the model's performance.

\noindent \textbf{Coordinate Parameterization:} TABLE \ref{tab:ab_cc} shows the ablation study of coordinate parameterization. Different from occupancy labels, the photometric loss assumes that the images perceive an infinite range. The contracted coordinate aims to represent the unbounded scene in a bounded occupancy. From the table, we can see that the contracted coordinate greatly improves the model's performance. In addition, since the parameterized coordinate is not the Euclidean 3D space, the proposed sampling strategy works better than normal uniform sampling in the original ego coordinate. 

\begin{table}[t]
	\caption{\textbf{The ablation study of coordinate parameterization}. `CC' means whether we adopt contracted coordinates. `Resample' indicates whether we leverage the proposed sampling strategy.}
	\scriptsize
	\centering
	\resizebox{0.48\textwidth}{!}{
		\begin{tabular}{|c|c|c|c|c|}

            \hline
            CC & Resample &\cellcolor{col1}Abs Rel & \cellcolor{col1}RMSE &  \cellcolor{col2}$\delta < 1.25 $\\
            \hline
        
            & &   0.216 &  8.465 & 0.694  \\
             \checkmark&  & 0.208 & 7.339  & 0.743\\
             \checkmark& \checkmark & \textbf{0.202} & \textbf{6.697}  & \textbf{0.768}  \\
            
            \hline
        \end{tabular}}
\label{tab:ab_cc}
\end{table}

\begin{figure}[t]%{r}{0.52\textwidth}
    \centering
    \setlength\tabcolsep{1.0pt} % default value: 6pt
    \renewcommand{\arraystretch}{1.0}
    
    \begin{tabular}{ccc}
    {\includegraphics[width=0.33\linewidth]{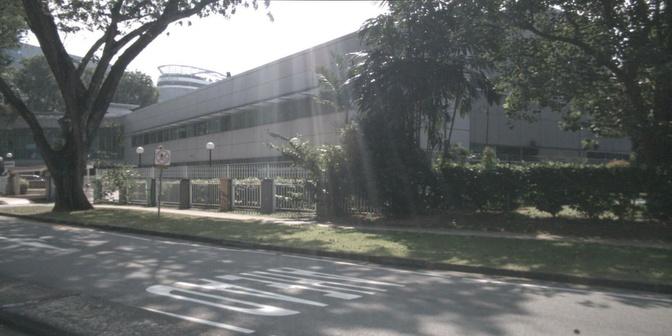}} &
    {\includegraphics[width=0.33\linewidth]{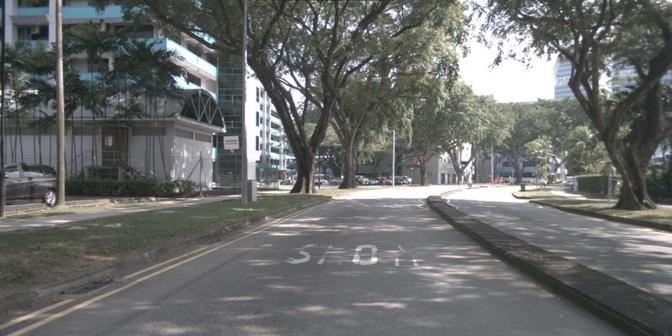}} &
    {\includegraphics[width=0.33\linewidth]{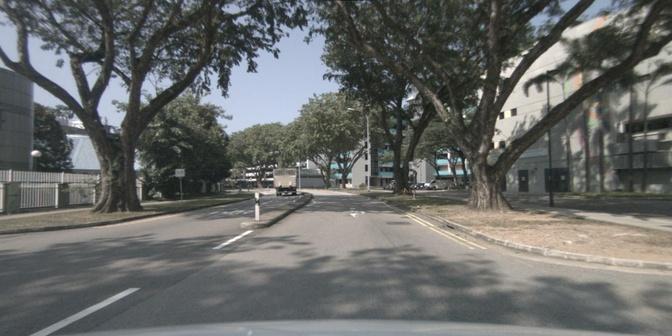}} \\
    
    {\includegraphics[width=0.33\linewidth]{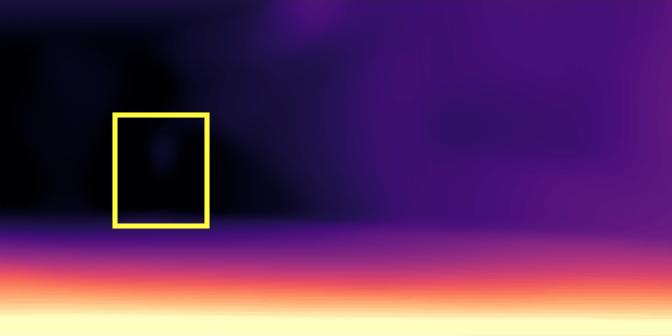}} &
    {\includegraphics[width=0.33\linewidth]{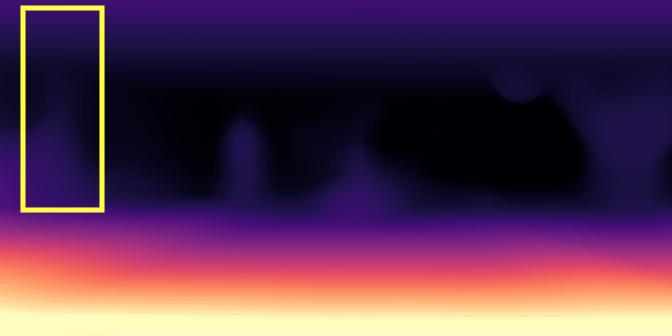}} &
    {\includegraphics[width=0.33\linewidth]{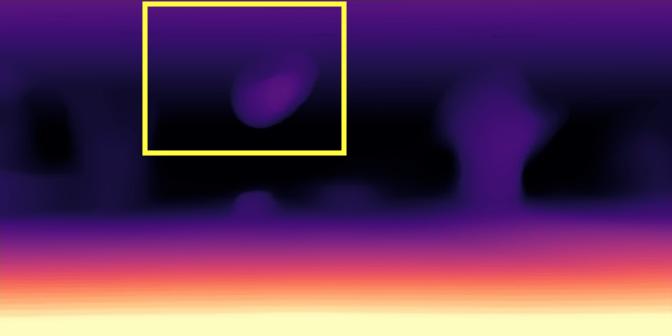}} \\
    
    {\includegraphics[width=0.33\linewidth]{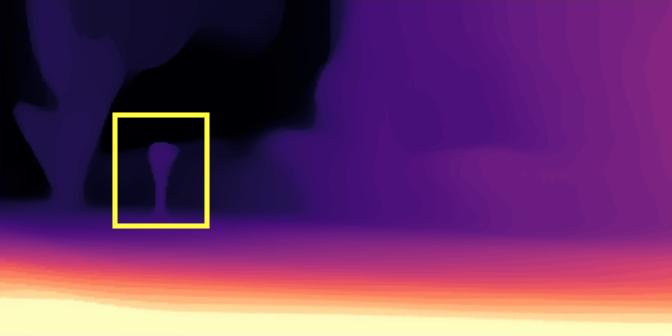}} &
    {\includegraphics[width=0.33\linewidth]{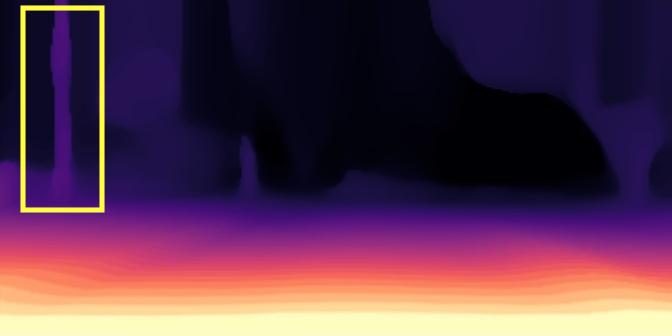}} &
    {\includegraphics[width=0.33\linewidth]{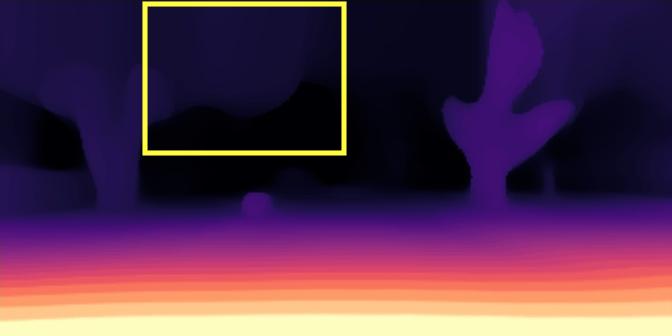}} \\
    \end{tabular}
    \caption{\textbf{Qualitative comparison of different coordinates.} The second line indicates the results without using coordinate parameterization. With the ability to represent unbounded environments, our method can get better results in far scenes, such as the sky.}
    % \vspace{-20pt}
    \label{fig:ablation_cc}
\end{figure}

\begin{table}[t]
    \centering
    \caption{\textbf{The ablation study of semantic label generation}. `SAM logits' means that we directly use the logits from SAM~\cite{sam}. 'Catagory names' means that we do not conduct the prompting strategies.}
    \resizebox{0.48\textwidth}{!}{
    \begin{tabular}{l|c|c|c }
        \toprule
        Method & SAM logits & Catagory names & Ours \\
        \midrule
        mIoU & 7.50 & 8.23 & 10.81 \\
        \bottomrule
    \end{tabular}}
\label{tab:ab_occ}
\end{table}

\begin{figure}[t]%{r}{0.6\textwidth}
    \centering
    \scriptsize
    \begin{subfigure}{0.48\linewidth}
        \includegraphics[width=\textwidth]{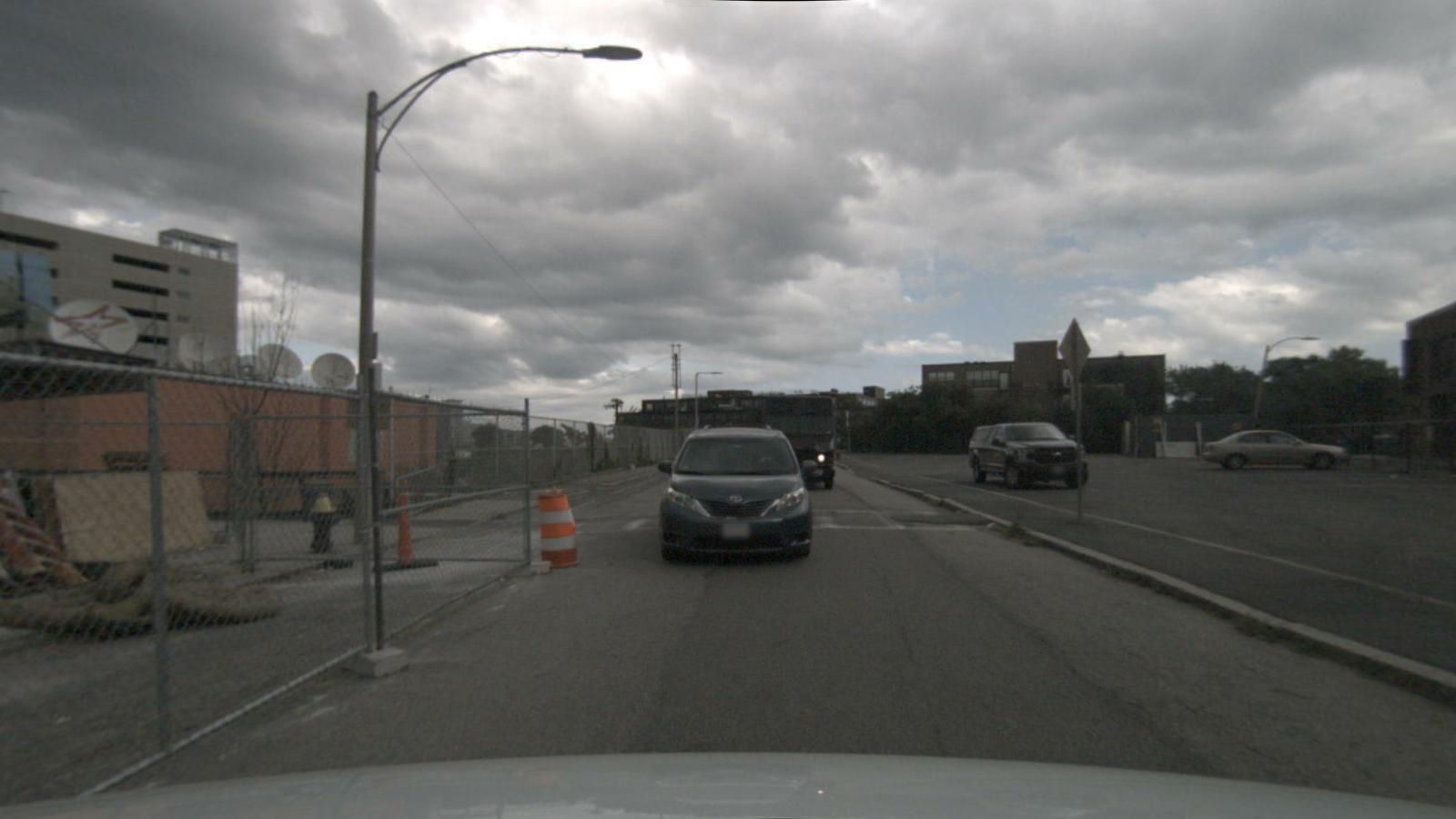}
        \caption{RGB image}
    \end{subfigure}
    \begin{subfigure}{0.48\linewidth}
        \includegraphics[width=\textwidth]{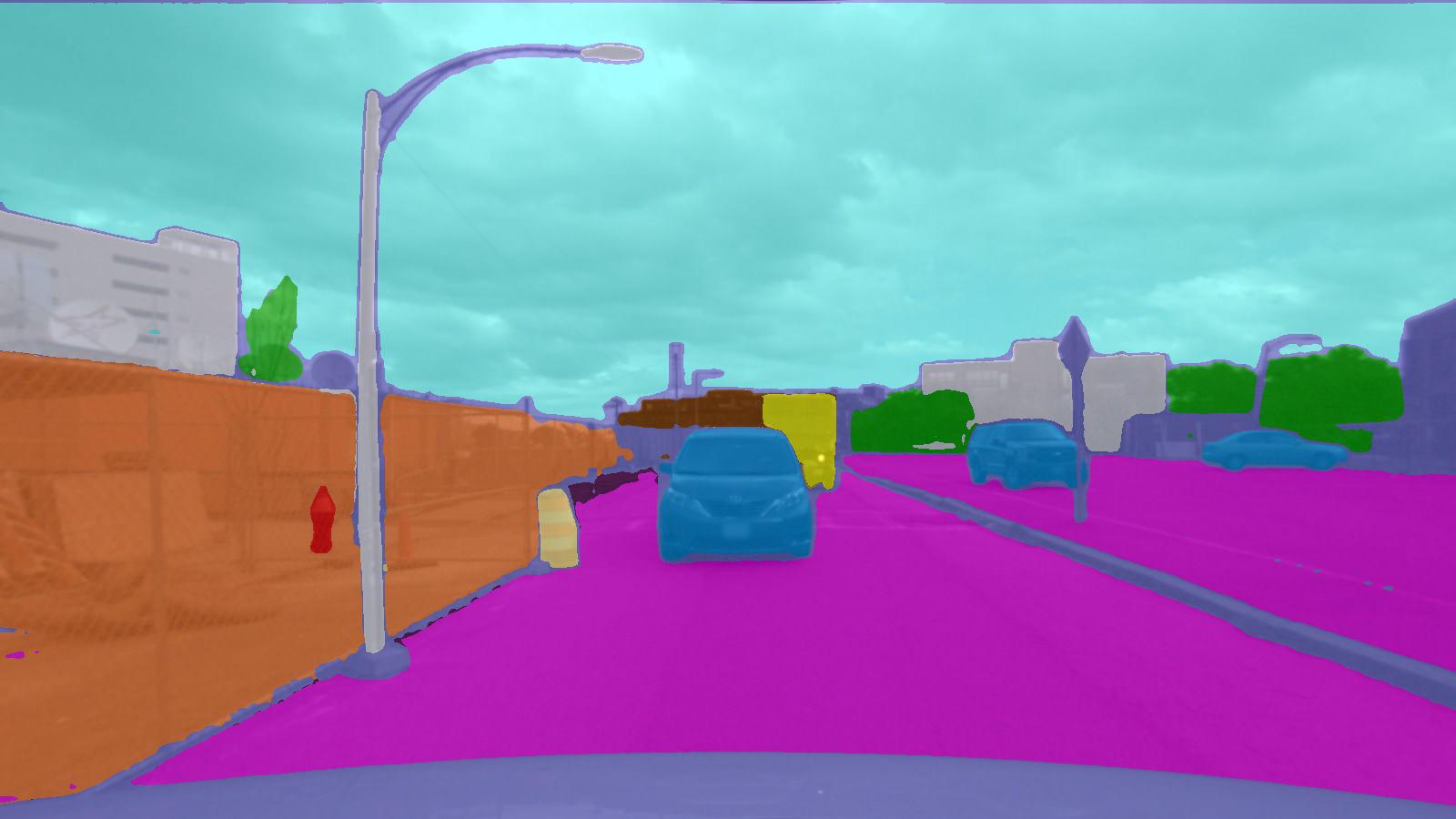}
        \caption{Our Grounded-SAM labels}
    \end{subfigure} 
    \begin{subfigure}{0.48\linewidth}
        \includegraphics[width=\textwidth]{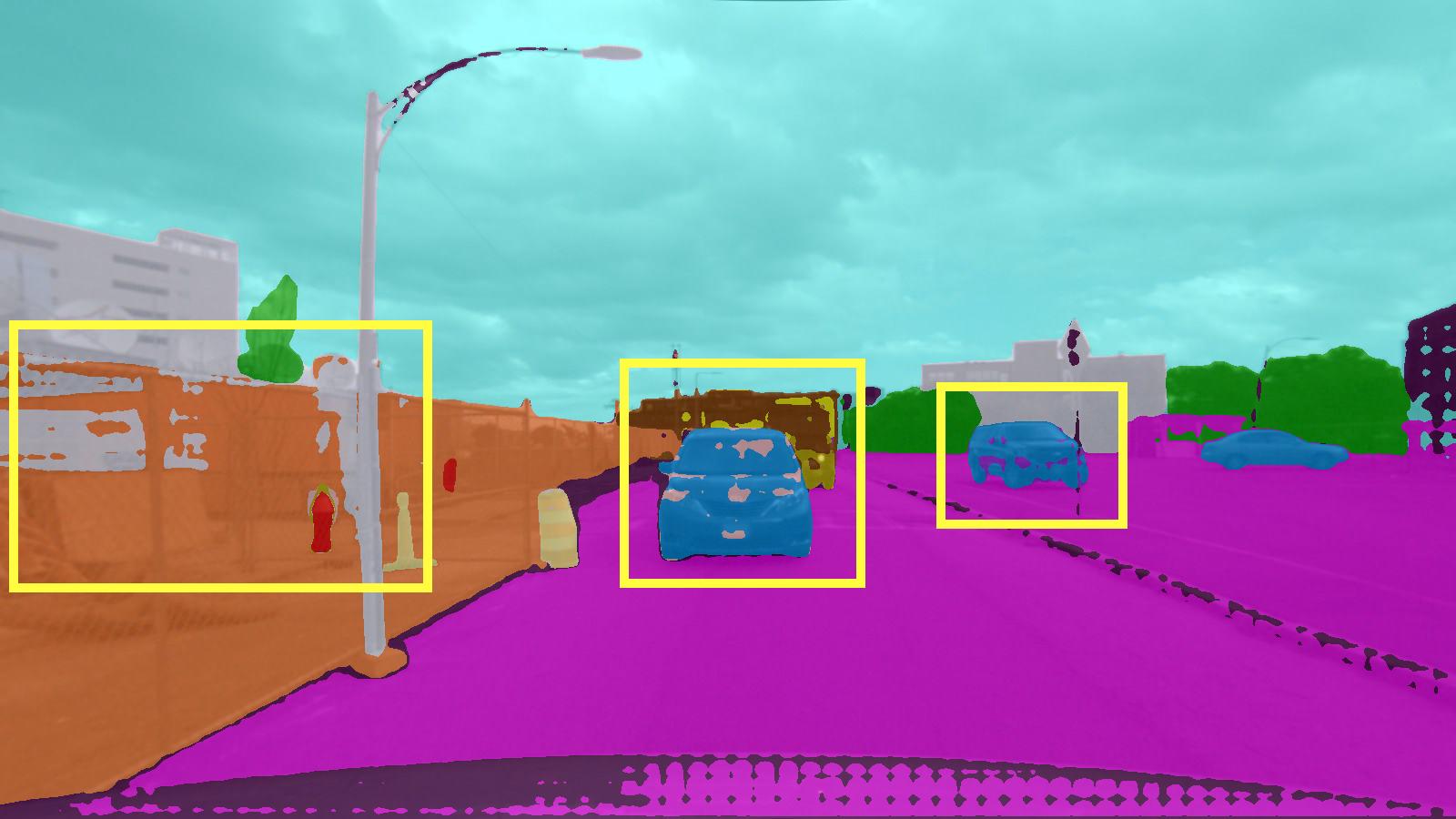}
        \caption{Labels with SAM logits}
    \end{subfigure}
    \begin{subfigure}{0.48\linewidth}
        \includegraphics[width=\textwidth]{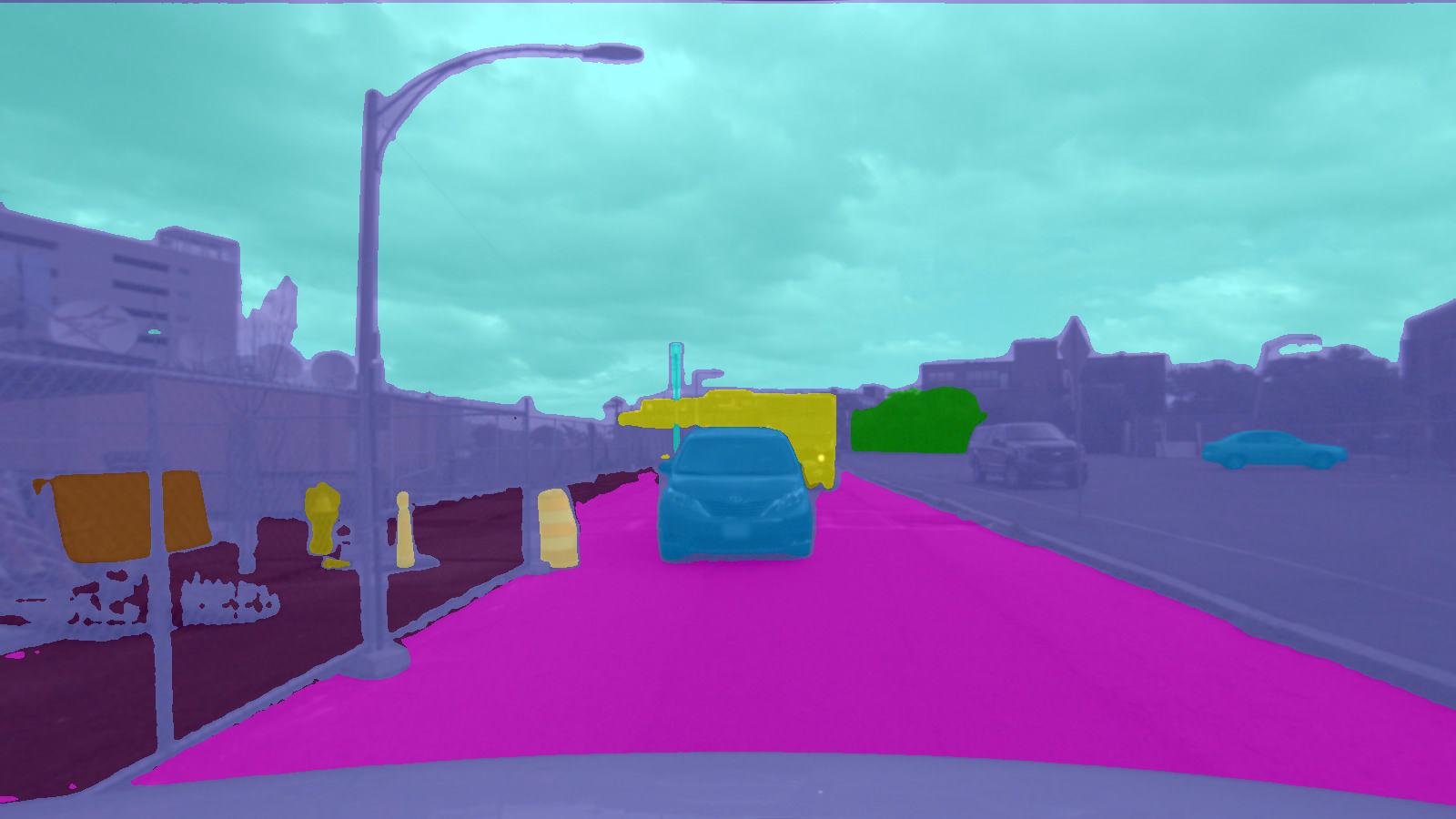}
        \caption{Labels w/o prompt strategies}
    \end{subfigure}
    \caption{\textbf{Comparison of different semantic label generation methods}. Compared with generating semantic labels with SAM logits or feeding raw category names, our semantic labels are preciser and have better continuity.
    } 
\label{fig:ablation_sam_category}
\end{figure}

\noindent \textbf{Semantic Label Generation:} In this subsection, we conduct ablation studies of semantic label generation on the nuScenes~\cite{nuscenes} dataset. First, we change grounding DINO~\cite{dino} logits as SAM logits~\cite{sam} to get semantic labels. As shown in TABLE \ref{tab:ab_occ} and Fig.~\ref{fig:ablation_sam_category}, we find that the SAM logits are more noisy and discontinuous. Then, we also feed raw category names to the open-vocabulary model without proposed prompting strategies. However, this method leads to worse results since the original class names cannot provide fine-grained semantic guidance and bring ambiguity.

\subsection{Visualization}

To further demonstrate the superiority of our method, we provide some qualitative results in Fig.~\ref{fig:qualitative} and \ref{fig:semantic}. From Fig.~\ref{fig:qualitative}, we can see that our method can generate high-quality depth maps and occupancy with fine-grained details. See supplementary material for more qualitative comparisons with other methods. For semantic occupancy prediction, as shown in Fig.~\ref{fig:semantic}, our OccNeRF can reconstruct dense results of the surrounding scenes, especially for the large-area categories, such as `drivable space' and `manmade'. 

\subsection{Analysis}

\begin{figure}[tb]
\centering
% \scriptsize
\resizebox{0.4\textwidth}{!}{
    \begin{tikzpicture}
    \begin{axis}[
        xlabel={\% of 3D labels},
        ylabel={mIoU},
        xmin=0, xmax=100,
        ymin=10, ymax=26,
        xtick={0,25,50,75,100},
        ytick={10,15,20,25},
        legend pos=south east,
        ymajorgrids=true,
        grid style=dashed,
        nodes near coords, % This enables the display of the mIoU values
        point meta=explicit symbolic, % Use the meta data for the label
        every node near coord/.append style={
            xshift=6mm,
            yshift=-4mm,
        }
    ]
    
    % With pretraining
    \addplot[
        color=blue,
        mark=square,
        ]
        coordinates {
        (0,10.81) [10.81]
        (25,21.21) [21.21]
        (50,22.82) [22.82]
        (75,24.03) [24.03]
        (100,24.88) [24.88]
        };
        \addlegendentry{SFT}
    
    % Without pretraining
    \addplot[
        color=red,
        mark=*,
        ]
        coordinates {
        (100,23.47) [23.47]
        };
        \addlegendentry{SCR}
    
    \end{axis}
    \end{tikzpicture}
}
    \caption{\textbf{Supervised fine-tuning experiment.} `SFT' stands for fine-tuning the pretrained model with 3D labels. `SCR' means training from scratch.}
    % \vspace{-20pt}
\label{fig:sft}
\end{figure}
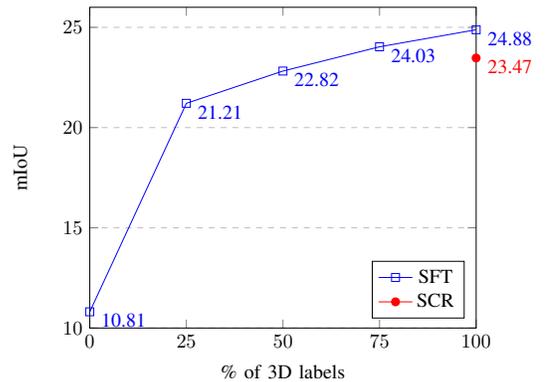

\noindent
\textbf{Supervised fine-tuning (SFT).} A large success has been achieved in the field of natural language processing by utilizing a methodology combining self-supervised pre-training and supervised fine-tuning. Our work makes it possible to extend this mothodology to 3D occupancy prediction. We explored supervised fine-tuning with our model using 3D occupancy labels from Occ3D~\cite{occ3d}. Fig.~\ref{fig:sft} demonstrates that integrating a fraction of the 3D ground truth significantly boosts performance, nearing that of fully supervised models. Complete fine-tuning with all 3D ground truth data even surpasses the non-pretrained model, highlighting our method's data efficiency.

\begin{table}[htb]
    \centering
    \resizebox{0.45\textwidth}{!}{
        \begin{tabular}{l|cc}
        \toprule
            Method & Latency (s) & Memory (G)\\ 
        \midrule
            BEVFormer~\cite{bevformer} & 0.28 & \textbf{4.5}\\
            TPVFormer~\cite{tpvformer} & 0.29 & 5.1\\
            MonoScene~\cite{monoscene} & 0.77 & 20.3\\
            SimpleOcc~\cite{simpleocc} & \textbf{0.16} & 8.7\\
            SurroundOcc~\cite{surroundocc} &0.31 & 5.9\\ 
        \midrule
            Ours &0.18 & 11.0\\
        \bottomrule
        \end{tabular}
    }
    \caption{\textbf{Computational costs.} The results are measured on a single NVIDIA A100.}
\label{tab:comp_cost}
\end{table}

\noindent
\textbf{Computational cost.} The inference latency and memory requirements for our approach are detailed in TABLE~\ref{tab:comp_cost}. It is important to note that our method does not require rendering during inference. Our method achieves a comparable efficiency with convolution-based methods like SimpleOcc~\cite{simpleocc} and is significantly faster than transformer-based methods like SurroundOcc~\cite{surroundocc}. However, volume rendering will increase training time. For an epoch on 8 A100 GPUs, our method takes 128 mins while SurroundOcc takes 75 mins.

\section{Conclusion}
\label{sec:conclusion}

In this paper, we have introduced OccNeRF, a novel approach to train multi-camera occupancy networks without relying on 3D supervision. By leveraging temporal photometric constraints and pretrained open-vocabulary segmentation models, OccNeRF has effectively addressed the challenges of sparse surrounding views and the representation of unbounded scenes. Our method has offered a cost-effective and scalable solution for autonomous driving by eliminating the dependency on expensive LiDAR data and utilizing vast amounts of unlabeled multi-camera image data, similar to the successful self-supervised method used in natural language processing.

Our experiments on the nuScenes and SemanticKITTI datasets have demonstrated that OccNeRF significantly outperforms existing methods in self-supervised multi-camera depth estimation, providing accurate and consistent depth predictions across multiple views. In semantic occupancy prediction, OccNeRF has achieved competitive performance with fully-supervised methods, excelling in large-area categories such as `drivable space' and `manmade' structures. However, its performance in detecting smaller objects like bicycles and pedestrians has been limited by the capabilities of open-vocabulary segmentation models in capturing fine details.

While promising, OccNeRF has some limitations, including its current inability to predict dynamic occupancy flows due to the lack of multi-frame information during inference. Future work could address this by incorporating optical flow models and multi-frame inputs. Additionally, improving the granularity and accuracy of open-vocabulary segmentation models could further enhance our system's detection and representation of smaller objects. Overall, OccNeRF represents a significant advancement in vision-centric 3D scene understanding for autonomous driving, offering a robust and efficient approach to developing autonomous systems.

\begin{figure*}[tb]
\setlength\tabcolsep{0.2 pt}
\renewcommand{\arraystretch}{0.2}
\centering

\begin{tabular}{lcccccc}

& F.Left & Front & F.Right & B.Right & Back & B.Left \\

\rotatebox[origin=c]{90}{Input} &
\begin{tabular}{l}\includegraphics[width=0.15\linewidth]{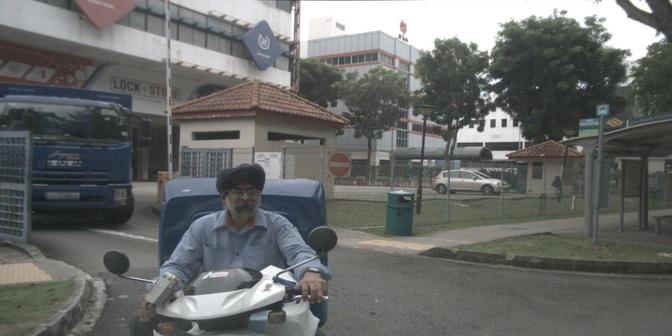}\end{tabular} &
\begin{tabular}{l}\includegraphics[width=0.15\linewidth]{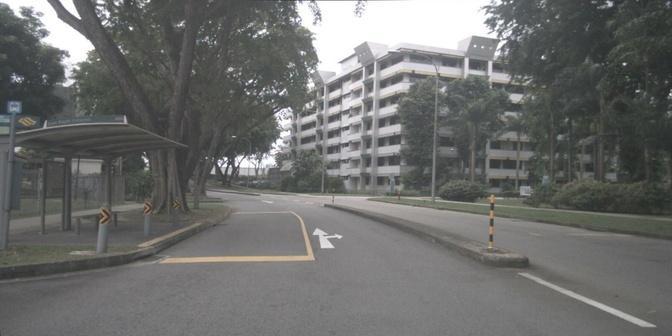}\end{tabular} &
\begin{tabular}{l}\includegraphics[width=0.15\linewidth]{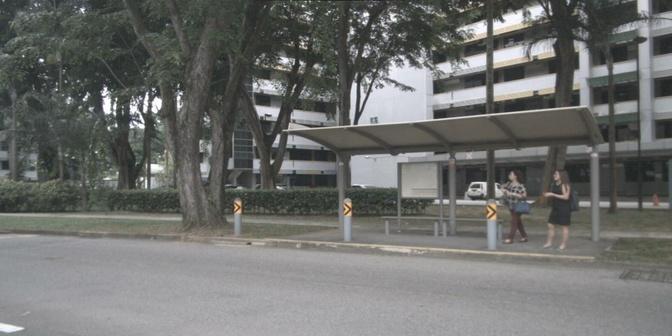}\end{tabular} &
\begin{tabular}{l}\includegraphics[width=0.15\linewidth]{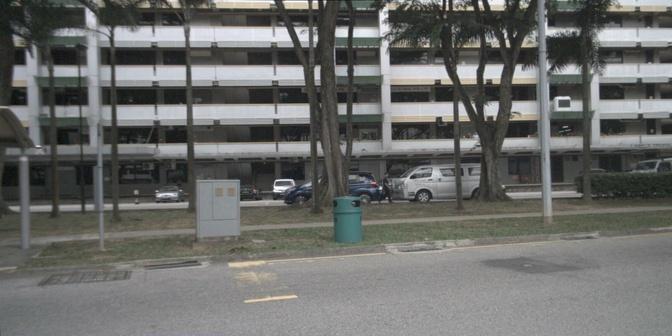}\end{tabular} &
\begin{tabular}{l}\includegraphics[width=0.15\linewidth]{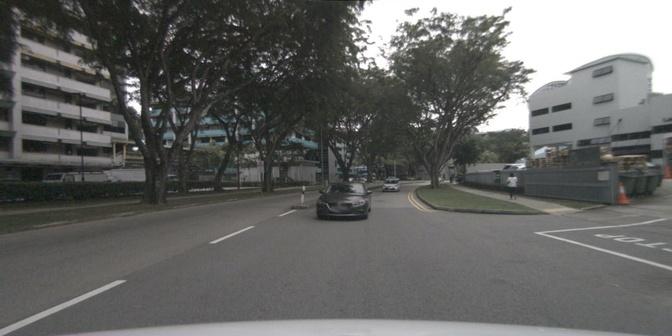}\end{tabular} &
\begin{tabular}{l}\includegraphics[width=0.15\linewidth]{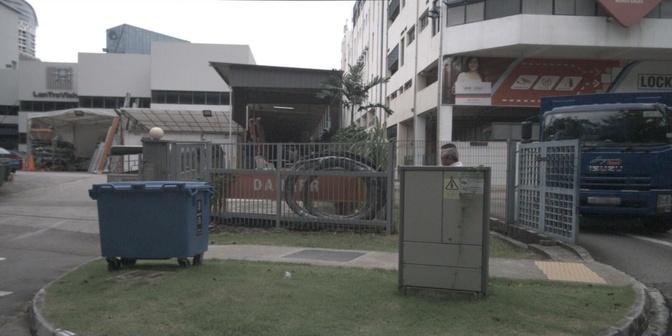}\end{tabular} \\

\rotatebox[origin=c]{90}{SD~\cite{surrounddepth}} &
\begin{tabular}{l}\includegraphics[width=0.15\linewidth]{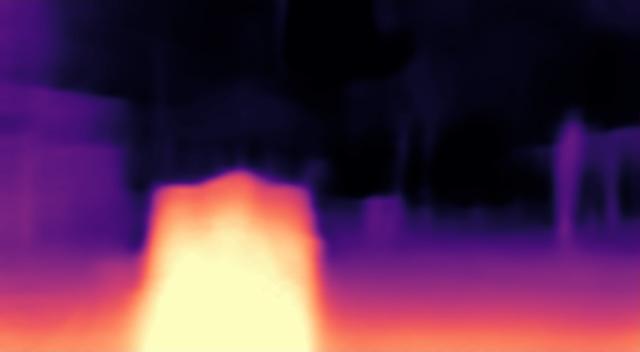}\end{tabular} &
\begin{tabular}{l}\includegraphics[width=0.15\linewidth]{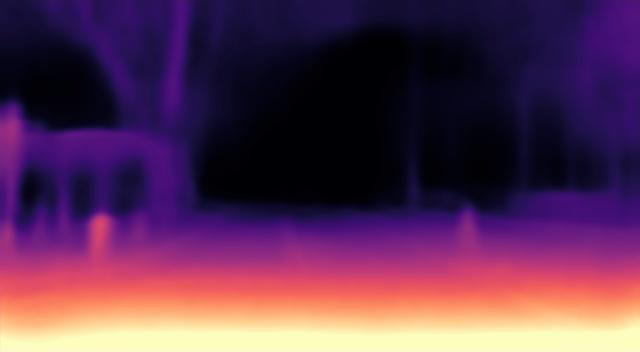}\end{tabular} &
\begin{tabular}{l}\includegraphics[width=0.15\linewidth]{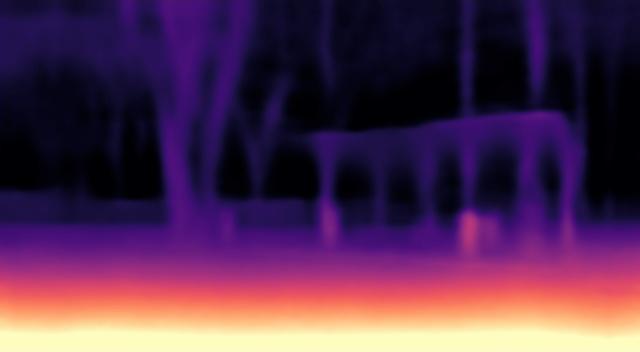}\end{tabular} &
\begin{tabular}{l}\includegraphics[width=0.15\linewidth]{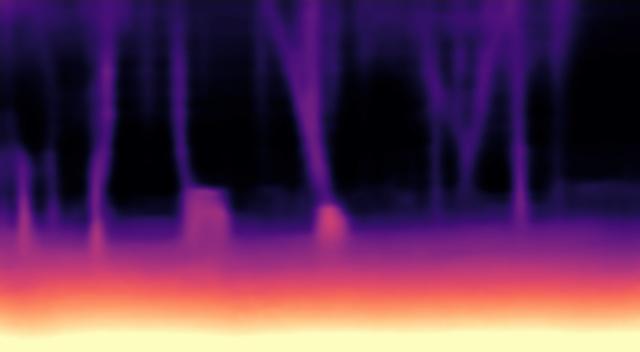}\end{tabular} &
\begin{tabular}{l}\includegraphics[width=0.15\linewidth]{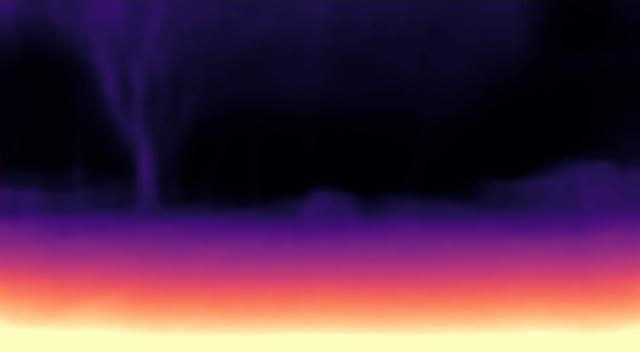}\end{tabular} &
\begin{tabular}{l}\includegraphics[width=0.15\linewidth]{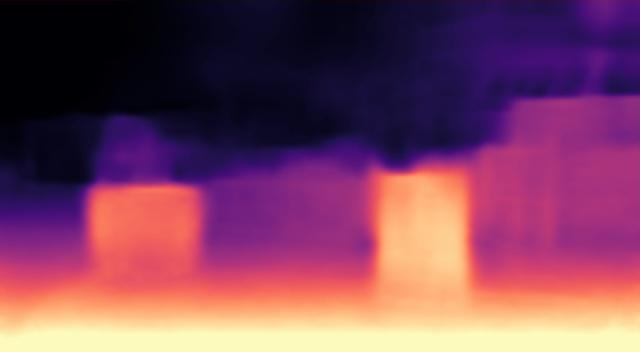}\end{tabular} \\

\rotatebox[origin=c]{90}{R3D3~\cite{r3d3}} &
\begin{tabular}{l}\includegraphics[width=0.15\linewidth]{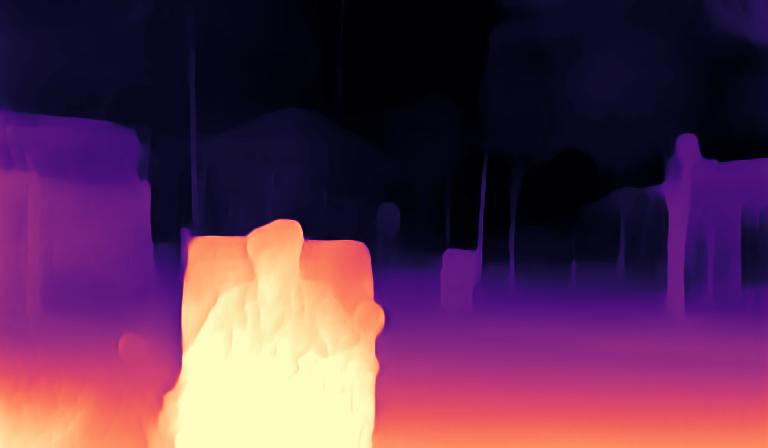}\end{tabular} &
\begin{tabular}{l}\includegraphics[width=0.15\linewidth]{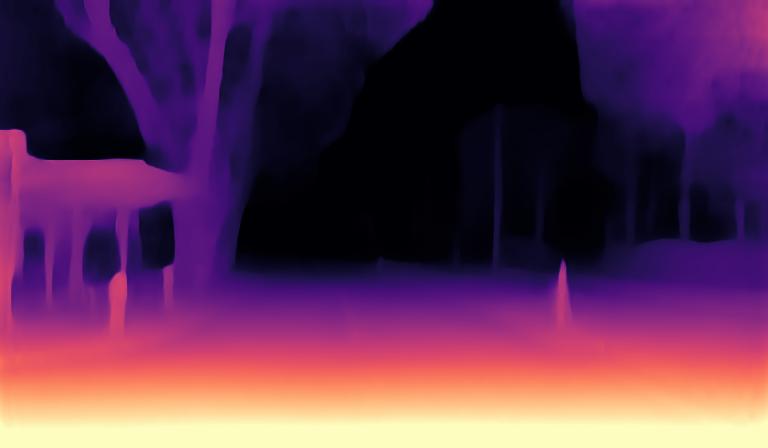}\end{tabular} &
\begin{tabular}{l}\includegraphics[width=0.15\linewidth]{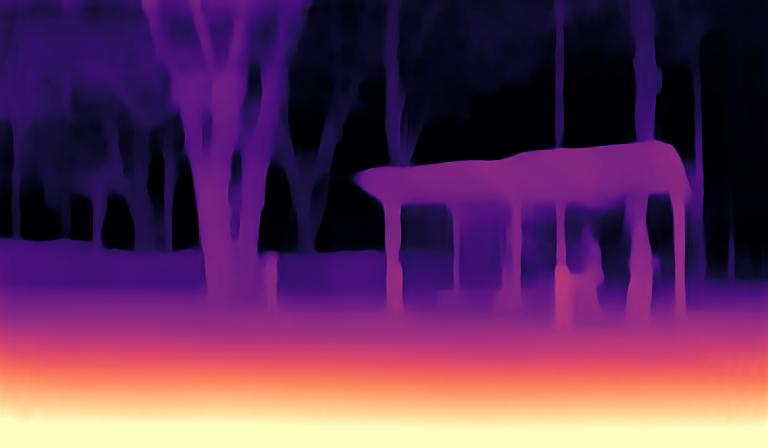}\end{tabular} &
\begin{tabular}{l}\includegraphics[width=0.15\linewidth]{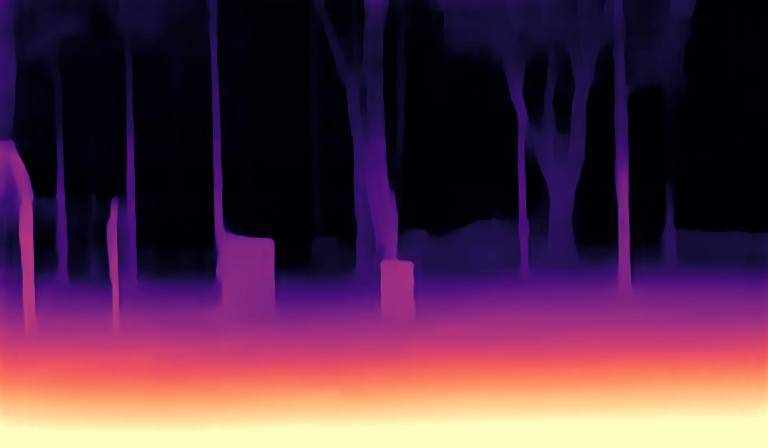}\end{tabular} &
\begin{tabular}{l}\includegraphics[width=0.15\linewidth]{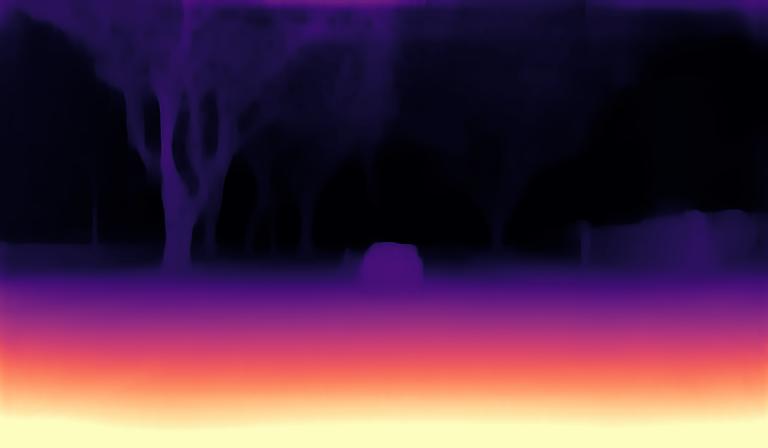}\end{tabular} &
\begin{tabular}{l}\includegraphics[width=0.15\linewidth]{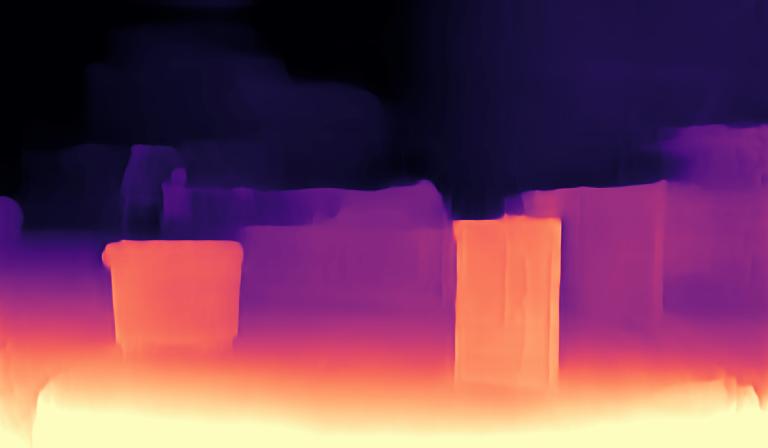}\end{tabular} \\

\rotatebox[origin=c]{90}{SO~\cite{simpleocc}} &
\begin{tabular}{l}\includegraphics[width=0.15\linewidth]{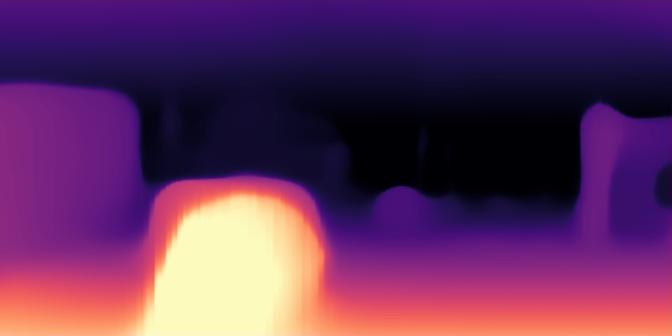}\end{tabular} &
\begin{tabular}{l}\includegraphics[width=0.15\linewidth]{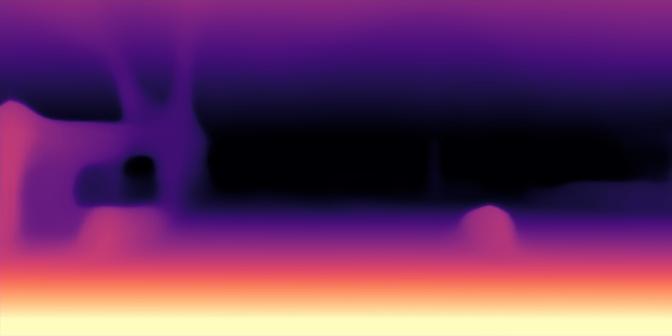}\end{tabular} &
\begin{tabular}{l}\includegraphics[width=0.15\linewidth]{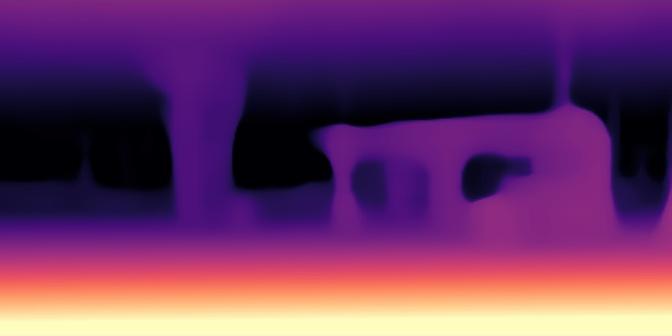}\end{tabular} &
\begin{tabular}{l}\includegraphics[width=0.15\linewidth]{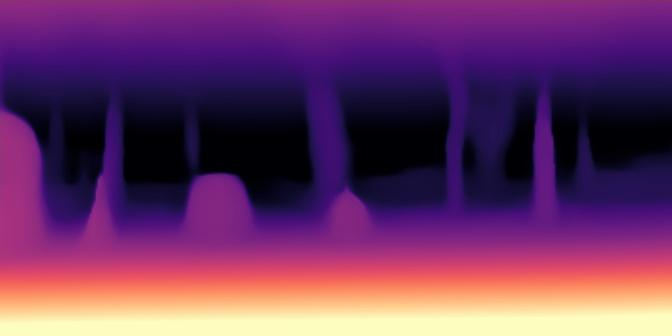}\end{tabular} &
\begin{tabular}{l}\includegraphics[width=0.15\linewidth]{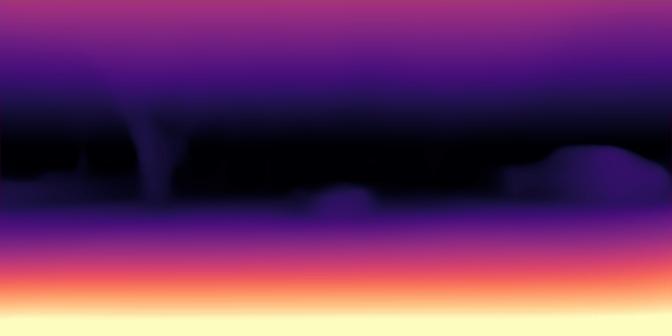}\end{tabular} &
\begin{tabular}{l}\includegraphics[width=0.15\linewidth]{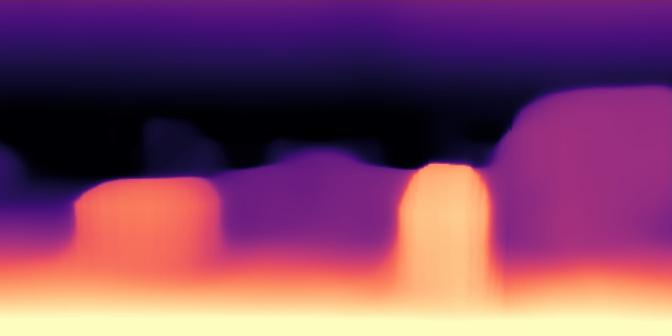}\end{tabular} \\

\rotatebox[origin=c]{90}{Ours} &
\begin{tabular}{l}\includegraphics[width=0.15\linewidth]{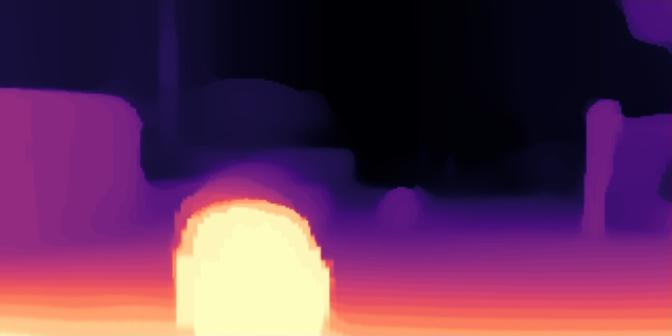}\end{tabular} &
\begin{tabular}{l}\includegraphics[width=0.15\linewidth]{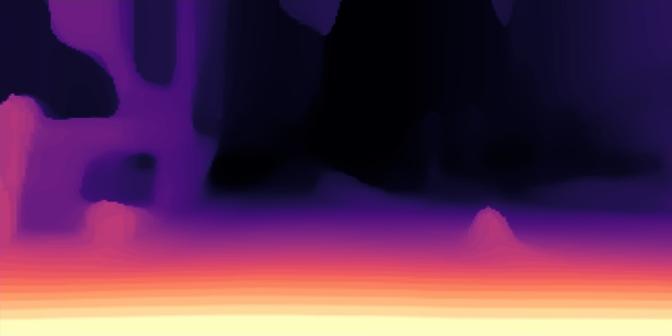}\end{tabular} &
\begin{tabular}{l}\includegraphics[width=0.15\linewidth]{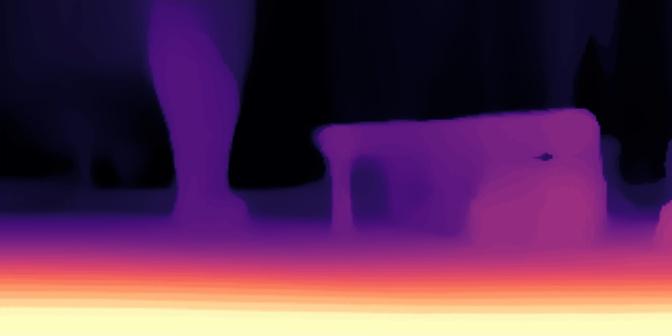}\end{tabular} &
\begin{tabular}{l}\includegraphics[width=0.15\linewidth]{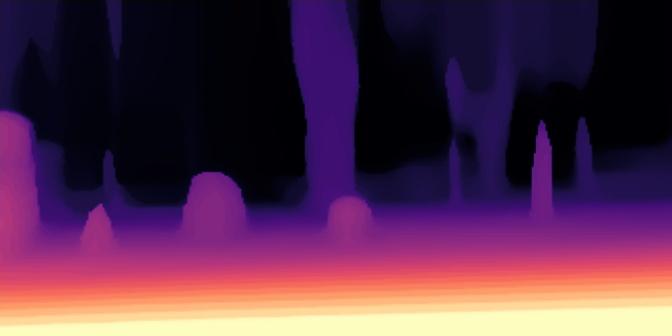}\end{tabular} &
\begin{tabular}{l}\includegraphics[width=0.15\linewidth]{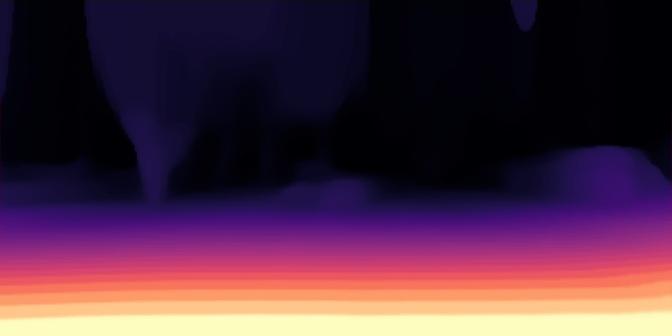}\end{tabular} &
\begin{tabular}{l}\includegraphics[width=0.15\linewidth]{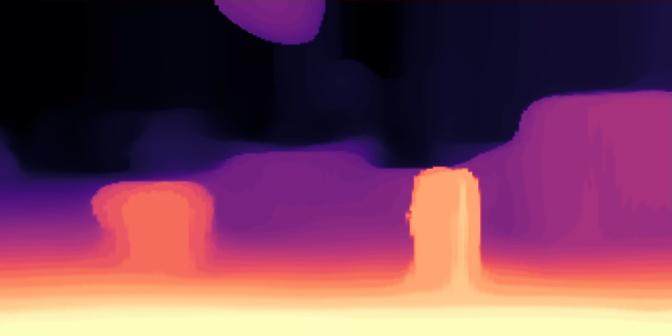}\end{tabular} \\

\midrule

\rotatebox[origin=c]{90}{Input} &
\begin{tabular}{l}\includegraphics[width=0.15\linewidth]{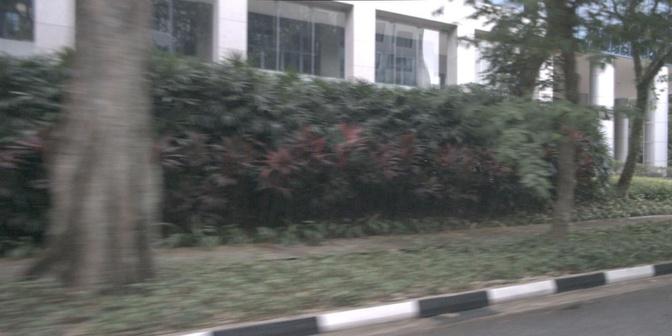}\end{tabular} &
\begin{tabular}{l}\includegraphics[width=0.15\linewidth]{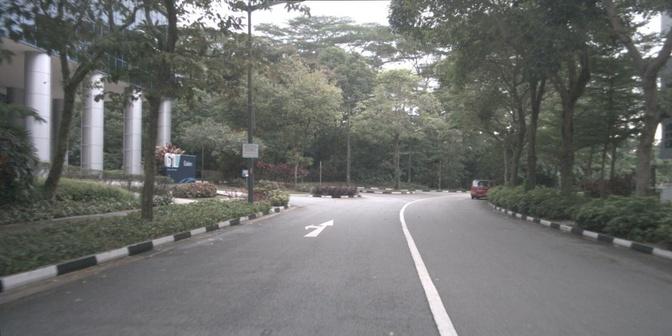}\end{tabular} &
\begin{tabular}{l}\includegraphics[width=0.15\linewidth]{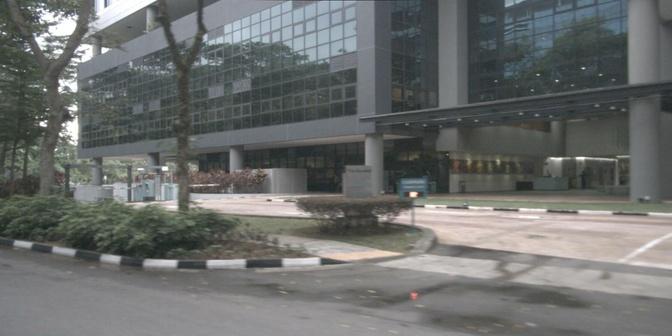}\end{tabular} &
\begin{tabular}{l}\includegraphics[width=0.15\linewidth]{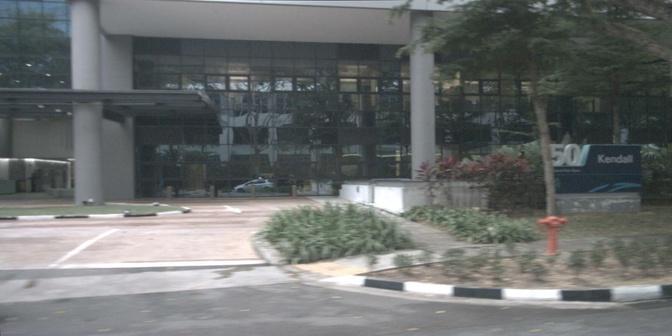}\end{tabular} &
\begin{tabular}{l}\includegraphics[width=0.15\linewidth]{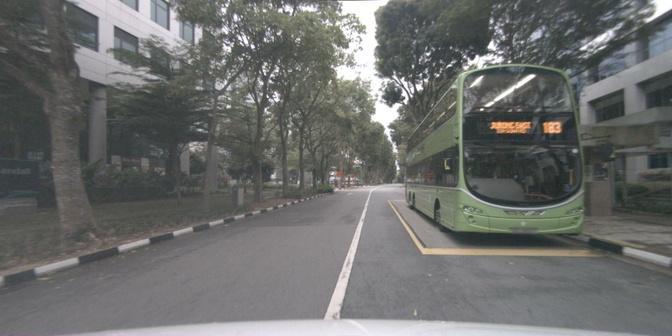}\end{tabular} &
\begin{tabular}{l}\includegraphics[width=0.15\linewidth]{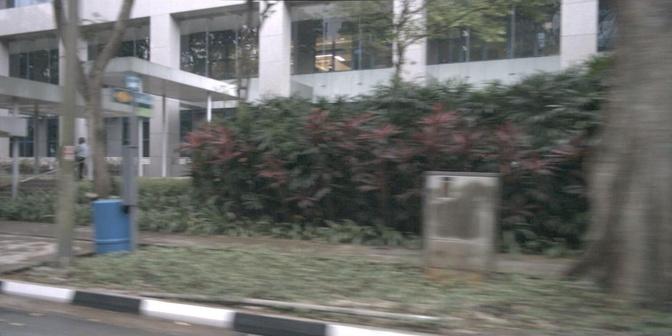}\end{tabular} \\

\rotatebox[origin=c]{90}{SD~\cite{surrounddepth}} &
\begin{tabular}{l}\includegraphics[width=0.15\linewidth]{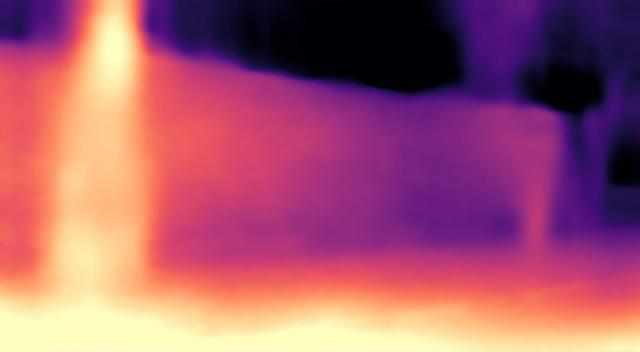}\end{tabular} &
\begin{tabular}{l}\includegraphics[width=0.15\linewidth]{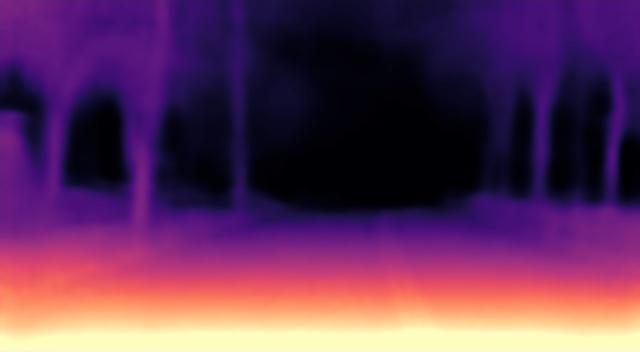}\end{tabular} &
\begin{tabular}{l}\includegraphics[width=0.15\linewidth]{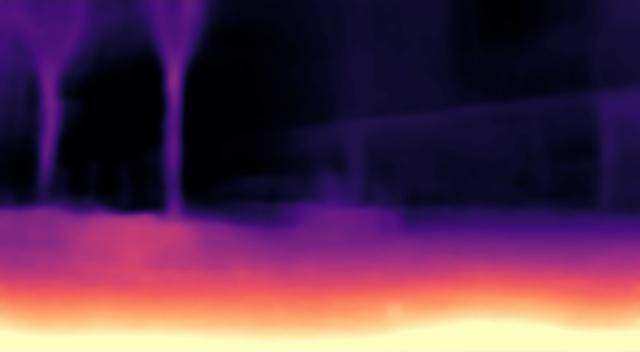}\end{tabular} &
\begin{tabular}{l}\includegraphics[width=0.15\linewidth]{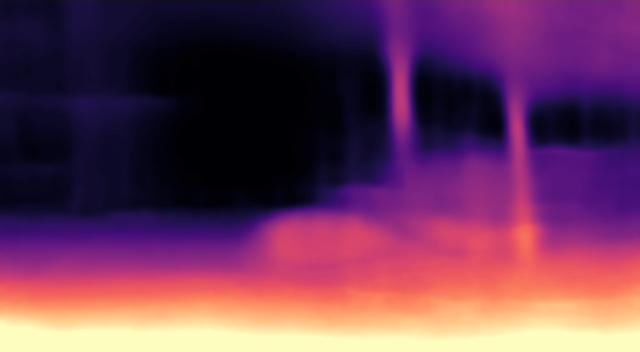}\end{tabular} &
\begin{tabular}{l}\includegraphics[width=0.15\linewidth]{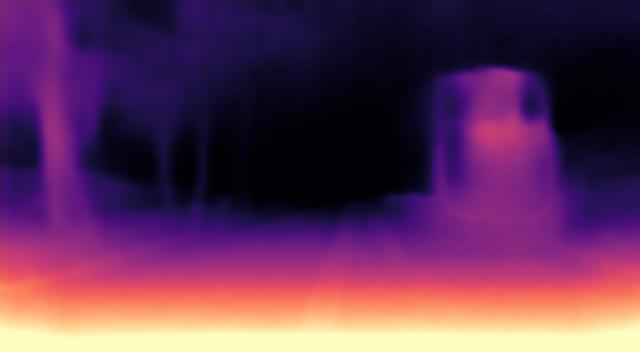}\end{tabular} &
\begin{tabular}{l}\includegraphics[width=0.15\linewidth]{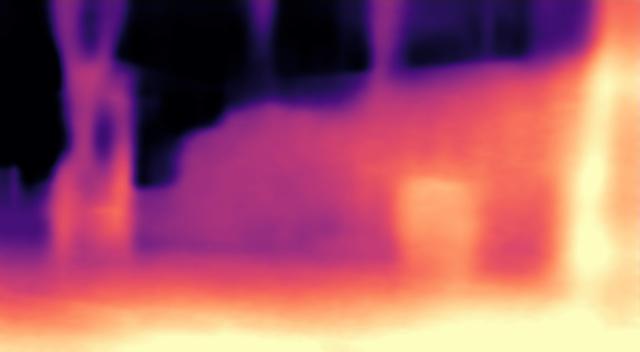}\end{tabular} \\

\rotatebox[origin=c]{90}{R3D3~\cite{r3d3}} &
\begin{tabular}{l}\includegraphics[width=0.15\linewidth]{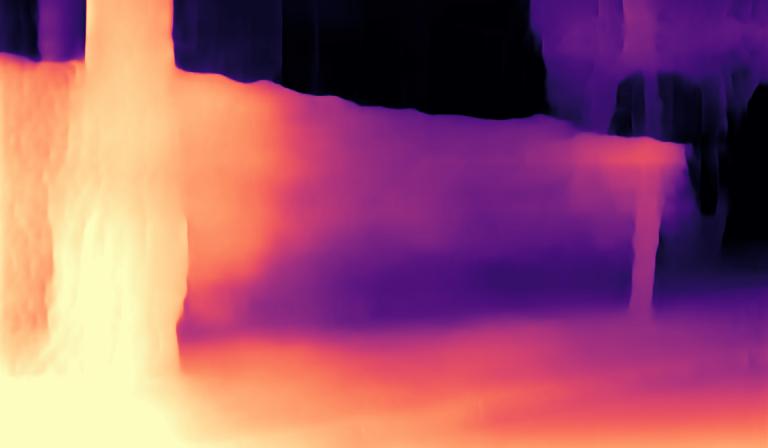}\end{tabular} &
\begin{tabular}{l}\includegraphics[width=0.15\linewidth]{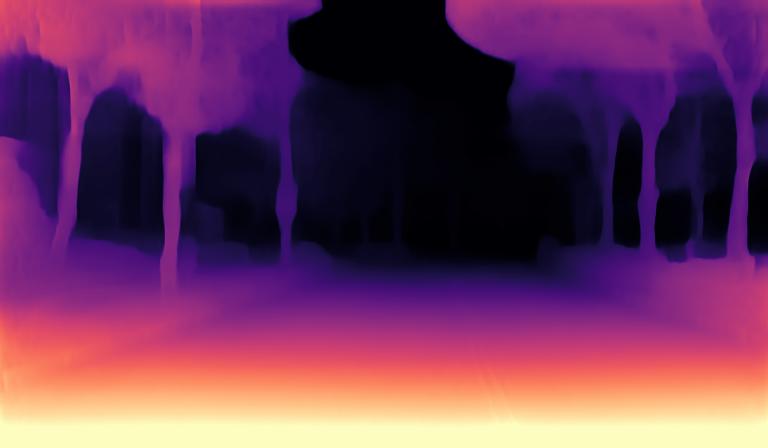}\end{tabular} &
\begin{tabular}{l}\includegraphics[width=0.15\linewidth]{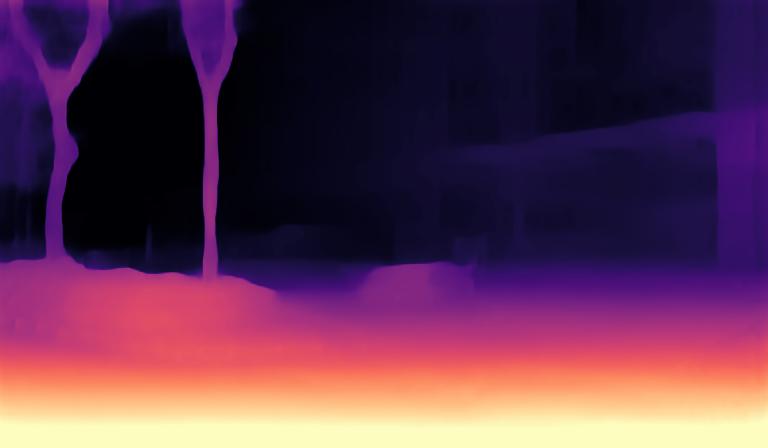}\end{tabular} &
\begin{tabular}{l}\includegraphics[width=0.15\linewidth]{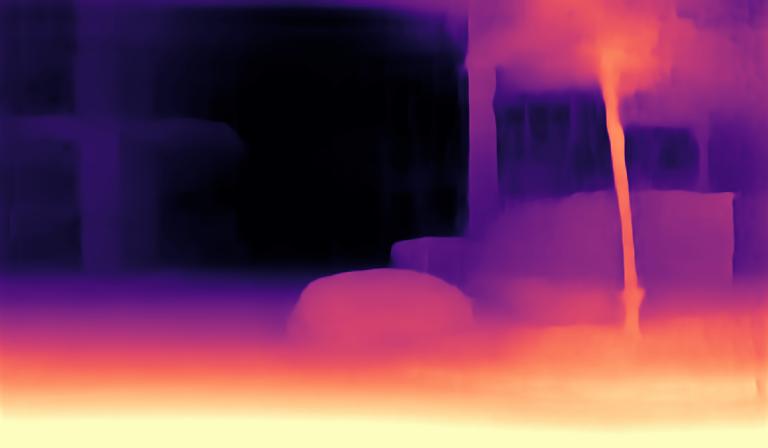}\end{tabular} &
\begin{tabular}{l}\includegraphics[width=0.15\linewidth]{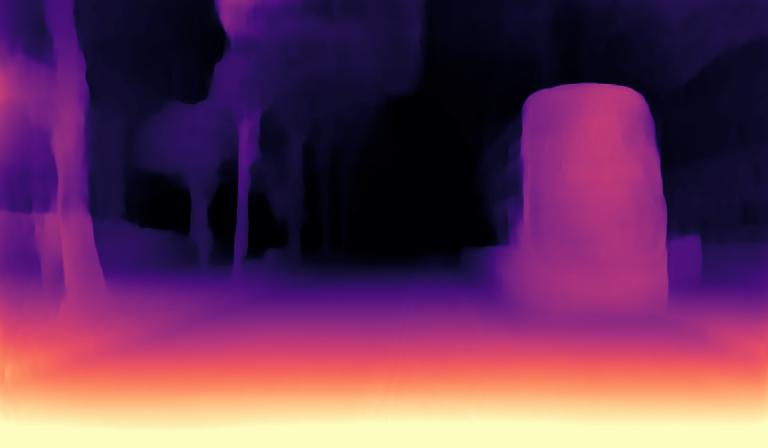}\end{tabular} &
\begin{tabular}{l}\includegraphics[width=0.15\linewidth]{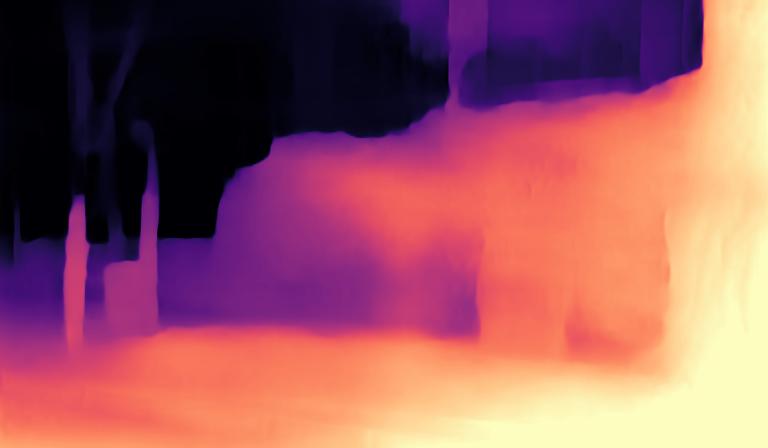}\end{tabular} \\

\rotatebox[origin=c]{90}{SO~\cite{simpleocc}} &
\begin{tabular}{l}\includegraphics[width=0.15\linewidth]{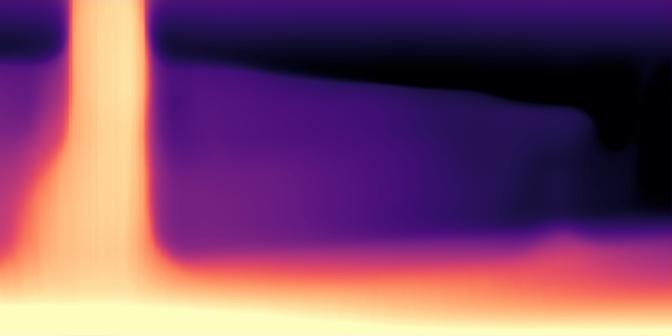}\end{tabular} &
\begin{tabular}{l}\includegraphics[width=0.15\linewidth]{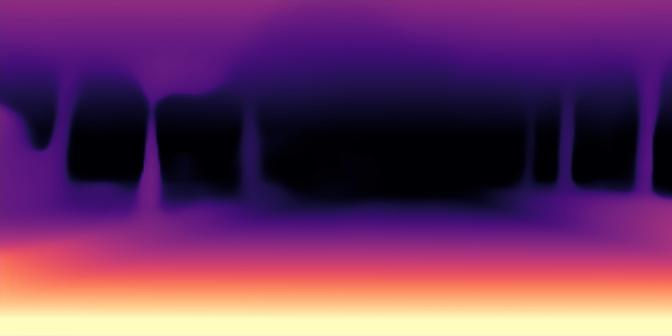}\end{tabular} &
\begin{tabular}{l}\includegraphics[width=0.15\linewidth]{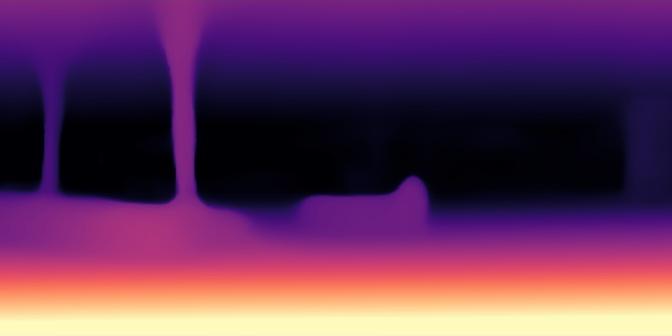}\end{tabular} &
\begin{tabular}{l}\includegraphics[width=0.15\linewidth]{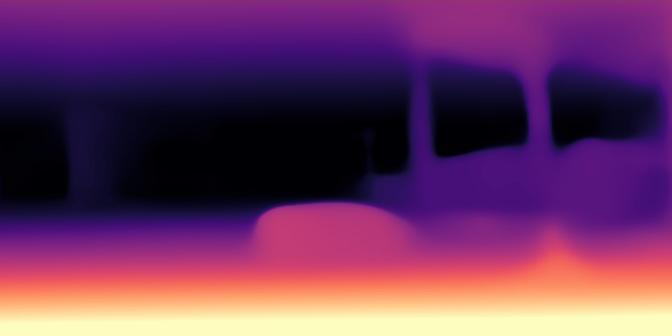}\end{tabular} &
\begin{tabular}{l}\includegraphics[width=0.15\linewidth]{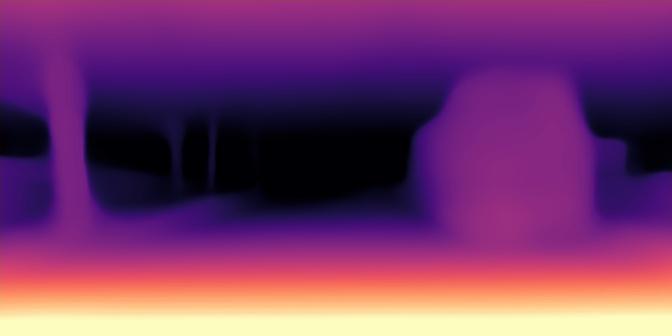}\end{tabular} &
\begin{tabular}{l}\includegraphics[width=0.15\linewidth]{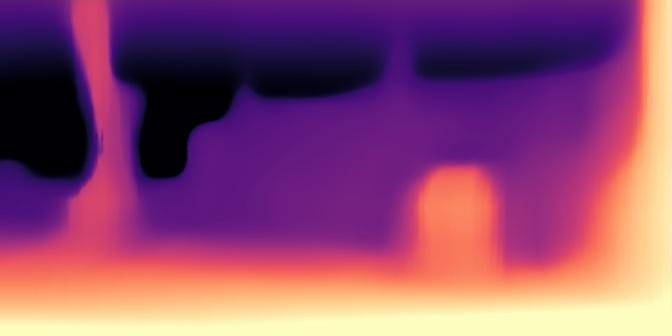}\end{tabular} \\

\rotatebox[origin=c]{90}{Ours} &
\begin{tabular}{l}\includegraphics[width=0.15\linewidth]{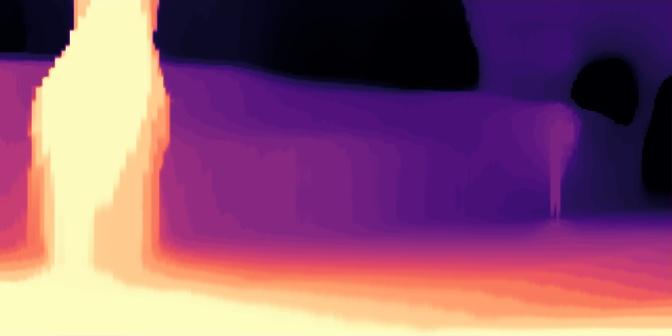}\end{tabular} &
\begin{tabular}{l}\includegraphics[width=0.15\linewidth]{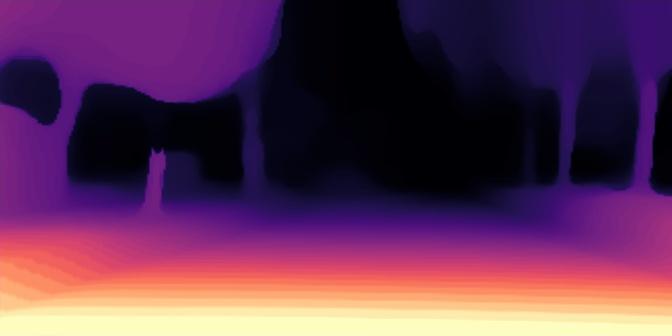}\end{tabular} &
\begin{tabular}{l}\includegraphics[width=0.15\linewidth]{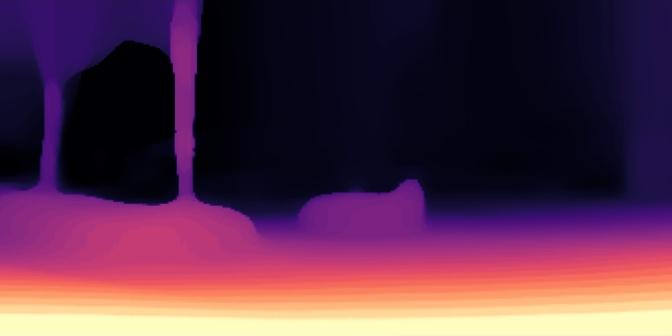}\end{tabular} &
\begin{tabular}{l}\includegraphics[width=0.15\linewidth]{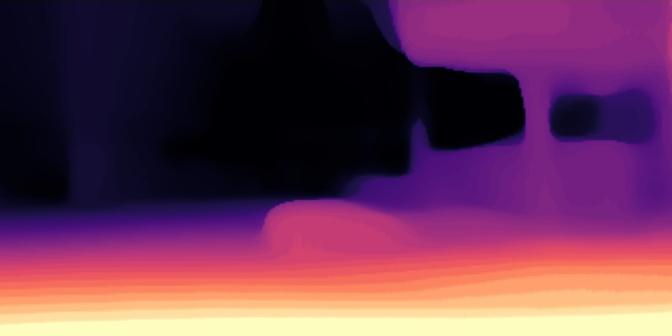}\end{tabular} &
\begin{tabular}{l}\includegraphics[width=0.15\linewidth]{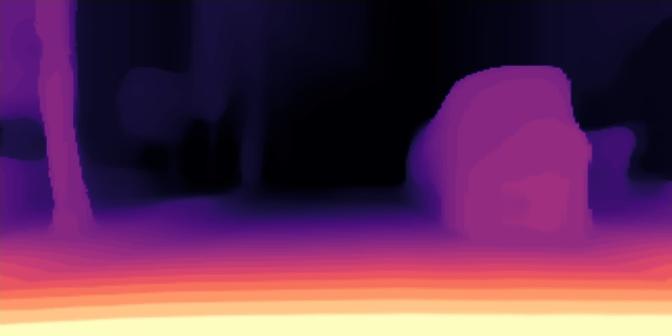}\end{tabular} &
\begin{tabular}{l}\includegraphics[width=0.15\linewidth]{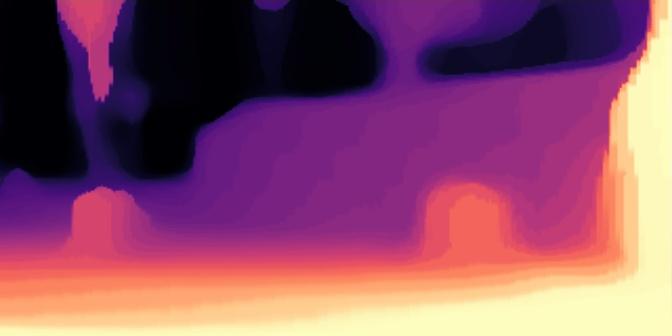}\end{tabular} \\

\end{tabular}

\caption{\textbf{Qualitative comparison of the depth estimation task on the nuScenes~\cite{nuscenes} dataset.} The output depths are presented in absolute terms and normalized between their maximum and minimum values for better visualization.}
\label{fig:fig_quali_comp}
% \vspace{-2mm}
\end{figure*}

\section*{Acknowledgment}
This work was supported in part by the National Natural Science Foundation of China under Grant 62125603, Grant 62321005, and Grant 62336004, and Shenzhen
Key Laboratory of Ubiquitous Data Enabling (Grant No. ZDSYS20220527171406015).

% \newpage
\appendices
\section{More Experimental Results}

\noindent \textbf{Per-camera evaluation:}
We present per-camera comparisons of our method for the depth estimation task against previous works~\cite{fsm,surrounddepth,r3d3} using the nuScenes dataset~\cite{nuscenes} in TABLE \ref{tab:nuscenes_per_cam}. Our approach consistently outperforms other methods across all camera views, showing particularly significant improvements in side views. Side views typically contain more intricate details, posing greater challenges for models to accurately estimate the depths.

\begin{table}[h]
\caption{\textbf{Per-camera comparisons for scale-aware multi-camera depth estimation on the nuScenes dataset}. Tests are conducted within 80 meters.}
\centering
\resizebox{1.0\linewidth}{!}{
\begin{tabular}{@{}l|cccccc @{}}
\toprule & \multicolumn{6}{c}{Abs Rel $\downarrow$}                          \\ \cmidrule(l){2-7} 
\textbf{Method} & \textit{Front} & \textit{F.Left} & \textit{F.Right} & \textit{B.Left} & \textit{B.Right} & \textit{Back} \\
\midrule
FSM~\cite{fsm}  & 0.186 & 0.287 & 0.375 & 0.296 & 0.418 & 0.221 \\
SurroundDepth~\cite{surrounddepth}  & 0.179 & 0.260 & 0.340 & 0.282 & 0.403 & 0.212 \\ 
R3D3~\cite{r3d3} & 0.174 & 0.230 & 0.302 & 0.249 & 0.360 & 0.201 \\
\midrule
\textbf{Ours} & \textbf{0.132} & \textbf{0.190} & \textbf{0.227} & \textbf{0.204} & \textbf{0.289} & \textbf{0.169} \\
\bottomrule %
\end{tabular}}
\label{tab:nuscenes_per_cam}
\end{table}

\noindent \textbf{Qualitative Comparisons:}
Fig. \ref{fig:fig_quali_comp} presents qualitative comparisons on the nuScenes~\cite{nuscenes} validation set. We visualize results from various state-of-the-art depth estimation and occupancy prediction methods using their official implementations. Compared to these methods, our occupancy-based approach exhibits fewer artifacts and enhanced overall accuracy. 

\section{The benefits of open-vocabulary.}

\begin{figure}[th]
    \centering
    \begin{subfigure}{0.32\linewidth}
        \includegraphics[width=\textwidth]{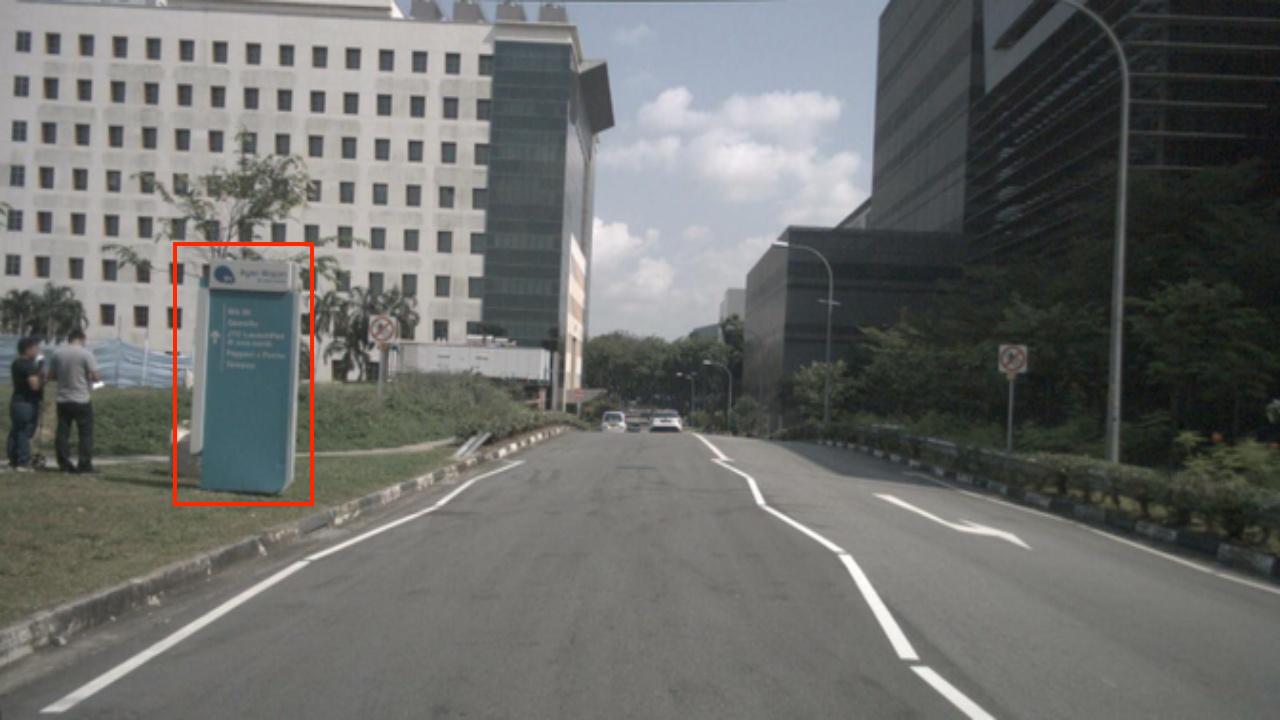}
        \caption{Input}
    \end{subfigure}
    \begin{subfigure}{0.32\linewidth}
        \includegraphics[width=\textwidth]{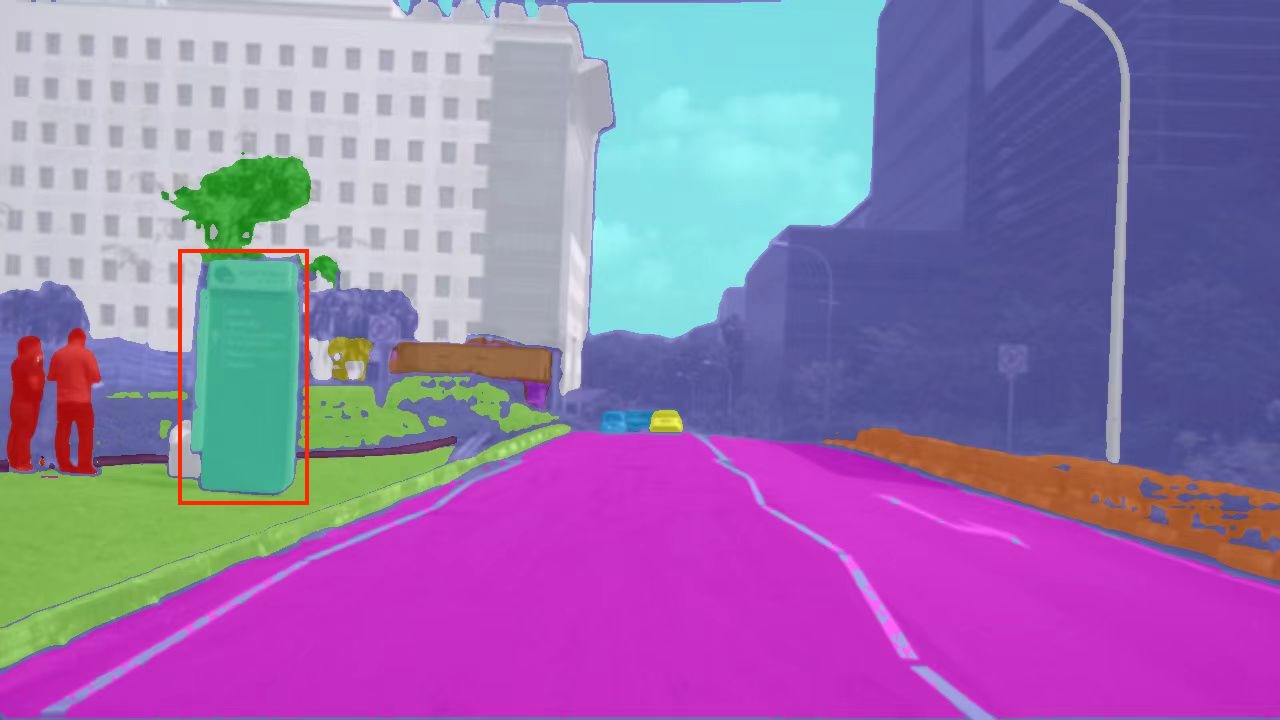}
        \caption{2D pseudo labels}
    \end{subfigure} 
    \begin{subfigure}{0.32\linewidth}
        \includegraphics[width=\textwidth]{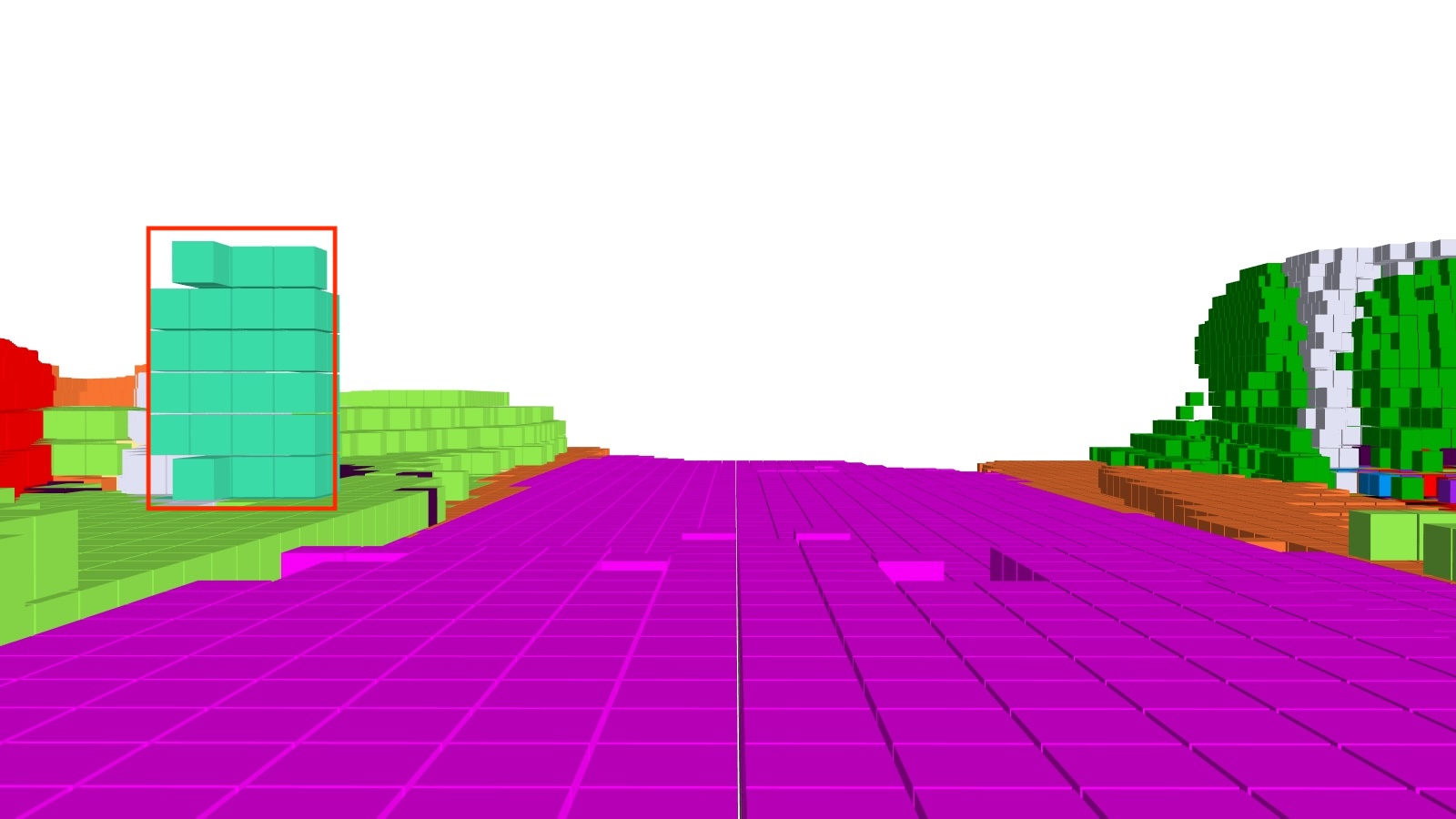}
        \caption{Occupancy}
    \end{subfigure} 
\caption{\textbf{Generating 3D occupancy for an unlabeled class in Occ3D~\cite{occ3d}}. When prompted with the term 'billboard', our model is capable of predicting the corresponding occupancy, despite the absence of this category in the official annotations.}
\label{fig:ov}
\end{figure}

Fig. \ref{fig:ov} illustrates that our pipeline utilizing the open-vocabulary models can generate results for undefined categories within the nuScenes dataset (e.g., `billboard' in the demonstration case). This adaptability enables our method to extend to arbitrary classes and be applicable to both publicly available and internally collected datasets. Furthermore, it demonstrates the potential for generalizing to rare classes, often referred to as corner cases, that are crucial in autonomous driving scenarios yet typically suffer from insufficient data. In contrast, traditional segmentation models pre-trained on the nuScenes dataset are confined to predefined classes and lack the flexibility to deal with extraordinary situations.


% % you can choose not to have a title for an appendix
% % if you want by leaving the argument blank
% \section{}

% \section*{Acknowledgment}
% The authors would like to thank Peiyang Li, 1008
% Danyang Zhang, Yu Zheng, Simin Wang, Yongming Rao, and 1009
% Tianmin Shu for their generous help.

\ifCLASSOPTIONcaptionsoff
  \newpage
\fi



%\bibliographystyle{IEEEtran}
\bibliographystyle{IEEEtran}
\bibliography{ref}

% \end{IEEEbiography}


% % insert where needed to balance the two columns on the last page with
% %\newpage



%\vfill



\vfill

% \enlargethispage{-10in}
\end{document}